\documentclass[final,12pt]{msml2021} % Anonymized submission
\usepackage{graphicx}
\graphicspath{{Figures/}}
\newcommand{\figscale}{0.26}
\usepackage{comment} %Comment environment
\usepackage{float}
\usepackage{multirow} %Tables
\usepackage{hyperref}
\usepackage{dsfont}
\usepackage{amsmath}

\newcommand{\ssobs}{\mathrm{obs}} % Observation subscript
\newcommand{\ssprior}{\mathrm{pr}} % Prior subscript
\newcommand{\sspost}{\mathrm{post}} % Posterior subscript
\newcommand{\sslkhd}{\mathrm{lkhd}} % Likelihood subscript
\newcommand{\sstrue}{\mathrm{true}} % True subscript
\newcommand{\ssmap}{\mathrm{map}} % MAP estimate subscript
\newcommand{\ssdraw}{\mathrm{draw}} % Draw estimate subscript
\newcommand{\dimparam}{D} % Parameter dimensions
\newcommand{\dimobs}{O} % Observation dimensions
\newcommand{\functparam}{u} % Parameter function
\newcommand{\functstate}{y} % State observations function
\newcommand{\param}{\textbf{u}} % Parameter bold
\newcommand{\state}{\textbf{y}} % State observations bold
\newcommand{\stateobs}{\textbf{y}_{\ssobs}} % State Observations
\newcommand{\forwardmap}{\mathcal{F}} % Forward map
\newcommand{\regmap}{\mathcal{R}} % Forward map
\newcommand{\NN}{\Psi} % Neural Network
\newcommand{\weights}{\boldsymbol{W}} % Weights
\newcommand{\weightsde}{\weights_{\mathrm{d}}} % Weights
\newcommand{\bias}{\boldsymbol{b}} % Biases
\newcommand{\model}{\mathcal{F}} % Forward Model
\newcommand{\modellinear}{\boldsymbol{F}} % Forward Model Linear
\newcommand{\decoder}{\Psi_{\mathrm{d}}} % Decoder
\newcommand{\numdatatrain}{M} % Number of training data
\newcommand{\ssnumdatatrain}{m} % Number of training data subscript
\newcommand{\numdatatest}{L} % Number of testing data
\newcommand{\ssnumdatatest}{l} % Number of testing data subscript
\newcommand{\dist}{p}
\newcommand{\paramrandom}{U}
\newcommand{\datarandom}{Y}
\newcommand{\randnoise}{E}
\newcommand{\distpost}{\dist_{\sspost}}
\newcommand{\distlkhd}{\dist_{\sslkhd}}
\newcommand{\distprior}{\dist_{\ssprior}}
\newcommand{\mean}{\boldsymbol{\mu}}
\newcommand{\cov}{\boldsymbol{\Gamma}}
\newcommand{\meanprior}{\mean_{\ssprior}}
\newcommand{\covprior}{\cov_{\ssprior}}
\newcommand{\meannoise}{\mean_{\randnoise}}
\newcommand{\noisevec}{\textbf{e}}
\newcommand{\regnoisemap}{\mathcal{M}} % Noise regularization
\newcommand{\regpriormap}{\mathcal{P}} % Prior regularization
\newcommand{\covnoise}{\cov_{\randnoise}} % Noise regularization
\newcommand{\regnoise}{\cov_{\randnoise}^{-1}} % Noise regularization
\newcommand{\regprior}{\cov_{\ssprior}^{-1}} % Prior regularization
\newcommand{\jacobian}{\boldsymbol{J}} % Jacobian
\newcommand{\noiselevel}{\eta} % Signal to noise ratio

\newcommand{\vardata}{\textbf{x}}
\newcommand{\paramgen}{\boldsymbol{\theta}}
\newcommand{\expct}{\mathbb{E}}

\newcommand{\varlatent}{\textbf{z}}
\newcommand{\distlatent}{\dist\left(\varlatent\right)}

\newcommand{\gaussian}{\mathcal{N}}
\newcommand{\isogaussian}{\gaussian\left(\varlatent|\boldsymbol{0}, \mateye\right)}
\newcommand{\distmodelpost}{q}
\newcommand{\paraminfer}{\boldsymbol{\phi}}
\newcommand{\KL}{\mathrm{KL}}

\newcommand{\meanpost}{\mean_{\sspost}}
\newcommand{\covpost}{\cov_{\sspost}}

\newcommand{\drawiso}{\boldsymbol{\epsilon}}

\newcommand{\JS}{\mathrm{JS}}

\newcommand{\penJS}{\alpha}
\newcommand{\truepost}{\dist(\param|\state)}

\newcommand{\truejoint}{\dist(\state,\param)}

\newcommand{\diststate}{\dist(\state)}
\newcommand{\paramdisttrue}{\param \sim \truepost}
\newcommand{\paramdistmodel}{\param \sim \distmodelpost_{\paraminfer}}
\newcommand{\modelpost}{\distmodelpost_{\paraminfer}(\param|\state)}

\newcommand{\covinfprior}{\mathcal{C}_{\ssprior}} % Infinite dimensional prior
\newcommand{\oppoisson}{\Delta} % Poisson Operator
\newcommand{\opinfpde}{\mathcal{A}} % Infinite dimensional prior
\newcommand{\blpcoeffone}{\delta} % BiLaplacian Prior Coefficient delta
\newcommand{\blpcoefftwo}{\gamma} % BiLaplacian Prior Coefficient gamma
\newcommand{\blpcoeffbnd}{\beta} % BiLaplacian Prior Coefficient Boundary

\newcommand{\std}{\sigma} % Standard deviation
\newcommand{\corr}{\rho} % Correlation
\newcommand{\domain}{\Omega} % Computational domain
\newcommand{\vecnormal}{\hat{\textbf{n}}} % Normal vector

\newcommand{\LRp}[1]{\left( #1 \right)}
\newcommand{\LRs}[1]{\left[ #1 \right]}

\newcommand{\mateye}{\boldsymbol{I}} % Identity matrix
\newcommand{\One}{\mathds{1}} % Vector of ones
\newcommand{\rmvec}{\mathrm{vec}} % Mathrm vec
\newcommand{\vecstd}{\boldsymbol{\std}} % Vector of standard deviation
\newcommand{\lowtri}{\boldsymbol{L}} % Lower portion of triangular matrix
\newcommand{\lowtrione}{\lowtri_{\One}} % Upper triangular matrix of ones
\title[Solving Bayesian Inverse Problems via Variational Autoencoders]{Solving Bayesian Inverse Problems via Variational Autoencoders}
\usepackage{times}

\msmlauthor{\Name{Hwan Goh} \Email{hwan.goh@gmail.com}\\
 \Name{Sheroze Sheriffdeen} \Email{sheroze@oden.utexas.edu}\\
 \Name{Jonathan Wittmer} \Email{jonathan.wittmer@utexas.edu}\\
 \Name{Tan Bui-Thanh} \Email{tanbui@oden.utexas.edu}\\
 \addr Oden Institute for Computational Engineering and Sciences\\
       The University of Texas at Austin\\
       Austin, TX 78712 \\
}

\makeatletter
 \let\Ginclude@graphics\@org@Ginclude@graphics
\makeatother

\begin{document}

\maketitle

\begin{abstract}%
    In recent years, the field of machine learning has made phenomenal progress
    in the pursuit of simulating real-world data generation processes. One
    notable example of such success is the variational autoencoder (VAE). In
    this work, with a small shift in perspective, we leverage and adapt VAEs for
    a different purpose: uncertainty quantification (UQ) in scientific inverse
    problems. We introduce UQ-VAE: a flexible, adaptive, hybrid
    data/model-constrained framework for training neural networks capable of rapid
    modelling of the posterior distribution representing the unknown parameter
    of interest. Specifically, from divergence-based variational inference, our
    framework is derived such that most of the information usually present in
    scientific inverse problems is fully utilized in the training procedure.
    Additionally, this framework includes an adjustable hyperparameter that
    allows selection of the notion of distance between the posterior model and
    the target distribution. This introduces more flexibility in controlling how
    optimization directs the learning of the posterior model. Further, this
    framework possesses an inherent adaptive optimization property that emerges
    through the learning of the posterior uncertainty. Numerical results for an
    elliptic PDE-constrained Bayesian inverse problem are provided to verify the
    proposed framework.
\end{abstract}


%==============================================================================
\section{Introduction}
%==============================================================================
The challenge of generative modelling is often approached through optimization
of a parameterized likelihood function $\dist_{\paramgen}(\vardata)$ to learn
the optimal parameters $\paramgen$. Latent variable models augment this
likelihood function using an unobservable latent variable in hopes that one
can also learn useful latent representations that facilitate reconstruction and generation
of high-dimensional distributions. Variational autoencoders (VAE) were
introduced in \cite{kingma2013auto} to train a generating
function that maps from the latent space to the target data distribution.  The
approach in \cite{kingma2013auto} assumes the latent distribution to simply be
an isotropic Gaussian $\distlatent = \isogaussian$; thereby shifting all the
burden of generative modelling to the learning of the generating function. To
address the infeasibly large number of draws from the
isotropic Gaussian required to ensure significant contribution to the likelihood
function, \cite{kingma2013auto} proposes to instead draw from the posterior
distribution $\dist_{\paramgen}(\varlatent|\vardata)$. This introduces the
subtask of variational inference; the modelling of the posterior with a
distribution $\distmodelpost_{\paraminfer}(\varlatent|\vardata)$ parameterized
by an additional set of learnable parameters $\paraminfer$. The learning of the
parameters $\paraminfer$ can be achieved through the minimization of the Evidence
Lower Bound (ELBO) \cite{blei2017variational}; a quantity obtainable through the
Kullback-Leibler divergence (KLD) between the model posterior and the target
posterior.

Although the goal of generative modelling fundamentally differs from that of
scientific inverse problems, we claim that VAEs are also well
suited for the latter challenge. In the setting of physical systems, scientific
inverse problems task us with determining parameters-of-interest (PoI) given
observations of a state variable. Often, these parameters are coefficients in a partial
differential equation (PDE) governing the system and the observations are
discrete measurements of the state variables of the PDE. Assuming that our
observations are corrupted by additive noise, then the equation
\begin{align}
    \stateobs
    =
    \model(\param)
    +
    \noisevec
\end{align}
represents the connection between the PoI and observational data. Here, $\param$
denotes a discretized representation of the PoI, $\model$ denotes the
parameter-to-observable (PtO) map and $\stateobs$ denotes our observational data.

To bridge the two contexts, we consider the latent space and generating function
in latent variable models to be the PoI space and the
PtO map in inverse problems. Therefore, while the former
uses inference to facilitate generative modelling, the latter uses generative
modelling to facilitate inference. We contrast the two settings under this
viewpoint in order to expose the utility of VAEs for scientific inverse
problems:
\begin{enumerate}
    \item For generative modelling, the focus of training VAEs is to learn the
        generating function. In contrast, for scientific inverse problems, the
        PtO map need not be learnt as
        it can often be modelled using physics-based numerical methods.
    \item For generative modelling, the latent space of VAEs is an unspecified
        quantity with properties emerging almost entirely though the process of
        learning the generating function. In contrast, for scientific inverse
        problems, by considering the latent space to be the PoI space,
        the latent space therefore possesses structure that represents the physical
        properties of the PoI.
\end{enumerate}
Therefore, the context of scientific inverse problems provides more
information that can be exploited to improve the training procedure of a VAE.
Indeed, in reference to the first point, one can surrogate the
generating function with a physics-based numerical model of the
PtO map. In doing so, the focus of the training is shifted
completely towards inference. In reference to the second point, knowledge of the
physical properties of the PoI can be encoded in the prior
model which, therefore, allows for a more informative latent distribution than the
isotropic Gaussian used in \cite{kingma2013auto}.
Further, through scientific experimentation or simulation, one may have on
hand a dataset of PoI and observation pairs
$\left\{\left(\param^{(\ssnumdatatrain)}
,\stateobs^{(\ssnumdatatrain)}\right)\right\}_{\ssnumdatatrain=1}^{\numdatatrain}$.
To draw an analogy within the context of generative modelling, the datapoints
$\param^{(\ssnumdatatrain)}$ are analogous to latent data that is usually not
available. This advantage provides the main motivation for our proposed UQ-VAE
framework: a mathematically justified training procedure for a VAE
that can utilize the latent data we possess.

%==============================================================================
\section{Preliminaries and Motivation}
%==============================================================================
In scientific inverse problems, the solution process for obtaining estimates of
the PoI often involves the optimization of a functional of the
form
\begin{align}
    \min_{\param}
    \left\lVert \stateobs - \model(\param) \right\rVert_{2}^{2}.
    \label{EqDataMisfit}
\end{align}
Whilst the observations are finite-dimensional entities, the
PoI often exists in an infinite-dimensional function space.
Thus, our task is complicated by the inherent ill-posed nature of inverse
problems; there are many possible estimates of the PoI that
are consistent with the observations. To alleviate this issue, regularization
can be introduced to essentially reduce the size of the solution space. With
the introduction of a regularization term, the optimization problem
(\ref{EqDataMisfit}) becomes
\begin{align}
    \min_{\param}
    \left\lVert \stateobs - \forwardmap(\param)
    \right\rVert_{2}^{2}
    + \regmap(\param). \label{EqRegDataMisfit}
\end{align}
The optimization of (\ref{EqRegDataMisfit}) is often a time-consuming process;
for example when using derivative-based iterative optimization algorithms. This
motivates the need for an inverse problems solver that can rapidly output PoI
estimates from given observational data.

%------------------------------------------------------------------------------
\subsection{Learning a Solver for Deterministic Inverse Problems}
\label{SecDeterministicInverseProblemSolver}
%------------------------------------------------------------------------------
We briefly discuss data-driven deterministic inverse problem solvers to motivate
both the concept of UQ as well as our proposed training procedure.

Suppose that we are tasked with learning an inverse problem solver using a
training dataset of $\numdatatrain$ PoI and observation pairs
$\left\{\left(\param^{(\ssnumdatatrain)}
,\stateobs^{(\ssnumdatatrain)}\right)\right\}_{\ssnumdatatrain=1}^{\numdatatrain}$.
If we elect to use a neural network $\NN$, then our solver is parameterized by
the weights $\weights$ of the network. Analogous to (\ref{EqRegDataMisfit}), the
training of this solver requires the optimization of the following functional:
\begin{align}
    \min_{\weights}
    \frac{1}{\numdatatrain}\sum_{\ssnumdatatrain=1}^{\numdatatrain}
    \left\lVert \param^{(\ssnumdatatrain)}
        -
    \NN\left(\stateobs^{(\ssnumdatatrain)}, \weights\right)\right\rVert_{2}^{2}
        +
    \mathcal{R}(\weights). \label{EqNNDataMisfit}
\end{align}
Following the offline training stage of the neural network, we obtain a rapid
online inverse problem solver that can output an estimate of our PoI given
observation data.  However, the ill-posed nature of inverse problems is
inherited by this task; there are many possible weights that can be used to
parameterize a solver that outputs PoI estimates consistent with the input
observational data. Further, while the regularization term in
(\ref{EqRegDataMisfit}) can be designed based on some prior knowledge of the
physical properties of the PoI, the physical interpretation of the neural
network weights is not so clear. Therefore, there is no natural choice for the
regularization of the optimization problem (\ref{EqNNDataMisfit}).

Instead of regularizing the weights of the network directly, one possible
approach is to regularize the output of the network. For example, we could make
use of the PtO map so that (\ref{EqNNDataMisfit}) becomes
\begin{align}
    \min_{\weights}
    \frac{1}{\numdatatrain}\sum_{\ssnumdatatrain=1}^{\numdatatrain}
    \left\lVert \param^{(\ssnumdatatrain)}
        -
        \NN\left(\stateobs^{(\ssnumdatatrain)},
        \weights\right)\right\rVert_{2}^{2}
        +
        \left\lVert \stateobs^{(\ssnumdatatrain)} -
            \model\left(\NN\left(\stateobs^{(\ssnumdatatrain)},
        \weights\right)\right)\right\rVert_{2}^{2}. \label{EqRegTrueNNDataMisfit}
\end{align}
In doing so, the PtO map indirectly informs the weights of
the neural network of the inversion task. An additional benefit with the
inclusion of the PtO map is that we are also able to include
noise regularization. That is, if we possess some knowledge about the properties of
the noise $\noisevec$ afflicting our observational data, we may quantify it and include it
in the second term to yield
\begin{align}
    \min_{\weights}
    \frac{1}{\numdatatrain}\sum_{\ssnumdatatrain=1}^{\numdatatrain}
    \left\lVert \param^{(\ssnumdatatrain)}
        -
        \NN\left(\stateobs^{(\ssnumdatatrain)},
        \weights\right)\right\rVert_{2}^{2}
        +
        \left\lVert \regnoisemap\left(\stateobs^{(\ssnumdatatrain)} -
                \model\left(\NN\left(\stateobs^{(\ssnumdatatrain)},
    \weights\right)\right)\right)\right\rVert_{2}^{2}
\end{align}
where $\regnoisemap$ represents some noise regularization mapping. Furthermore, if
we possess some prior knowledge about the physical properties of the parameter
of interest, we may also include it as follows:
\begin{subequations}
    \begin{align}
        \min_{\weights}
        \frac{1}{\numdatatrain}\sum_{\ssnumdatatrain=1}^{\numdatatrain}
        \left\lVert \param^{(\ssnumdatatrain)}
            - \NN\left(\stateobs^{(\ssnumdatatrain)},
        \weights\right)\right\rVert_{2}^{2}
        &+
        \left\lVert \regnoisemap\left(\stateobs^{(\ssnumdatatrain)} -
                \model\left(\NN\left(\stateobs^{(\ssnumdatatrain)},
        \weights\right)\right)\right)\right\rVert_{2}^{2}\\
        &+
        \left\lVert
            \regpriormap\left(\NN\left(\stateobs^{(\ssnumdatatrain)}, \weights \right)
        \right)\right\rVert_{2}^{2}
    \end{align}\label{Eqfwdinvloss}
\end{subequations}
where $\regpriormap$ is some prior regularization mapping.

Training a neural network through the optimization problem (\ref{Eqfwdinvloss})
yields a learned inverse problems solver that outputs a point estimate of our
PoI. As it is, this deterministic solver is unable
to provide information about the accuracy of the estimate. It would be more
ideal to have a probabilistic interpretation of our learned solver that
facilitates UQ. With this in mind, we are motivated to
view inverse problems under the framework of Bayesian statisics.
\emph{In this setting, we instead work towards a solver for Bayesian inverse
    problems which, in turn, allows us to formally establish the regularization
    terms in (\ref{Eqfwdinvloss}). Conversely, by ignoring the uncertainty
    estimate, the UQ-VAE solver recovers the deterministic setting
    (\ref{Eqfwdinvloss}).}

%------------------------------------------------------------------------------
\subsection{Learning a Solver for Bayesian Inverse Problems} \label{SecLearningBIP}
%------------------------------------------------------------------------------
Under the statistical framework, the PoI of an inverse problem
is considered to be a random variable instead of an unknown value. Consequently, the
solution of the statistical inverse problem is a probability distribution
instead of a single estimated value. By adopting a statistical framework instead of
deterministic, the question asked by the inverse problem essentially
changes from ``what is the value of our parameter?" to ``how accurate
is the estimate of our parameter?". Further, the statistical framework
attempts to remove the ill-posedness of inverse problems by restating the
inverse problem as a well-posed extension in a larger space of probability
distributions \cite{kaipio2006statistical, dashti2013bayesian}.

With an additive noise assumption, we consider the following
observational model
\begin{align}
\datarandom = \model\left(\paramrandom\right) + \randnoise \label{EqRandObsModel}
\end{align}
where $\model$ is the PtO map and $\datarandom$, $\paramrandom$ and $\randnoise$
are the random variables representing the observational data, PoI and noise
model respectively. Since $\paramrandom$ is unknown to us, we represent its
uncertainty with a probability distribution constructed from prior
knowledge. However, we do not use equation (\ref{EqRandObsModel}) directly when
working under a statistical framework. Instead, we primarily consider the
relations between the probability distributions of the random variables involved
to answer the question: ``given the observational data $\stateobs$, what is the
distribution of the PoI $\paramrandom$ responsible for our measurement?''.
Therefore, the conditional density
$\dist_{\paramrandom|\datarandom}\left(\param|\datarandom=\stateobs\right)$ is
the solution to the statistical parameter estimation problem under the Bayesian
framework. To approximate this conditional density, we utilize Bayes' Theorem to
form a model of
$\dist_{\paramrandom|\datarandom}\left(\param|\datarandom=\stateobs\right)$
called the posterior distribution which we denote as $\distpost$:
\begin{align}
    \distpost\left(\param|\stateobs\right)\propto
    \distlkhd\left(\stateobs|\param\right)\distprior\left(\param\right).
    \label{EqPostProb}
\end{align}
where $\distlkhd$ is the likelihood model and $\distprior$ is the
prior model.

Formulating the parameter estimation problem under a Bayesian
framework requires us to obtain (\ref{EqPostProb}). This challenges us with
the completion of three tasks:
\begin{enumerate}
    \item construct the likelihood model $\distlkhd$ that expresses
        the interrelation between the data and the unknown,
    \item using prior
        information we may possess about the unknown $\param$, construct a prior
        probability density $\distprior$ that expresses this information,
    \item develop methods which extract meaningful information from the
        posterior density $\distpost$.
\end{enumerate}
To address these three tasks, two assumptions are
often made. The first assumption supposes that the noise $\randnoise$ is mutually independent
with respect to the PoI $\paramrandom$. Then, using our
observation model (\ref{EqRandObsModel}) and marginalization of the noise
$\randnoise$, we obtain the following likelihood model:
\begin{align}
    \distlkhd = \dist_{\randnoise}\left(\stateobs -
    \model\left(\param\right)\right).
\end{align}
The second assumption supposes that all are random variables are Gaussian with
$\mathcal{N}(\meannoise,\covnoise)$ and $\mathcal{N}(\meanprior,\covprior)$.
With this, our posterior model becomes
\begin{subequations}
    \begin{align}
        \distpost(\param|\stateobs)
        &\propto
            \dist_{\randnoise}(\stateobs-\model(\param))\distprior(\param)\\
        &=
        \exp {\left(-\frac{1}{2}\left(\left\lVert \stateobs-\model(\param) -
                    \meannoise\right\rVert_{\regnoise}^2 +
    \left\lVert\param-\meanprior\right\rVert_{\regprior}^2\right) \right)}.
    \label{EqBayesPostExp}
    \end{align}
    \label{EqBayesPost}
\end{subequations}
In order to perform UQ with given observational data
$\stateobs$, one often seeks an approximation of the posterior covariance of
(\ref{EqBayesPost}). A commonly used approximation is the Laplace approximation
\cite{evans2000approximating,press2009subjective,stigler1986laplace,tierney1986accurate,wong2001asymptotic}
\begin{align}
    \covpost = \left(
    \jacobian_{\model}(\param_{\ssmap})^\mathrm{T}
    \covnoise^{-1}
    \jacobian_{\model}(\param_{\ssmap})
    +
    \covprior^{-1}
    \right)^{-1}
    \label{EqLaplaceApprox}
\end{align}
where $\param_{\ssmap}$ is the maximum a posteriori (MAP) estimate and
$\jacobian_{\model}(\param_{\ssmap})$ is the Jacobian of $\model$ at the MAP
estimate. The MAP estimate can be obtained through optimization of the
exponentiated functional in (\ref{EqBayesPostExp}); which often requires a
potentially expensive derivative-based iterative optimization procedure.
Markov Chain Monte Carlo (MCMC) methods can also be employed but such methods
notably suffer the curse of dimensionality.

This desire for efficient UQ also motivates our work.
Although the training procedure of a Bayesian inverse problems solver may be
computationally demanding, it represents an offline stage. Once trained, the
online stage of UQ has a computational cost governed
only by the forward propagation of the observational data through the trained
neural network; which is often a very computationally efficient procedure.

Additionally, the PoI data $\param^{(\ssnumdatatrain)}$ in the training dataset
$\left\{\left(\param^{(\ssnumdatatrain)}
,\stateobs^{(\ssnumdatatrain)}\right)\right\}_{\ssnumdatatrain=1}^{\numdatatrain}$
usually plays no role in traditional methods. Information about the PoI is often
only encoded in the prior model. This motivates the data-driven aspect of our approach in
that we not only use prior information but we also take advantage of PoI data
in the cases where paired PoI-observation datasets are readily
obtainable.

With this motivation in mind, we now detail a proposed framework for training our solver.
When comparing with (\ref{Eqfwdinvloss}) from Section
\ref{SecDeterministicInverseProblemSolver}, the exponentiated functional in
(\ref{EqBayesPostExp}) looks like an appealing candidate for the regularization
terms. We formalize the inclusion of these terms as regularizers through the
derivation of our proposed framework for data-driven UQ.

%==============================================================================
\section{Flexible, Adaptive, Hybrid Data/Model-Constrained Framework for UQ}
%==============================================================================
We begin by detailing the derivation of our proposed UQ-VAE framework before
discussing from the more tractable perspective of optimization. While the former
centers the discussion more in the setting of variational inference, the latter
is centered more towards regularization of the training procedure for our
solver. Further, whilst the flexibility of our framework emerges through the
viewpoint of variational inference, the optimization viewpoint uncovers
its inherent adaptive properties.

%------------------------------------------------------------------------------
\subsection{Derivation of the UQ-VAE Framework} \label{SecDeriveUQVAE}
%------------------------------------------------------------------------------
Let $\dist(\param|\state)$ denote the target posterior density we wish to
estimate and let $\distmodelpost_{\paraminfer}(\param|\state)$ denote our model
of the target density with statistics parameterized by $\paraminfer$. To optimize for the
parameters $\paraminfer$, we require some notion of distance between our model
posterior and target posterior. In this work, instead of using the KLD to obtain
the ELBO, we elect to use the following family of Jensen-Shannon
divergences (JSD)
\cite{nielsen2010family}:
    \begin{align}
        \JS_{\penJS}(\distmodelpost||\dist)
        &=
        \penJS\KL(\distmodelpost||(1-\penJS)\distmodelpost + \penJS\dist)+
        (1-\penJS)\KL(\dist||(1-\penJS)\distmodelpost + \penJS\dist).
        \label{EqJSDFamily}
    \end{align}
With this, we introduce the following theorem:
\begin{theorem} \label{ThmGenJSVariationalInference}
   Let $\penJS \in (0,1)$. Then
   \begin{subequations}
        \begin{align}
            \frac{1}{\penJS}\JS_{\penJS}(\distmodelpost_{\paraminfer}(\param|\state)||
            \dist(\param|\state))
            =
            &- \expct_{\param \sim
                \distmodelpost_{\paraminfer}}
                \left[\log \left(\penJS +
                        \frac{(1-\penJS)\distmodelpost_{\paraminfer}(\param|\state)}
                {\dist(\param|\state)}\right)\right] \label{Eqlog1}\\
            &+\log(\dist(\state))\\
            &- L_{\JS_{\penJS}}(\paraminfer, \state)
        \end{align} \label{EqGenJSmodeldensityequality}
    \end{subequations}
    where
    \begin{align}
        L_{\JS_{\penJS}}(\paraminfer, \state)
        &=
        \frac{1-\penJS}{\penJS}\expct_{\param \sim
        \dist(\param|\state)}
        \left[\log\left(\penJS +
                \frac{(1-\penJS)\distmodelpost_{\paraminfer}(\param|\state)}
        {\dist(\param| \state)}\right)\right] \label{Eqlog2}
        +
        \expct_{\param \sim
        \distmodelpost_{\paraminfer}}
        \left[
        \log\left(\frac{\dist(\state, \param)}{
        \distmodelpost_{\paraminfer}(\param|\state)}\right)\right].
    \end{align} \label{EqGenJSlowerbound}
\end{theorem}
The proof is given in Section \ref{SecProofThmGenJS} of the Appendix. From an
implementation perspective, Theorem \ref{ThmGenJSVariationalInference} does not
offer much insight. In particular, we would like to have direct access to the
quotients within the expectations. Therefore, we offer the following corollary
which yields useful KLD terms:
\begin{corollary} \label{CorrGenJSLowerbound}
    Let $\penJS \in (0,1)$ and consider again (\ref{EqGenJSmodeldensityequality}).
    Equation (\ref{EqGenJSmodeldensityequality}) is bounded above
    such that:
    \begin{subequations}
        \begin{align}
            \frac{1}{\penJS}\JS_{\penJS}(\distmodelpost_{\paraminfer}(\param|\state)||
            \dist(\param|\state))
            \leq
            &-\KL(\distmodelpost_{\paraminfer}(\param|\state)||
               \dist(\param|\state))\\
            &+\log(\dist(\state))
            - \log(1-\penJS) - \frac{(1-\penJS)\log(1-\penJS)}{\penJS}\\
            &+
            \frac{1-\penJS}{\penJS}\KL(\dist(\param|\state)||
            \distmodelpost_{\paraminfer}(\param|\state))\\
            &-\expct_{\param \sim \distmodelpost_{\paraminfer}}
            \left[\log\left(\dist(\state|\param)\right)\right]
            +
            \KL\left(\distmodelpost_{\paraminfer}(\param|\state)||\dist(\param)\right).
        \end{align} \label{EqNewGenVLB}
    \end{subequations}
    In particular, we have that
    \begin{subequations}
        \begin{align}
            -L_{\JS_{\penJS}}(\paraminfer, \state)
            \leq
            -\frac{(1-\penJS)\log(1-\penJS)}{\penJS}
            &+
            \frac{1-\penJS}{\penJS}\KL(\dist(\param|\state)||
            \distmodelpost_{\paraminfer}(\param|\state))\\
            &-\expct_{\param \sim \distmodelpost_{\paraminfer}}
            \left[\log\left(\dist(\state|\param)\right)\right]
            +
            \KL\left(\distmodelpost_{\paraminfer}(\param|\state)||\dist(\param)\right).
            \label{EqLBofGenJSVLB}
        \end{align}
    \end{subequations}
    \begin{proof}
        The bounds are obtained simply by dropping the $\penJS$ from within the
        logarithms in (\ref{Eqlog1}) and (\ref{Eqlog2}). From there, the
        assertions are easily obtained.
    \end{proof}
\end{corollary}
The significance of Corollary \ref{CorrGenJSLowerbound} is that the minimization
of the upper bound
\begin{align}
    \frac{1-\penJS}{\penJS}\KL(\dist(\param|\state)||
    \distmodelpost_{\paraminfer}(\param|\state))
    -\expct_{\param \sim \distmodelpost_{\paraminfer}}
    \left[\log\left(\dist(\state|\param)\right)\right]
    + \KL\left(\distmodelpost_{\paraminfer}(\param|\state)||\dist(\param)\right)
    \label{EqJSVUB}
\end{align}
with respect to $\paraminfer$ could minimize
\begin{align}
    \frac{1}{\penJS}\JS_{\penJS}(\distmodelpost_{\paraminfer}(\param|\state)||
    \dist(\param|\state))
    +
    \KL(\distmodelpost_{\paraminfer}(\param|\state)||
    \dist(\param|\state)) \label{EqJSKLInference}
\end{align}
which would, therefore, perform the task of variational inference. From the
perspective of standard VAEs, notice that the second and third term of (\ref{EqJSVUB})
together form the negative of the ELBO. Relating back to the
exponentiated functional (\ref{EqBayesPostExp}), from the perspective of
Bayesian inverse problems, the second term in (\ref{EqJSVUB}) is the likelihood
model containing the PtO map and the third term contains the prior model
$\dist(\param)$ representing information about the PoI. The first term in
(\ref{EqJSVUB}) represents information regarding the target posterior. This is
the key achievement of utilizing the JSD as the notion of distance between the
model posterior and target posterior as it is this term that allows for the
inclusion of paired PoI-observation datasets.

Through the choice of $\penJS \in (0,1)$, the family of JSDs (\ref{EqJSDFamily})
allows for interpolation between the zero-forcing $\KL(\dist||\distmodelpost)$
as $\penJS \to 0$ and the zero-avoiding $\KL(\distmodelpost||\dist)$ as $\penJS
\to 1$ \cite{bishop2006pattern,murphy2012machine}. With this in mind, we make
the following observations:
\begin{itemize}
    \item In (\ref{EqJSDFamily}), it is clear that if $\penJS=1$ then we recover
        the usual zero-avoiding $\KL(\distmodelpost||\dist)$ used in \cite{kingma2013auto}.
        Indeed, as $\penJS \to 1$, then the first term in (\ref{EqJSVUB}) tends
        to $0$ which recovers the negative of the ELBO.
    \item In (\ref{EqJSKLInference}), the presence of the KLD term ensures that
        our model posterior will inherently be zero-avoiding. However, the
        $\frac{1}{\penJS}$ scaling factor ensures that as $\penJS \to 0$,
        the consequently zero-forcing JSD dominates the KLD.
\end{itemize}
Therefore, our UQ-VAE framework essentially retains the full flexibility of the
JSD family. Thus, with only an adjustment of a single scalar value, our
framework allows the selection of the notion of distance used by the
optimization routine to direct the model posterior towards the target posterior.
This, in turn, translates to control of the balance of data-fitting and
regularization used in the training procedure as discussed in the following
section.

%------------------------------------------------------------------------------
\subsection{Regularized Optimization Problem} \label{SecRegOptProblem}
%------------------------------------------------------------------------------
We now move our discussion towards implementation by detailing the
optimization problem implied by the UQ-VAE framework. In particular, we
employ approximations of (\ref{EqJSVUB}) from Section
\ref{SecDeriveUQVAE} to yield an implementable loss functional.

We begin by forming an approximation to the first term in (\ref{EqJSVUB}).
First, we apply the well-known fact that the minimization of the KLD between the
empirical and model distributions is equivalent to maximization of the
likelihood function with respect to $\paraminfer$. Next, we form a Monte-Carlo
estimation using our PoI data. However, since every observation datapoint
$\stateobs^{(\ssnumdatatrain)}$ is associated with only one PoI datapoint
$\param^{(\ssnumdatatrain)}$, our approximation is somewhat crude. \footnote{Note
that more than one prior sample can be used here.}
Finally, we adopt a Gaussian model for our model posterior:
$\distmodelpost_{\paraminfer}\left(\param|\stateobs^{(\ssnumdatatrain)}\right) =
\gaussian\left(\param|\meanpost^{(\ssnumdatatrain)},\covpost^{(\ssnumdatatrain)}\right)$.
These assumptions can be summarized by the following chain of
equations:
\begin{subequations}
    \begin{align*}
        \expct_{\param\sim
        \dist\left(\param|\stateobs^{(\ssnumdatatrain)}\right)}
        \left[-\log\left(
        \distmodelpost_{\paraminfer}(\param|\stateobs^{(\ssnumdatatrain)})\right)\right]
        & \lesssim \expct_{\param\sim
        \dist\left(\param\right)}
        \left[-\log\left(
        \distmodelpost_{\paraminfer}(\param|\stateobs^{(\ssnumdatatrain)})\right)\right]
        \approx
        -\log\left(
        \distmodelpost_{\paraminfer}(\param^{(\ssnumdatatrain)}|
        \stateobs^{(\ssnumdatatrain)})\right)\\
        &=
        \frac{\dimparam}{2}\log(2\pi) +
        \frac{1}{2}\log\left|\covpost^{(\ssnumdatatrain)}\right| +
        \frac{1}{2}\left\|\meanpost^{(\ssnumdatatrain)} -
        \param^{(\ssnumdatatrain)}\right\|_{\covpost^{(\ssnumdatatrain)-1}}^{2}.
    \end{align*}
\end{subequations}
Now we incorporate deep learning into the framework. We consider a neural network
that takes in our observation data $\stateobs^{(\ssnumdatatrain)}$ as an input
to output the statistics $\left(\meanpost^{(\ssnumdatatrain)},
\covpost^{(\ssnumdatatrain)}\right)$ of our posterior model. In doing so, we are
reparameterizing the statistics of our Gaussian posterior model represented by
$\paraminfer$ with the weights $\weights$ of the neural network $\NN$; thereby increasing the
learning capacity of our model.

As mentioned in Section \ref{SecLearningBIP}, it is common to adopt Gaussian noise
and prior models $\mathcal{N}(\meannoise,\covnoise)$ and
$\mathcal{N}(\meanprior,\covprior)$. With this and
our approximation above, we obtain the following optimization problem:
\begin{subequations}
    \begin{alignat}{3}
        \min_{\weights}
        &\frac{1}{\numdatatrain}\sum_{\ssnumdatatrain=1}^{\numdatatrain}
        &&\frac{1-\penJS}{\penJS}
        \left(
            \log
            \left|
                \covpost^{(\ssnumdatatrain)}
            \right|
            +
            \left\|
                \meanpost^{(\ssnumdatatrain)}
                -
                \param^{(\ssnumdatatrain)}
            \right\|_{\covpost^{(\ssnumdatatrain)-1}}^{2}
        \right)
        \label{EqOptPost}\\
        & &&+
        \left\lVert
            \stateobs^{(\ssnumdatatrain)}
            -
            \model\left(
                \param_{\ssdraw}^{(\ssnumdatatrain)}(\weights)
            \right)
            -
            \meannoise
        \right\rVert_{\regnoise}^{2}
        \label{EqOptLkhd}\\
        & &&+
        \mathrm{tr}\left(
            \covprior^{-1}
            \covpost^{(\ssnumdatatrain)}
        \right)
        +
        \left\|
            \meanpost^{(\ssnumdatatrain)}
            -
            \meanprior
        \right\|_{\regprior}^{2}
        +
        \log\frac{
            \left|
                \covprior
            \right|
        }
        {
            \left|
                \covpost^{(\ssnumdatatrain)}
            \right|
        }
        \label{EqOptPrior}\\
        &\mathrm{where} &&
        \left(
            \meanpost^{(\ssnumdatatrain)},
            \covpost^{\frac{1}{2}(\ssnumdatatrain)}
        \right)
        =
        \NN\left(
            \stateobs^{(\ssnumdatatrain)},\weights
        \right),\\
        & && \param_{\ssdraw}^{(\ssnumdatatrain)}(\weights)
        =
        \meanpost^{(\ssnumdatatrain)}
        +
        \covpost^{\frac{1}{2}(\ssnumdatatrain)}
        \drawiso^{(\ssnumdatatrain)},\\
        & && \drawiso^{(\ssnumdatatrain)}
        \sim
        \gaussian(\boldsymbol{0},\mateye_{\dimparam})
    \end{alignat}
    \label{EqOpt}
\end{subequations}
Drawing connections with our discussion in Section
\ref{SecDeterministicInverseProblemSolver}, our data-misfit
(\ref{EqOptPost}) containing our PoI data is regularized by
the likelihood model (\ref{EqOptLkhd}) containing the PtO
map and the prior model (\ref{EqOptPrior}) containing information on our
PoI space.

\begin{figure}[H]
    \centering
    \includegraphics[scale=0.4]{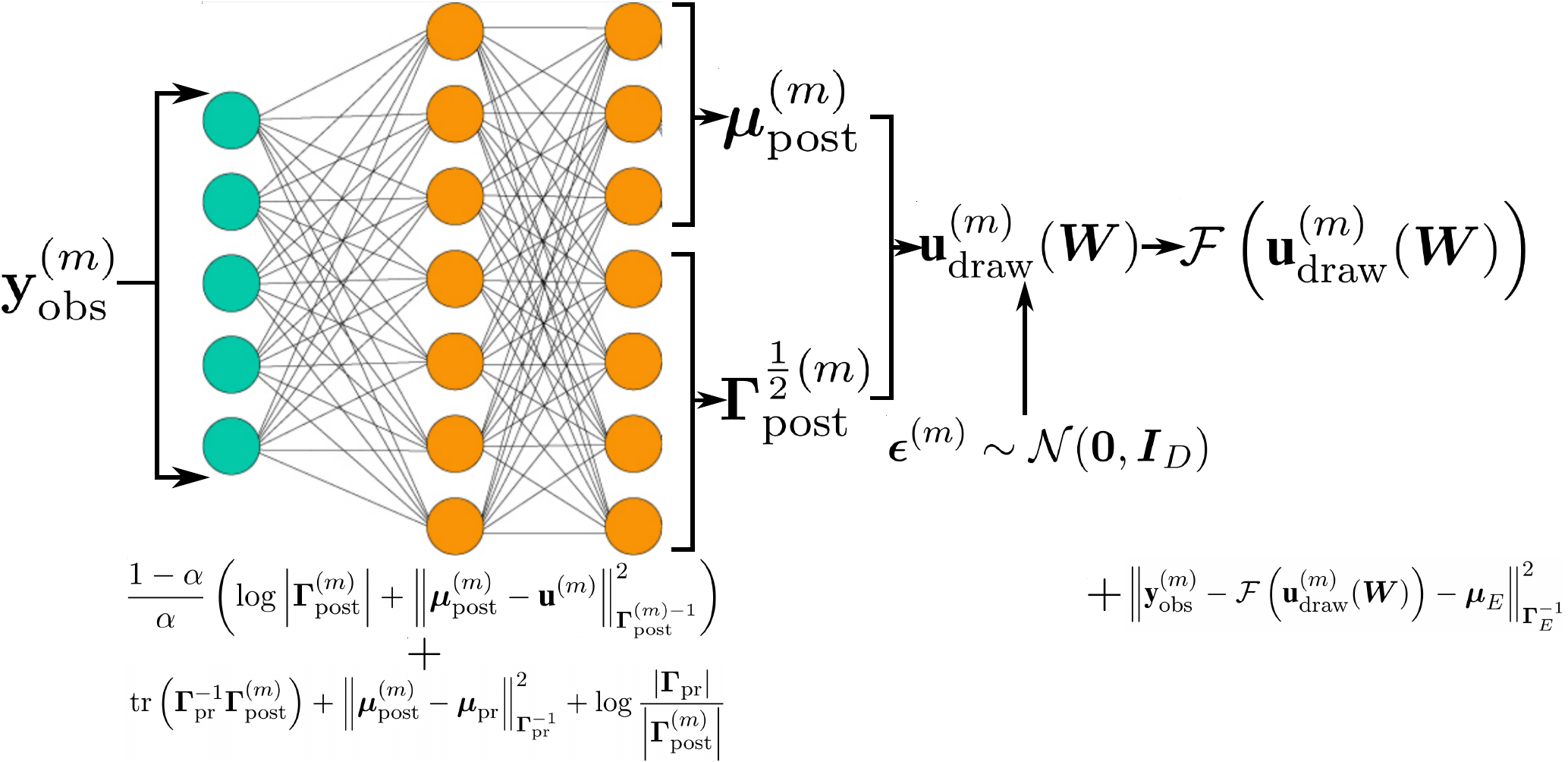}
    \caption{Schematic of the UQ-VAE framework with the exact PtO map.}
    \label{FigUQVAE}
\end{figure}

In (\ref{EqOptPost}), the $\frac{1-\penJS}{\penJS}$ factor allows $\penJS$ to
adjust the influence of the PoI data and regularization over the optimization procedure.
However, the onus of balancing data-fitting and regularization is not completely
on the hyperparameter $\penJS$. The presence of the matrix
$\covpost^{(\ssnumdatatrain)-1}$ in the weighted norm of
(\ref{EqOptPost}) acts as an adaptive penalty for the data-misfit term. Indeed,
during training, this matrix changes as the network weights $\weights$, on which
the matrix depends on, are optimized. With this in mind, we make the following
observations about the two terms in (\ref{EqOptPost}):
\begin{itemize}
    \item Since $\lim_{\penJS \to 0} \frac{1-\penJS}{\penJS} = \infty$, then a
        choice of $\penJS \approx 0$ emphasizes the
        $\log\left|\covpost^{(\ssnumdatatrain)}\right|$ term. This causes a
        preference for a small posterior variance which, in turn, creates a
        large penalization of the data-misfit term by the inverse of the
        posterior covariance.
    \item In contrast, a choice of $\penJS \approx 0$ relieves the requirement
        of a small posterior variance to promote the influence of the PoI
        data on the optimization problem.
\end{itemize}
We postulate that this counterbalancing dynamic between the two terms of (\ref{EqOptPost})
induced by $\penJS$ prevents the posterior variance from converging to too low
or too high values as the optimization procedure progresses.
A schematic of the UQ-VAE framework is displayed in Figure \ref{FigUQVAE}.

\begin{figure}[H]
    \centering
    \includegraphics[scale=0.31]{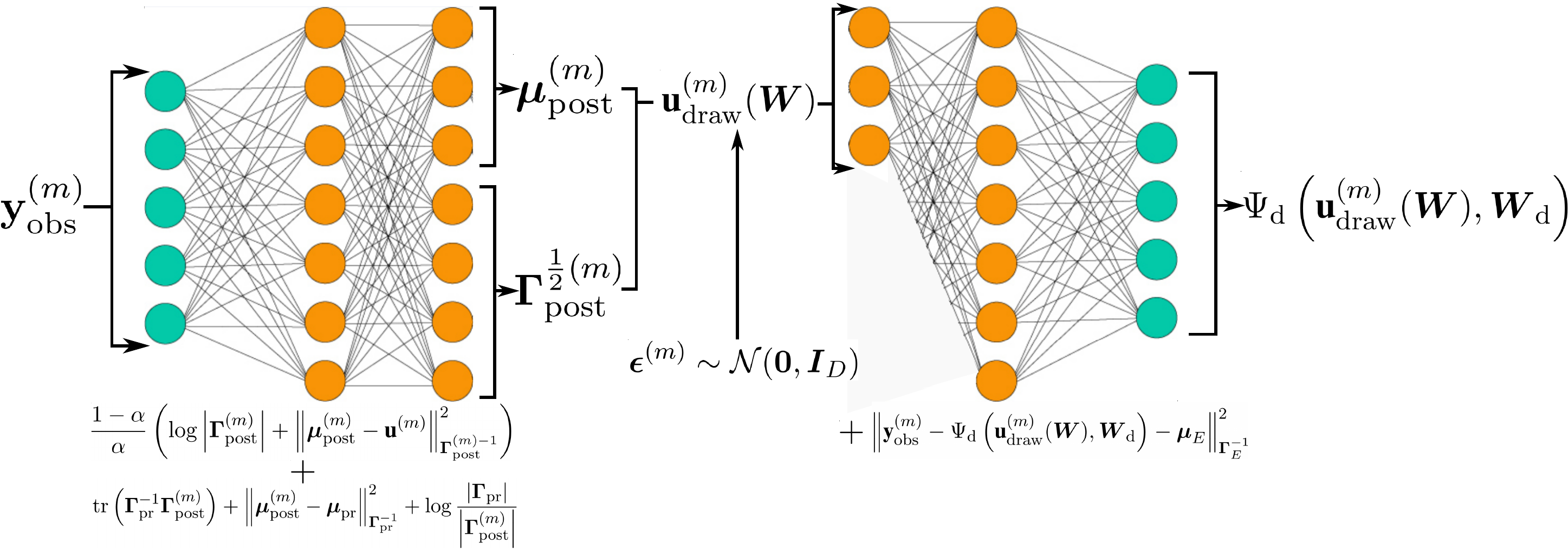}
    \caption{Schematic of the UQ-VAE framework with learned PtO map.}
    \label{FigUQVAEDecoder}
\end{figure}

During the training procedure, the repeated operation of the
PtO map on our dataset
$\left\{\left(\param^{(\ssnumdatatrain)}
,\stateobs^{(\ssnumdatatrain)}\right)\right\}_{\ssnumdatatrain=1}^{\numdatatrain}$
may incur a significant computational cost. To alleviate this, we also consider a modification
to the optimization problem (\ref{EqOpt}) where we
simultaneously learn the PtO map along with the inverse problem
solver. To this end, we replace the PtO map $\model$
with another neural network $\decoder$ parameterized by weights $\weightsde$.
With this, the term (\ref{EqOptLkhd}) becomes
\begin{align}
    \left\lVert
        \stateobs^{(\ssnumdatatrain)}
        -
        \decoder\left(\param_{\ssdraw}^{(\ssnumdatatrain)}(\weights), \weightsde\right)
        -
        \meannoise
    \right\rVert_{\regnoise}^{2}
\end{align}
and (\ref{EqOpt}) becomes an optimization problem over both sets of weights
$\left(\weights,\weightsde\right)$. Notice that the resulting optimization
problem resembles those typically used to train an autoencoder. It is for this
reason that we elect to use the subscript `d' to indicate the decoder. Since the
learned model $\decoder$ of the PtO map $\model$ is a neural network, the
evaluation of this map during the training phase is computationally inexpensive.
However, this efficiency comes at the expense of utilizing knowledge of the
governing physics of the inverse problems. Thus, the modified framework is
almost purely data-driven and resembles more the original utility of VAEs for
generative modelling.  A schematic of the UQ-VAE framework with learned PtO map
is displayed in Figure \ref{FigUQVAEDecoder}.

%------------------------------------------------------------------------------
\subsection{Analytical Result}
%------------------------------------------------------------------------------
We now present theoretical justification for the model-constrained UQ-VAE
framework for the case where the PtO map $\model$ is linear. In this case, the
target posterior $\dist(\param|\stateobs)$ is Gaussian provided that the prior
model $\gaussian(\mean_{\ssprior},\cov_{\ssprior})$ and noise model
$\gaussian(\mean_{\randnoise}, \cov_{\randnoise})$ are also Gaussian. We
show that linear, one-layer neural networks outputting the posterior mean and
Cholesky factor of the posterior covariance are able to recover exactly the
target posterior. The proof is given in Section
\ref{SecProofThmLinGaussPostAnalyticalResult} of the Appendix.
\begin{theorem} \label{ThmLinGaussPostAnalyticalResult}
    Let $\stateobs \in \mathbb{R}^{\dimobs}$.
    Consider a Gaussian prior model
    $\gaussian\left(\mean_{\ssprior}, \cov_{\ssprior}\right)$ and Gaussian noise
    model $\gaussian\left(\mean_{\randnoise},\cov_{\randnoise}\right)$.
    Suppose the target posterior $\dist(\param|\stateobs) =
    \gaussian\left(\mean_{\sstrue},\cov_{\sstrue}\right)$ is such that
    \begin{subequations}
        \begin{align}
            \cov_{\sstrue}
            &=
            \left(
                \modellinear^\mathrm{T}\regnoise\modellinear
                +
                \regprior
            \right)^{-1}\\
            \mean_{\sstrue}
            &=
            \cov_{\sstrue}
            \left(
                \modellinear^\mathrm{T}\regnoise
                \left(
                    \stateobs
                    -
                    \mean_{\randnoise}
                \right)
                +
                \regprior\mean_{\ssprior}
            \right)
        \end{align}
    \end{subequations}
    where $\modellinear \in \mathbb{R}^{\dimobs \times \dimparam}$.
    Suppose the model posterior $\distmodelpost_{\paraminfer}(\param|\stateobs)=
    \gaussian\left(\meanpost,\covpost\right)$ is such that
    \begin{subequations}
        \begin{align}
            \meanpost
            &=
            \weights_{\mean}\stateobs + \bias_{\mean}\\
            \covpost^{\frac{1}{2}}
            &=
            \lowtri
            \odot
            \lowtrione
            +
            \mathrm{diag}\left(\vecstd\right)\\
            \log\left(\vecstd\right)
            &=
            \weights_{\vecstd}\stateobs + \bias_{\vecstd}\\
            \rmvec\left(\lowtri\right)
            &=
            \weights_{\lowtri}\stateobs + \bias_{\lowtri}
            \label{EqLowTri}
        \end{align}
    \end{subequations}
    where $\lowtrione$ is a lower triangular matrix of ones with zeros on the
    diagonal. Let $\penJS = \frac{1}{2}$. Then the optimization problem
    \begin{subequations}
        \begin{align}
            \min_{
                \weights_{\mean}, \bias_{\mean},
                \weights_{\vecstd}, \bias_{\vecstd},
                \weights_{\lowtri}, \bias_{\lowtri}
            }
            &\frac{1-\penJS}{\penJS}
            \left(
                \log\left|\covpost\right|
                +
                \mathrm{tr}\left(
                    \covpost^{-1}\cov_{\sstrue}
                \right)
                +
                \left\|
                    \meanpost
                    -
                    \mean_{\sstrue}
                \right\|_{\covpost^{-1}}^{2}
            \right)\\
            &+
            \mathrm{tr}\left(
                    \regnoise
                    \modellinear
                    \covpost
                    \modellinear^\mathrm{T}
                \right)
            +
            \left\lVert
                \stateobs
                -
                \modellinear
                \meanpost
                -
                \meannoise
            \right\rVert_{\regnoise}^{2}\\
            &+
            \mathrm{tr}\left(\covprior^{-1} \covpost\right)
            +
            \left\|
                \meanpost
                -
                \meanprior
            \right\|_{\regprior}^{2}
            +
            \log\frac{\left|\covprior\right|}
            {\left|\covpost\right|}
        \end{align}
        \label{EqOptLinExpctThm}
    \end{subequations}
    achieves its minimum if and only if
    $\weights_{\mean}, \bias_{\mean},
    \weights_{\vecstd}, \bias_{\vecstd},
    \weights_{\lowtri}, \bias_{\lowtri}$
    are such that $\meanpost = \mean_{\sstrue}$ and $\covpost = \cov_{\sstrue}$.
\end{theorem}
Although this result provides theoretical guarantees, the requirement of
learning a full matrix in (\ref{EqLowTri}) can be computational infeasible. For
this reason, it is often the case that only the diagonal of the model posterior
covariance is learned \cite{kingma2013auto}.

%==============================================================================
\section{Related Work}
%==============================================================================
In generative modelling, it is beneficial to learn better representations of the
latent space. To this end, the use of the JSD for VAEs is explored in
\cite{deasy2020constraining, sutter2020multimodal}. These works consider using
the JSD in place of the KLD in the ELBO between the posterior and the prior. Our
work differs from these approaches as the JSD is applied not directly to the
ELBO.  Rather, the JSD is applied at the inference level to yield a different
bound that includes an expression of the latent data.

In scientific inverse problems, neural networks have been used to augment the
solution process in \cite{adler2017solving, jin2017deep, li2018nett,
patel2019bayesian, peng2020solving, seo2019learning,
sheriffdeen2019accelerating}. Some of these works use deep learning to assist a
more traditional solution algorithm. In \cite{adler2017solving}, a convolutional
neural network is used in a partially learned gradient descent scheme. In
\cite{jin2017deep}, the component of the iterative solution algorithm requiring
convolutional operators is surrogated with convolutional neural networks. Other
approaches use deep learning directly for regularization of an ill-posed inverse
problem. In \cite{li2018nett}, neural networks are used in place of a Tikhonov
regularizer. The work in \cite{patel2019bayesian, gonzalez2019solving}, explored
prior modelling in Bayesian inverse problems using generative adversarial
networks (GANs) and VAEs respectively. In \cite{seo2019learning}, VAEs are used
to learn a low dimensional manifold representation of the PoI which introduces
regularization for ill-posed inverse problems. Similarly, in
\cite{peng2020solving}, an autoencoder is used to learn a latent representation
of the unknown of interest which thereby allows the inversion task to be
reframed into solving for the latent representation. In
\cite{sheriffdeen2019accelerating}, bounded neural network discrepancy models
were used to augment reduce order models in order to accelerate PDE-constrained
inverse solutions.  In summary, even with improvements through deep learning,
the computational cost of these approaches is still mostly inherited from the
cost of the base solution algorithms that were augmented. In our approach, the
solution process is itself propagation through a trained neural network; a
process largely unrivalled in terms of computational efficiency. Additionally,
most of the work mentioned above do not quantify uncertainty.

Broadly speaking, there are three types of approaches for performing UQ with
deep learning. In an analogous manner to above, the first type utilizes deep
learning to augment traditional methods for UQ. In
\cite{jiang2019combining}, an autoencoder is trained to reduce the
dimensionality of the PoI space. MCMC sampling is then conducted on the
low-dimensional representation and estimation in the full PoI space is recovered
by propagation through the trained decoder. Similarly, in \cite{hou2019solving},
a class of network training methods was proposed that can be combined with
sample-based Bayesian inference algorithms. For the second type of approach,
deep learning models that inherently possess some stochastic property are,
themselves, used to represent uncertainty. In \cite{caldeira2020deeply}, an
investigation into Bayesian Neural Networks, Concrete Dropout and Deep Ensembles
was provided in the setting of scientific problems. However, the discussion in
the paper does not directly pertain to inverse problems.

Our proposed method for UQ falls under the third type of
approach. Here, deep learning is used to directly learn and model the
uncertainty. In \cite{gabbard2019bayesian, tonolini2020variational,
chua2020learning, adler2018deep}, purely deep learning posterior modelling
approaches were proposed. In \cite{gabbard2019bayesian} and
\cite{tonolini2020variational}, conditional VAEs (CVAE) \cite{sohn2015learning}
were trained to perform variational inference for gravitational-wave astronomy
problems and computational imaging problems respectively. Unlike our UQ-VAE
approach, CVAEs do not marry the latent space with the PoI space; they require a
separate latent variable for its derivation. The work in \cite{chua2020learning}
trains a neural network to rapidly produce one or two dimensional projections of
Bayesian posteriors for gravitational-wave astronomy.  For more general and
higher-dimensional settings, the work in \cite{adler2018deep} introduces Deep
Posterior Sampling which trains a Wasserstein GAN to sample from the posterior.
However, the training of GANs is notoriously unstable; a limitation also met and
acknowledged by the authors of \cite{adler2018deep} in their paper.  Compared to
GANs, VAEs possess a mathematical rigour that is more directly related to
the data distribution it attempts to generate from. Therefore, with the VAE
at the core of our proposed framework, we postulate that there is more
potential for mathematically rigourous extension.

%==============================================================================
\section{Results}
%==============================================================================
For our numerical experiments, we consider the two dimensional steady state heat
conduction problem. The details of this problem setup can be found in Section
\ref{Sec2DHeat} of the Appendix. Although the non-linear forward mapping is
quite simple, elliptic inverse problems are severely ill-posed and serve as
excellent benchmark inference problems
\cite{kabanikhin2008definitions,kirsch2011introduction,beskos2017geometric,cui2016dimension,chen2019projected}.
Details of the neural network architectures and optimization procedure can be
found in Section \ref{SecTrainingProperties} of the Appendix. The codes
implementing the UQ-VAE framework used to produce these results can be obtained
from \cite{goh2020uqvae}. For comparison, we use the hIPPYlib library
\cite{VillaPetraGhattas19} to compute the Laplace approximation using gradients
and Hessians derived via the adjoint method, inexact Newton-CG, Armijo line
search and randomized eigensolvers.

For our numerical experiments, we consider the following cases:
\begin{enumerate}
    \item four noise levels $\noiselevel = 0, 0.01, 0.05, 0.1$
    \item four sizes of training datasets $\numdatatrain = 50, 500, 1000, 5000$
    \item four choices of the JSD family $\penJS = 0.00001, 0.001, 0.1, 0.5$
    \item exact PtO map and learned PtO map.
\end{enumerate}
For our numerical simulations, in some cases, selecting $\penJS \geq 0.5$ may
yield an unstable training procedure with exploding gradients and so we only
display results with $\penJS$ as high as $\penJS = 0.5$. The computational cost
of training is detailed in Section \ref{SecComputationalCost} of the Appendix.
The table of relative errors and figures are found in Section
\ref{SecResults2DHeat} of the Appendix with each figure set and subsection
representing a noise level and dataset size.  Note that we have elected to only
display the cross-sectional predictions and not the predictions over the full
domain since it is harder to visualize the uncertainty in the latter.

Before continuing to a detailed discussion, we address the comment in Section
\ref{SecRegOptProblem} regarding the adoption of a crude Monte-Carlo
approximation using our PoI data. The overall feasibility of our obtained
uncertainty estimates indicate that it is sufficient to use only one PoI
datapoint. We postulate that any inaccuracies resulting from such a crude
approximation is alleviated by the variety of information embedded in the
training dataset. Indeed, since the weights are used to capture the statistics
of a whole training dataset, the amortized nature of our inference method
perhaps reduces the requirement for accurate Monte-Carlo approximations.
Moreover, in the original VAE paper \cite{kingma2013auto}, the authors observed
that, providing the number of training data $\numdatatrain$ was large enough,
using only one draw from the model posterior to approximate the expected value
of the likelihood term was sufficient to obtain good results. This finding
regarding the approximation of the likelihood term mirrors ours regarding the
posterior data misfit term.

%------------------------------------------------------------------------------
\subsection{Comparison with the Laplace Approximation} \label{SecCompareLaplace}
%------------------------------------------------------------------------------
We summarize our results beginning with a comparison between UQ
using UQ-VAE and with the Laplace approximation. It is important
to point out that UQ-VAE provides a global Gaussian approximation while the
Laplace approximation is usually a local Gaussian approximation at the MAP
point.  From the results, we can see that the posterior mean estimated by the
UQ-VAE is closer to the true value than the MAP estimate. However, the Laplace
approximation provides larger uncertainty estimates which ensures they are still
feasible despite the underestimation of values by the MAP. In contrast,
especially for larger values of $\penJS$, the UQ-VAE uncertainty estimates are
smaller and, in some cases, the true parameter values at the location of the
anomaly lie slightly outside the uncertainty estimates.

Recall that the statistics of the noise are accurately modelled.
Therefore, the larger the value of $\noiselevel$, the more diminished the
influence of the likelihood model (\ref{EqOptLkhd}) on the training procedure;
through penalization by the weighting matrix $\regnoise$. For the Laplace
approximation (\ref{EqLaplaceApprox}), this translates to an increase in the
approximate posterior variance. The approximate posterior variance obtained
from the UQ-VAE also exhibits this response to the increased noise
regularization.

Further, the Laplace approximation uncertainty estimates are clearly responsive
to the sensor locations; providing smaller uncertainty where the sensors are
located. Whilst the UQ-VAE uncertainty estimates do exhibit a similar behaviour,
it is less pronounced due to the aforementioned global and
local nature of the UQ-VAE and Laplace approximation, respectively.
Furthermore, we postulate that this is because the UQ-VAE uncertainty estimates
represent two sources of information; the observational data fed into the
encoder as well as the PoI and observational pairs in the training dataset.
Whilst the former provides information only at the locations at the sensors, the
latter provides information at all areas of the domain. Indeed, the process of
training under the UQ-VAE framework works to embed information about the
training dataset into the neural network. UQ is then
performed with this embedded information as well as the input observational
data. Despite these differences, we believe that it is still instructive to
compare the two approximations.

Finally, the computational efficiency of
inference by propagation through a trained neural network $\NN$ is, on average, more
than 2750 times faster than that of the Laplace approximation.

\begin{figure}[H]
    \centering
    \includegraphics[scale=\figscale]{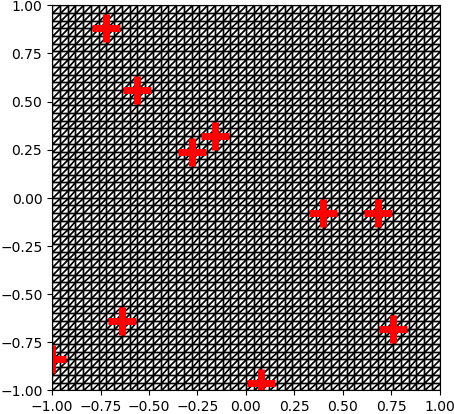}
    \includegraphics[scale=\figscale]{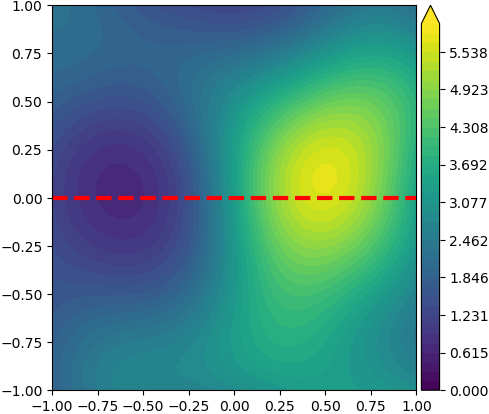}
    \includegraphics[scale=\figscale]{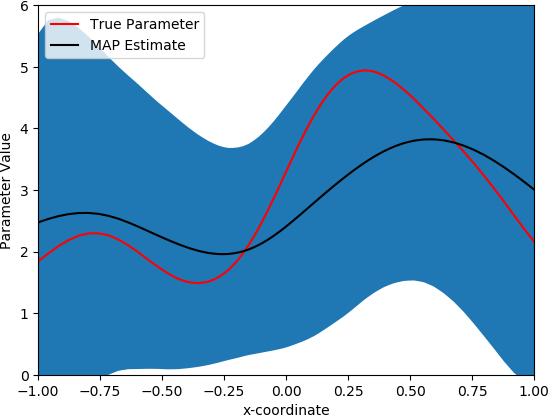}
    \includegraphics[scale=\figscale]{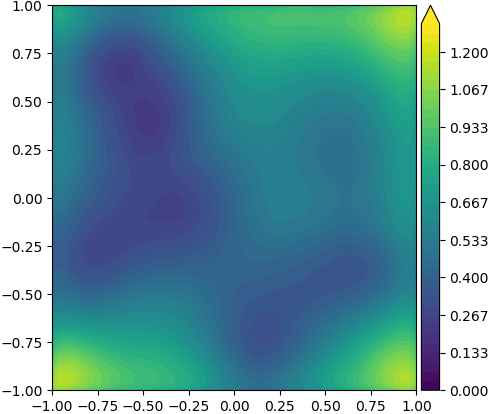}\\
    \includegraphics[scale=\figscale]{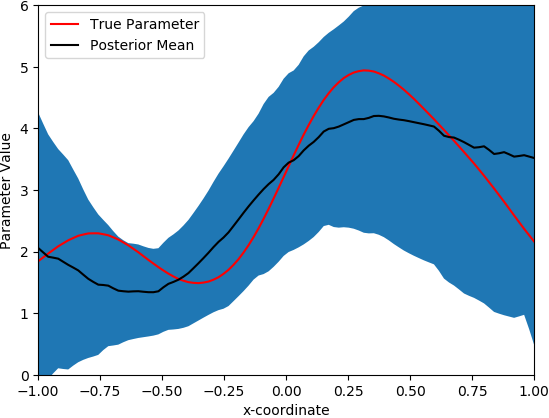}
    \includegraphics[scale=\figscale]{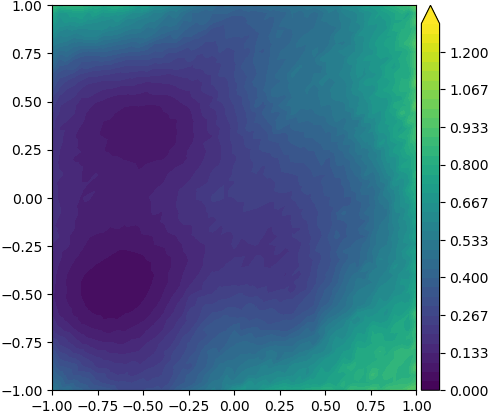}
    \includegraphics[scale=\figscale]{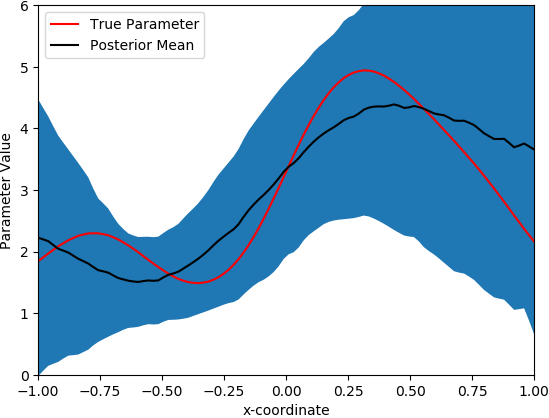}
    \includegraphics[scale=\figscale]{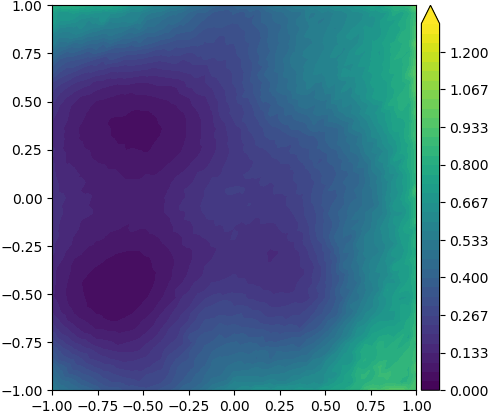}\\
    \caption{\scriptsize Top row left to right: mesh with sensors denoted with a red cross,
        true PoI,
        cross-sectional uncertainty estimate and pointwise posterior
        variance from Laplace approximation.
        Second row: UQ-VAE uncertainty estimates with $\penJS = 0.001$. First
        and third columns: cross-sectional uncertainty estimates. Second and
        fourth columns: approximate pointwise posterior variance. First and
        second columns: exact PtO map.  Third and fourth columns: learned PtO
        map.}
        \label{FigPoisson2Dn2601nspt01M1000Example}
\end{figure}

See Section \ref{SecComputationalCost} of the Appendix for more details on the
computational cost of inference.  An example of our computational results is
displayed in Figure \ref{FigPoisson2Dn2601nspt01M1000Example} which compares the
uncertainty estimates obtained using the Laplace approximation with those
obtained with our UQ-VAE framework.

%------------------------------------------------------------------------------
\subsection{Comparison Within the UQ-VAE Framework}
%------------------------------------------------------------------------------
Now, we discuss the results within the UQ-VAE framework by comparing between
noise levels, dataset size, modelling and learning of the PtO map and choices of
$\penJS$:
\begin{enumerate}
    \item Comparing between noise levels, feasible estimates were obtained for
        all noise levels. However, for $\noiselevel = 0.05$, feasible estimates
        were only obtained for $\numdatatrain = 50, 500, 1000$. With the exception
        of $\noiselevel = 0.05$ and $\numdatatrain = 5000$, the uncertainty
        estimates are larger for larger values of $\noiselevel$. For the noise
        level of $\noiselevel = 0.1$, no feasible estimates were obtained for
        any choice of dataset size $\numdatatrain$.
    \item Comparing between training dataset sizes, the uncertainty estimates
        are larger for smaller dataset sizes.  When $\noiselevel = 0.05$, larger
        dataset sizes appear to be detrimental to the accuracy of the
        estimation.
    \item Comparing between the case where the PtO is exact and when the PtO
        is learned, for the cases of $\noiselevel = 0$, $\numdatatrain = 500$,
        $\penJS = 0.5$ and $\noiselevel = 0.01$, $\numdatatrain = 500$, $\penJS
        = 0.1$, using the exact PtO clearly outperforms using the learned
        PtO.
    \item Comparing between choices of $\penJS$, for smaller values of $\penJS$
        the estimated posterior mean underestimates the true PoI values and so a
        larger uncertainty estimate is required to ensure feasible estimates.
        The reverse occurs for larger values of $\penJS$.
\end{enumerate}
We now offer some key interpretations our above observations.
For the first observation, the reasoning for larger uncertainty estimates
observed with larger values of $\noiselevel$ was discussed in Section
\ref{SecCompareLaplace}.

For the second observation, we see that the posterior uncertainty is responsive
to the size of the dataset. With less data, our uncertainty estimates are
larger. This behaviour is ideal as more information intuitively implies less
uncertainty. For the case of $\noiselevel = 0.05$, we conjecture that the use of
large datasets of heavily corrupted data yields an unfavourable influence on our
optimization procedure.  We believe that the limitation of our framework lies in
its potential dependence on an accurate training dataset. One of the key merits
of our proposed method is that we have introduced, in a mathematically justified
manner, a data-driven component to the inversion process. This expands the
toolkit at one's disposal when solving inverse problems. Indeed, in addition to
constructing models for the prior and the PtO map, information on the properties
of the PoI as well as the governing physics can be encoded into the training
dataset. However, it would be incomplete to view this added option purely as an
advantage. Although the utilization of datasets alleviates the burden on
accurate prior and physics modelling, poorly constructed or highly corrupted
datasets could completely sabotage the inversion process regardless of any
accuracy achieved by the prior and physics models.

For the third observation, it can be reasoned that the similarity of results for
the two cases is due to the success of the decoder in learning the PtO. The
elliptic forward problem is quite simple and so it is not unexpected that the
decoder learns it with ease. Thus, simultaneously learning an accurate PtO may
be a reasonable task that aids and does not detract from the main task of
learning the Bayesian inverse problems solver. This supports the strategy of
using a learned PtO to reduce the offline cost of training the neural network.
However, the two cases where use of the exact PtO outperforms using the
learned PtO with the small training dataset size of $\numdatatrain=500$ suggests
that including the physics of the problem allows for feasible uncertainty
estimates when large amounts of training data may not be readily available.

For the fourth observation, in alignment with our discussion in Section
\ref{SecRegOptProblem}, a lower value of $\penJS$ creates a larger penalty on
the posterior terms (\ref{EqOptPost}). Since our results suggest that a lower
value of $\penJS$ yields a larger posterior variance, then it is implied that
the minimization of $\log\left|\covpost^{(\ssnumdatatrain)}\right|$ is not a
priority during the training procedure; so long as the posterior data
$\param^{(\ssnumdatatrain)}$ has enough influence. One can also reason that
since lower values of $\penJS$ correspond to the zero-forcing KLD, one would
expect a larger model variance; for example in the case of approximating a
multimodal distribution with a unimodal model \cite{bishop2006pattern}. The
reverse occurs with higher values of $\penJS$ which, due to tendency towards the
zero-avoiding KLD, would result in a smaller model variance.

Finally, we address the instability of the training procedure as $\penJS \to 1$
which diminishes the influence of the PoI data on our optimization problem and
moves towards recovering the zero-avoiding KLD. Recall the more traditional
method of optimizing the exponentiated functional in (\ref{EqBayesPost}) where
the target optimization parameter is the physically meaningful $\param$. In
contrast, our method optimizes for the more arbitrary and often more plentiful
weights $\weights$. Intuitively, we are therefore faced with a more ill-posed
problem.  Moreover, from a perspective centered around the traditional posterior
(\ref{EqBayesPost}), one can instead view the posterior data terms
(\ref{EqOptPost}) as regularizers of the traditional loss functional involving
only the likelihood and prior term. This provides the much needed regularization
for our more ill-posed inverse problem; which therefore necessitates that
$\penJS \ll 1$ for this regularization to take effect. We leave concrete
theoretical analysis for future work.

%==============================================================================
\section{Conclusion}
%==============================================================================
In this paper, we propose a framework for training of neural networks capable of
rapid UQ. Although the VAE, on which this framework is
based, was originally motivated by methods in generative modelling, we have
shown that it is also well suited for scientific inverse problems. Indeed, it
utilizes three sources of information usually available for such problems:
\begin{enumerate}
    \item the physical laws governing the problem through the PtO map
    \item the physical properties of the PoI through the prior model
    \item simulation or experimentation procedures through the paired
        PoI-observation datasets.
\end{enumerate}
Furthermore, this framework is derived from a solid mathematical foundation and
possesses a complex, dynamic interplay of factors from variational inference as
well as regularization. Despite this complexity, this framework requires only a
few design decisions. Indeed, aside from the usual adjustment of hyperparameters
associated with neural network architecture, essentially only the tuning of the
hyperparameter $\penJS$ is required. The selection of the PtO map and prior model
can be guided by the underlying physical properties of the problem. Our results
also show that feasible estimates are achievable and, moreover, that these estimates
exhibit behaviour similar to that of existing inversion methods. Additionally, our
uncertainty estimates show a favourable response to the size of our training
dataset; the uncertainty is inversely proportional to the amount of training
data used which reflects the availability or lack of information. The preliminary
investigation offered in this paper employs only standard optimization
techniques and simple deep learning architectures with a very minimal amount of
tuning performed. This decision was made in order to showcase the robustness as
well as the limitations of our framework in its most basic form. Moreover, we
utilize somewhat crude statistical approximations. We believe that the results
presented in this paper can be improved with the employment of more sophisticated
optimization procedures, neural network architecture and statistical machinery.

% Acknowledgments---Will not appear in anonymized version
\acks{
    This research was partially funded by the National Science Foundation awards
    NSF-1808576 and NSF-CAREER-1845799; by the Defense Thread Reduction Agency award
    DTRA-M1802962; by the Department of Energy award DE-SC0018147; by KAUST; by 2018
    ConTex award; and by 2018 UT-Portugal CoLab award. The authors acknowledge the
    Texas Advanced Computing Center (TACC) at The University of Texas at Austin for
    providing HPC resources that have contributed to the research results reported
    within this paper. URL: \url{http://www.tacc.utexas.edu}. The authors would like
    to thank Jari Kaipio, Ruanui Nicholson and Rory Wittmer for the insightful
    discussions.
}

\bibliography{references}

\appendix

%==============================================================================
\section{Proofs}
%==============================================================================
%------------------------------------------------------------------------------
\subsection{Proof of Theorem \ref{ThmGenJSVariationalInference}} \label{SecProofThmGenJS}
%------------------------------------------------------------------------------
\begin{proof}
The clearest path forward is to manipulate
\begin{subequations}
    \begin{align}
        \JS_\penJS(\modelpost||\truepost)
        =
        &\penJS \KL(\modelpost||(1-\penJS)\modelpost + \penJS \truepost) \\
        &+ (1 - \penJS) \KL(\truepost||(1-\penJS)\modelpost + \penJS \truepost)
    \end{align}
\end{subequations}
term-by-term. Beginning with the first term,
\begin{subequations}
        \begin{align}
            &\penJS \KL(\modelpost||(1-\penJS)\modelpost + \penJS \truepost)\\
            &=
            \penJS \expct_{\paramdistmodel}\LRs{
            \log \LRp{\frac{\modelpost}{(1-\penJS)\modelpost + \penJS \truepost}}}\\
            &=
            -\penJS\expct_{\paramdistmodel}\LRs{
            \log \LRp{\frac{(1-\penJS)\modelpost + \penJS \truepost}{\modelpost}}}\\
            &= -\penJS \expct_{\paramdistmodel}\LRs{
            \log \LRp{\frac{\truepost}{\modelpost}\LRp{
            \penJS +  \frac{(1-\penJS)\modelpost}{\truepost}}}}\\
            &= -\penJS \expct_{\paramdistmodel} \LRs{
            \log \LRp{\frac{\truepost}{\modelpost}}}
            -\penJS \expct_{\paramdistmodel}\LRs{
            \log\LRp{\penJS +  \frac{(1-\penJS)\modelpost}{\truepost}}}\\
            &= -\penJS \expct_{\paramdistmodel} \LRs{
            \log \LRp{\frac{\truejoint }{\diststate \modelpost}}}
            -\penJS \expct_{\paramdistmodel} \LRs{
            \log \LRp{ \penJS +  \frac{(1-\penJS)\modelpost}{\truepost}}}\\
            &= \penJS\log\LRp{\diststate}
            -\penJS \expct_{\paramdistmodel} \LRs{
            \log \LRp{\frac{\truejoint}{\modelpost}}}
            -\penJS \expct_{\paramdistmodel} \LRs{
            \log \LRp{ \penJS +  \frac{(1-\penJS)\modelpost}{\truepost}}}.
        \end{align} \label{EqJSq}
\end{subequations}
Similarly, the second term can be decomposed as
\begin{subequations}
    \begin{align}
        &(1-\penJS)\KL(\truepost||(1-\penJS)\modelpost + \penJS \truepost)\\
        &=
        (1-\penJS)\expct_{\paramdisttrue}\LRs{
        \log \LRp{\frac{\truepost}{(1-\penJS)\modelpost + \penJS \truepost}}}\\
        &= -(1-\penJS) \expct_{\paramdisttrue} \LRs{
        \log \LRp{\frac{(1-\penJS)\modelpost + \penJS \truepost}{\truepost}}} \\
        &= -(1-\penJS) \expct_{\paramdisttrue} \LRs{
        \log \LRp{\penJS + \frac{(1-\penJS)\modelpost}{\truepost}}}.
    \end{align} \label{EqJSp}
\end{subequations}
Combining equations (\ref{EqJSq}) and (\ref{EqJSp}), we arrive at
\begin{subequations}
    \begin{align}
        \JS_{\penJS}(\modelpost||\truepost)
        =&\penJS \log \LRp{\diststate} \\
        &-\penJS \expct_{\paramdistmodel}\LRs{
        \log \LRp{\frac{ \truejoint}{\modelpost}}}\\
        &-\penJS \expct_{\paramdistmodel} \LRs{
        \log \LRp{\penJS +  \frac{(1-\penJS)\modelpost}{\truepost}}}\\
        &-(1 - \penJS) \expct_{\paramdisttrue} \LRs{
        \log \LRp{\penJS + \frac{(1-\penJS)\modelpost}{\truepost}}}.
    \end{align}
\end{subequations}
Finally, the expressions in Theorem \ref{EqGenJSlowerbound} follow through
division by $\penJS$ and rearrangement.
\end{proof}

%------------------------------------------------------------------------------
\subsection{Proof of Theorem \ref{ThmLinGaussPostAnalyticalResult}}
\label{SecProofThmLinGaussPostAnalyticalResult}
%------------------------------------------------------------------------------
\begin{proof}
    Set $\penJS = \frac{1}{2}$. Forcing the variations with respect to weights
    $\weights_{\mean}$ and biases $\bias_{\mean}$ to vanish gives
    \begin{subequations}
        \begin{align}
            &\covpost^{-1}
            \left(
                \mean_{\sspost}
                -
                \mean_{\sstrue}
            \right)\\
            &-
            \modellinear^\mathrm{T}
            \regnoise
            \left(
                \stateobs
                -
                \modellinear
                \mean_{\sspost}
                -
                \mean_{\randnoise}
            \right)
            +
            \regprior
            \left(
                \mean_{\sspost}
                -
                \mean_{\ssprior}
            \right)\\
            &=
            \textbf{0}
        \end{align}
    \end{subequations}
    which is true if and only if $\mean_{\sspost} = \mean_{\sstrue}$.
    Forcing the variations with respect to $\weights_{\lowtri}$ and biases
    $\bias_{\lowtri}$ to vanish gives
    \begin{subequations}
        \begin{align}
            &-
            \left(
                \covpost^{-\frac{1}{2}}
                \cov_{\sstrue}
                \otimes
                \covpost^{-\frac{1}{2}\mathrm{T}}
                +
                \covpost^{-\frac{1}{2}}
                \left(
                    \meanpost - \mean_{\sstrue}
                \right)
                \left(
                    \meanpost - \mean_{\sstrue}
                \right)^\mathrm{T}
                \otimes
                \covpost^{-\frac{1}{2}\mathrm{T}}
            \right)
            \rmvec\left(
                \covpost^{-\frac{1}{2}}
            \right)\\
            &+
            \left(
                \mateye_{\dimparam}
                \otimes
                \modellinear^\mathrm{T}\regnoise\modellinear
            \right)
            \rmvec\left(
                \covpost^{\frac{1}{2}}
            \right)
            +
            \left(
                \mateye_{\dimparam}
                \otimes
                \covprior^{-1}
            \right)
            \rmvec\left(
                \covpost^{\frac{1}{2}}
            \right)\\
            &=
            \textbf{0}
        \end{align}
    \end{subequations}
    where $\otimes$ denotes the Kronecker product. Using the two identities
    \cite{dhrymes1978mathematics}
    \begin{subequations}
        \begin{align}
            \rmvec\left(
                \boldsymbol{A}
                \boldsymbol{B}
                \boldsymbol{C}
                \boldsymbol{D}
            \right)
            &=
            \left(
                \boldsymbol{D}^\mathrm{T}
                \boldsymbol{C}^\mathrm{T}
                \otimes
                \boldsymbol{A}
            \right)
            \rmvec\left(
                \boldsymbol{B}
            \right)\\
            \rmvec\left(
                \boldsymbol{A}
                \boldsymbol{B}
            \right)
            &=
            \left(
                \mateye
                \otimes
                \boldsymbol{A}
            \right)
            \rmvec\left(
                \boldsymbol{B}
            \right)
        \end{align}
    \end{subequations}
    along with some basic manipulation gives
    \begin{align}
        &-\covpost^{-1}
        \left(
            \cov_{\sstrue}
            +
            \left(
                \meanpost - \mean_{\sstrue}
            \right)
            \left(
                \meanpost - \mean_{\sstrue}
            \right)^\mathrm{T}
        \right)
        +
        \left(
            \modellinear^\mathrm{T}\regnoise\modellinear
            +
            \covprior^{-1}
        \right)
        \covpost
        =
        \textbf{0}.
        \label{EqForceVariation0lowtri}
    \end{align}
    Forcing the variations with respect to $\weights_{\vecstd}$ and biases
    $\bias_{\vecstd}$ to vanish gives
    \begin{subequations}
        \begin{align}
            &-\covpost^{-1}
            \odot
            \left(
                \cov_{\sstrue}
                +
                \left(
                    \meanpost - \mean_{\sstrue}
                \right)
                \left(
                    \meanpost - \mean_{\sstrue}
                \right)^\mathrm{T}
            \right)
            \covpost^{-\frac{1}{2}\mathrm{T}}\\
            &+
            \left(
                \modellinear^\mathrm{T}\regnoise\modellinear
                \odot
                \covpost^{\frac{1}{2}}
            \right)
            +
            \left(
                \covprior^{-1}
                \odot
                \covpost^{\frac{1}{2}}
            \right)\\
            &=
            \textbf{0}
        \end{align}
        \label{EqForceVariation0vecstd}
    \end{subequations}
    where $\odot$ denotes the entrywise product. Here,
    (\ref{EqForceVariation0lowtri}) and (\ref{EqForceVariation0vecstd})
    with $\meanpost = \mean_{\sstrue}$
    are true if and only if $\covpost = \cov_{\sstrue}$ as required.
\end{proof}

%==============================================================================
\section{Two Dimensional Steady State Heat Conduction Problem} \label{Sec2DHeat}
%==============================================================================
We considered the heat equation with heat conductivity as the PoI and
temperature as the state. The governing PDE and associated boundary conditions
are displayed below
\begin{subequations}
    \begin{alignat}{3}
        -\nabla\cdot \functparam \nabla \functstate &= 0
        &&\quad\text{ in }\domain \\
        -\functparam(\nabla \functstate \cdot \vecnormal) &= \mathrm{Bi}\, \functstate
        &&\quad\text{ on }\domain^{\mathrm{ext}} \setminus
        \domain^{\mathrm{root}} \\
        -\functparam(\nabla \functstate \cdot \vecnormal) &= -1
        &&\quad\text{ on }\domain^{\mathrm{root}}
        \label{EqHeatEquation}
    \end{alignat}
\end{subequations}
where $\functparam$ denotes the thermal heat conductivity, $\mathrm{Bi}$ is the
Biot number set to $\mathrm{Bi}=0.5$, $\domain$ is the physical domain,
$\domain^{\mathrm{root}}$ is the bottom edge of the domain and
$\domain^{\mathrm{ext}}$ is the remaining edges of the domain.

The prior model has mean $\meanprior = 2$ and covariance $\covprior$ that is a
discretization of the infinite dimensional covariance operator
$\covinfprior = \opinfpde^{-2}$ where $\opinfpde$ is a differential operator
such that
\begin{align}
    \opinfpde \functparam
    =
    \begin{cases}
        -\blpcoefftwo\oppoisson \functparam + \blpcoeffone \functparam  &\text{ in }\domain\\
        \nabla \functparam \cdot \vecnormal + \blpcoeffbnd\functparam
                                                            &\text{ on }\partial\domain.
    \end{cases}
\end{align}
Here, $\blpcoeffone, \blpcoefftwo > 0$ controls the correlation length and
variance of the prior operator. We set $\blpcoefftwo = 0.1$, $\blpcoeffone =
0.5$ and $\blpcoeffbnd$ is chosen as in \cite{daon2016mitigating,
roininen2014whittle} to reduce boundary artifacts. Priors of this type ensure
that $\covinfprior$ is a trace-class operator which guarantees bounded pointwise
variance and a well-posed infinite-dimensional Bayesian inverse problem
\cite{stuart2010inverse,bui2013computational}.

The testing dataset of PoI values is drawn from a Gaussian autocorrelation
smoothness prior \cite{kaipio2006statistical} with mean $\mean = 2$ and
covariance
\begin{align}
    \cov_{ij}
    =
    \std^{2}\exp\left(-\frac{
            \left\lVert \vardata_{i} - \vardata_{j}\right\rVert_{2}^{2}}
            {2\corr^{2}}\right)
\end{align}
with $\std^{2} = 2$, $\corr = 0.5$. The training set for the UQ-VAE network is a
set of separate draws from the same distribution. From this, the corresponding
state observations are computed using the solvers from the FEniCS library
\cite{logg2012automated} on the PoI dataset. The same solver is used as our
numerical model of the PtO map during the training procedure.

We consider a computational domain consisting of $\dimparam = 2601$ degrees of
freedom. The observation data corresponds to sensor measurements from $10$
randomly selected locations and is afflicted with Gaussian
distributed additive noise on all sensors with zero mean and standard deviation
$\std = \noiselevel \max\left|\stateobs\right|$. For our noise model, we set
$\left(\meannoise,\covnoise\right)
=
\left(\textbf{0},\std^{2} \mateye_{\dimobs}\right)$ which
corresponds to the scenario where the statistics of the noise are accurately
modelled. Finally, we assume a diagonal matrix for our posterior model
covariance.

%==============================================================================
\section{Neural Network Architecture and Training Properties} \label{SecTrainingProperties}
%==============================================================================
%------------------------------------------------------------------------------
\subsection{Architecture}
%------------------------------------------------------------------------------
The architecture of our neural network $\NN$ consists of $5$ hidden layers of
$500$ nodes with the ReLU activation function. No activation function was used
at the output layer. The input layer has $\dimobs$ number of nodes to match the
dimension of the observational data which comprises of $\dimobs$ number of
measurement points.  The output layer has $2\dimparam$ nodes, with $\dimparam$
the dimension of the PoI, to represent the estimated posterior mean
$\meanpost$ and diagonal of the posterior covariance $\covpost$.

When the parameter-to-observable map is learnt, the corresponding decoder
network $\decoder$ has $2$ hidden layers also with $500$ nodes and the ReLU
activation function. Again, no activation function was used at the output layer.
The input layer has $\dimparam$ number of nodes to represent a draw from the
learned posterior and the output layer has $\dimobs$ number of nodes to match
the dimension of the observational data.

%------------------------------------------------------------------------------
\subsection{Training}
%------------------------------------------------------------------------------
For optimization, we use a batch size of $100$.  Therefore, the loss in
(\ref{EqOpt}) is averaged over the number of PoI and observation pairs and the
gradient is calculated for each batch. Optimization is conducted using
the Adam optimizer which performs mini-batch momentum-based stochastic gradient
descent \cite{kingma2014adam}. This training procedure was repeated for 400
epochs.

The metric used to measure accuracy is the averaged relative error
$\frac{1}{\numdatatest}\sum_{\ssnumdatatest=1}^{\numdatatest}
 \frac{\left\lVert \param^{(\ssnumdatatest)}
        - \meanpost\left(\stateobs^{(\ssnumdatatest)}\right)
        \right\rVert_{2}^{2}}
        {\left\lVert \param^{(\ssnumdatatest)}\right\rVert_{2}^{2}}$
where $\left\{\left(\param^{(\ssnumdatatest)},
\stateobs^{(\ssnumdatatest)}\right)\right\}_{\ssnumdatatest=1}^{\numdatatest}$
is a dataset unseen by the training procedure and
$\meanpost\left(\stateobs^{(\ssnumdatatest)}\right)$ is the estimated posterior
mean from the neural network $\NN$ taking a datapoint
$\stateobs^{(\ssnumdatatest)}$ as an input. An estimate is said to be
\textit{feasible} if and only if true value lies within the estimated
uncertainty bounds. The uncertainty bounds displayed represent three standard
deviations. Similarly, when the PtO map is learned by $\decoder$, we use the
relative error
$\frac{1}{\numdatatest}\sum_{\ssnumdatatest=1}^{\numdatatest}
\frac{\left\lVert \stateobs^{(\ssnumdatatest)}
        - \decoder\left(\param\left(\stateobs^{(\ssnumdatatest)}\right)\right)
        \right\rVert_{2}^{2}}
        {\left\lVert \stateobs^{(\ssnumdatatest)}\right\rVert_{2}^{2}}$
where $\param\left(\stateobs^{(\ssnumdatatest)}\right)$ is a draw model
posterior output by the trained encoder with a datapoint
$\stateobs^{(\ssnumdatatest)}$ as an input.

%------------------------------------------------------------------------------
\subsection{Computational Cost} \label{SecComputationalCost}
%------------------------------------------------------------------------------
When training a neural network with the (\ref{EqOptLkhd}) term in the loss
functional (\ref{EqOpt}) using the PtO map $\model$, the solving of a linear
system is required. Because of this, training on CPUs is more efficient. In
contrast, when the (\ref{EqOptLkhd}) term uses the learned PtO map $\decoder$
instead, no linear system solves are required and so the training procedure is
most efficient on GPUs. We monitor the cost of training each batch as wall-clock
time for the whole optimization procedure. The offline cost for training the
networks using $\model$ is, on average, 4.5 seconds per batch on a dual-socket
node with two Intel Xeon E5-2690 CPUs for a total of 24 cores. The offline cost
for training the networks using $\decoder$ is, on average, 0.35 seconds per
batch on a NVIDIA 1080-TI GPU.

We compare the computational efficiency of inference between the Laplace
approximation and propagation through the neural network $\NN$ trained under the
UQ-VAE framework on a Intel Core i7-8550U CPU using an average wall-clock time
over 20 evaluations.  The time taken to form the MAP estimate with hIPPYlib is,
on average, 40 seconds with a maximum number of descent iterations of 25. The
time taken to form the low-rank Gaussian approximation of the posterior
covariance with 50 requested eigenvectors is, on average, 70 seconds. Therefore,
in total, it takes on average 110 seconds to form the Laplace approximation. In
contrast, forming the model posterior by propagation through a trained neural
network takes, on average, 0.04 seconds; more than 2750 times faster.

%==============================================================================
\section{Results: Two Dimensional Steady State Heat Conduction Problem}
\label{SecResults2DHeat}
%==============================================================================
%------------------------------------------------------------------------------
\subsection{$\noiselevel=0$, $\numdatatrain=50$} \label{SecHeat2Dn2601ns0M50}
%------------------------------------------------------------------------------
\begin{table}[H]
    \scriptsize
    \centering
    \begin{tabular}{c | c | c | c |}
        \cline{2-4}
        & \multicolumn{2}{c|}{Relative Error: $\param$} & \multicolumn{1}{c|}{Relative Error: $\stateobs$}\\
        \hline
        \multicolumn{1}{|c|}{$\penJS$} & Exact PtO & Learned PtO & Learned PtO\\
        \hline
        \multicolumn{1}{|c|}{0.00001} & 28.88\% & 30.86\% & 24.61\% \\
        \multicolumn{1}{|c|}{0.001}   & 30.46\% & 30.76\% & 22.80\% \\
        \multicolumn{1}{|c|}{0.1}     & 30.70\% & 28.95\% & 49.38\% \\
        \multicolumn{1}{|c|}{0.5}     & 29.75\% & 32.73\% & 45.34\% \\
        \hline
    \end{tabular}
    \caption{\scriptsize Table displaying the relative errors for UQ-VAE.
            Relative error of MAP estimate: 15.15\%.}
    \label{TableRelativeErrorsn2601ns0M50}
\end{table}
\begin{figure}[H]
    \centering
    \includegraphics[scale=\figscale]{Figures/poisson_2d/general/mesh_n2601.png}
    \includegraphics[scale=\figscale]{Figures/poisson_2d/general/parameter_test_n2601_128.png}
    \includegraphics[scale=\figscale]{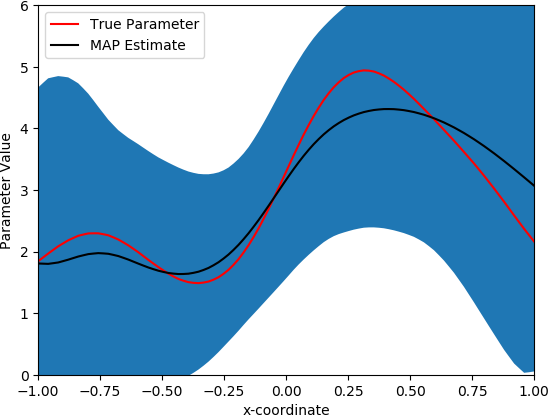}
    \includegraphics[scale=\figscale]{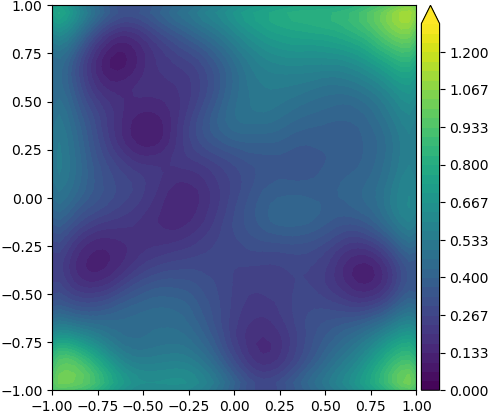}\\
    \includegraphics[scale=\figscale]{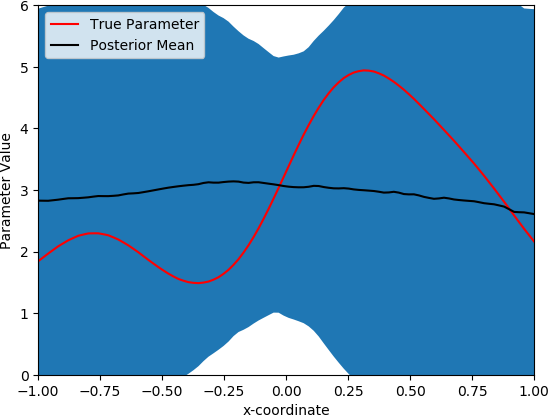}
    \includegraphics[scale=\figscale]{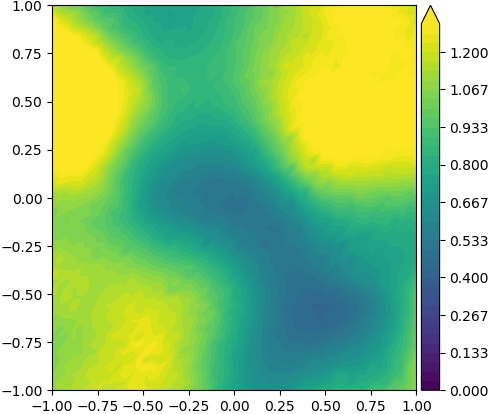}
    \includegraphics[scale=\figscale]{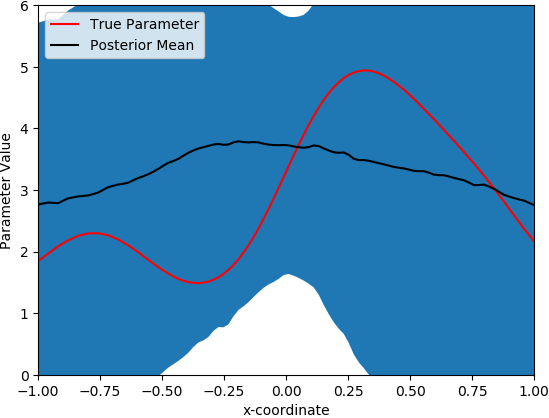}
    \includegraphics[scale=\figscale]{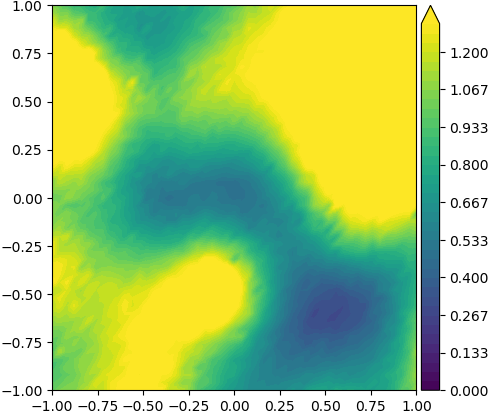}\\
    \includegraphics[scale=\figscale]{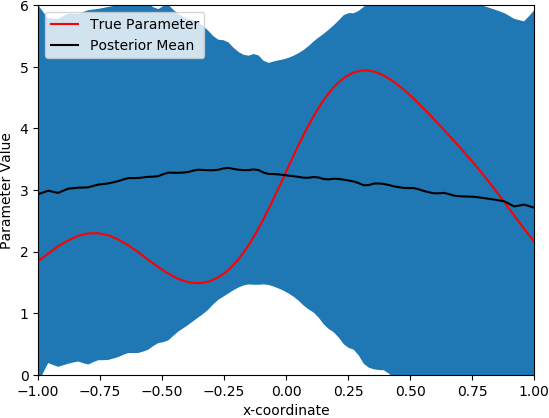}
    \includegraphics[scale=\figscale]{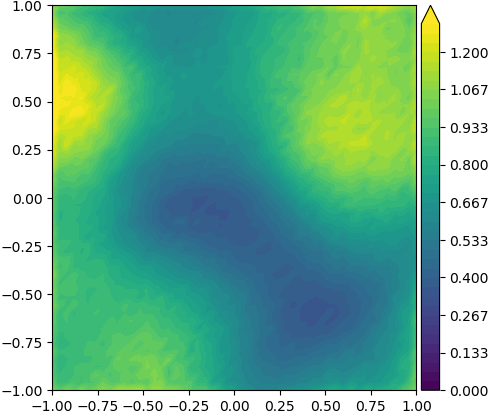}
    \includegraphics[scale=\figscale]{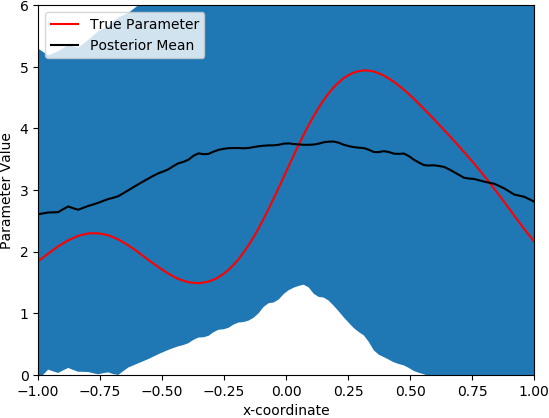}
    \includegraphics[scale=\figscale]{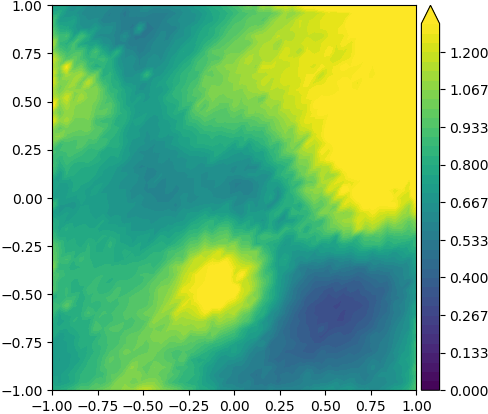}\\
    \includegraphics[scale=\figscale]{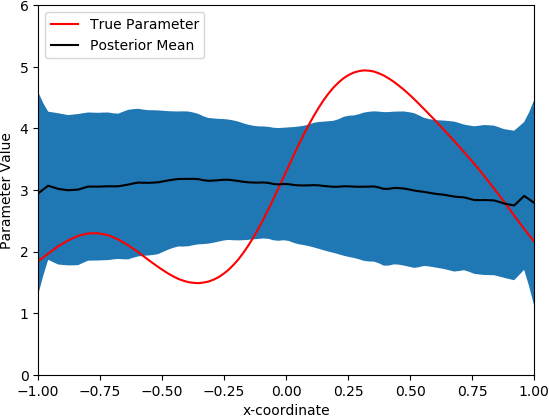}
    \includegraphics[scale=\figscale]{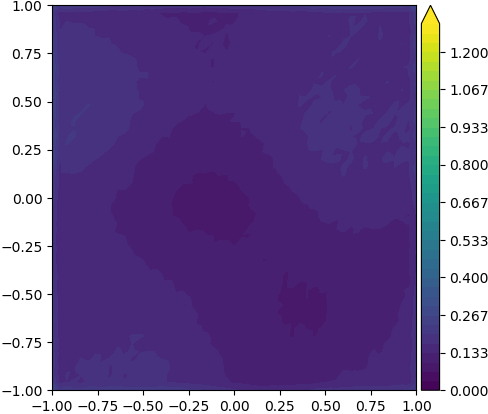}
    \includegraphics[scale=\figscale]{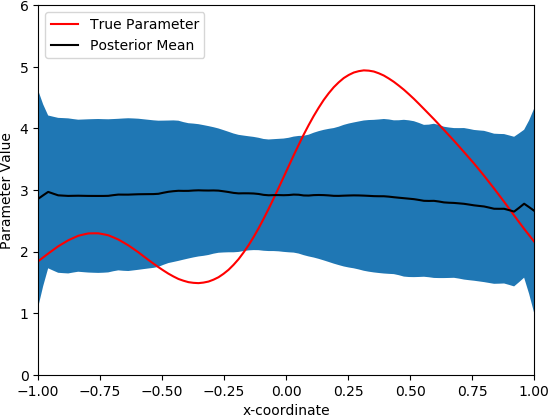}
    \includegraphics[scale=\figscale]{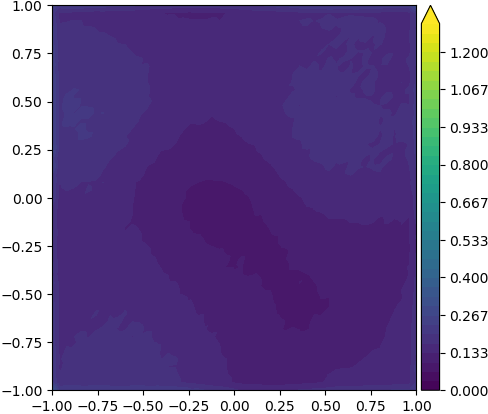}\\
    \includegraphics[scale=\figscale]{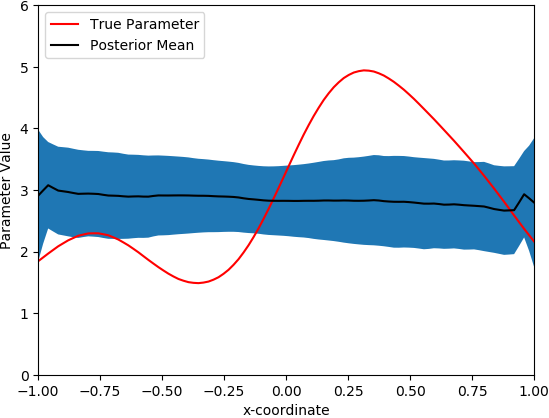}
    \includegraphics[scale=\figscale]{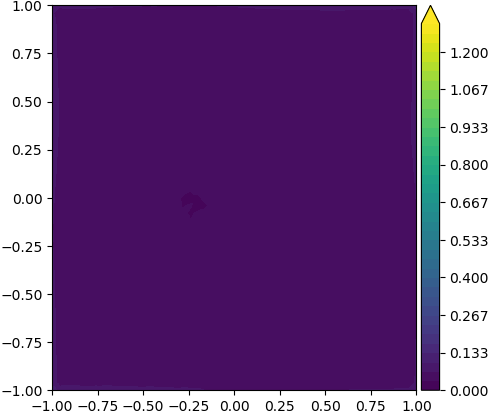}
    \includegraphics[scale=\figscale]{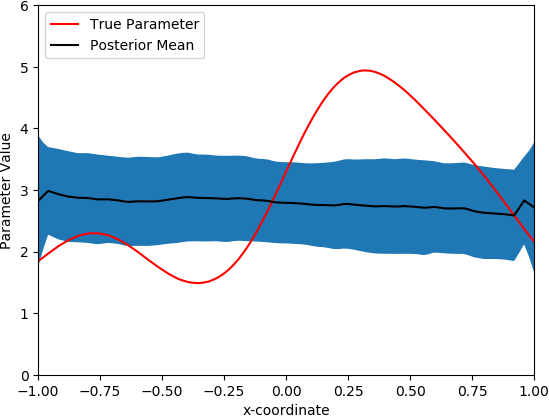}
    \includegraphics[scale=\figscale]{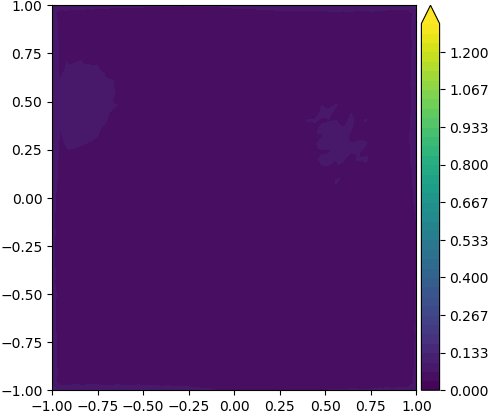}
    \caption{\scriptsize Top row left to right: mesh with sensors denoted with a red cross,
        true PoI,
        cross-sectional uncertainty estimate and pointwise posterior
        variance from Laplace approximation.
        Second to fourth rows: $\penJS =
        0.00001,0.001,0.1,0.5$. First and third columns: cross-sectional
        uncertainty estimates. Second and fourth columns: approximate pointwise
        posterior variance. First and second columns: exact PtO map.  Third and
        fourth columns: learned PtO map.}
        \label{FigPoisson2Dn2601ns0M50}
\end{figure}

%------------------------------------------------------------------------------
\subsection{$\noiselevel=0$, $\numdatatrain=500$} \label{SecHeat2Dn2601ns0M500}
%------------------------------------------------------------------------------
\begin{table}[H]
    \scriptsize
    \centering
    \begin{tabular}{c | c | c | c |}
        \cline{2-4}
        & \multicolumn{2}{c|}{Relative Error: $\param$} & \multicolumn{1}{c|}{Relative Error: $\stateobs$}\\
        \hline
        \multicolumn{1}{|c|}{$\penJS$} & Exact PtO & Learned PtO & Learned PtO\\
        \hline
        \multicolumn{1}{|c|}{0.00001} & 21.72\% & 21.70\% & 22.49\% \\
        \multicolumn{1}{|c|}{0.001}   & 21.33\% & 22.13\% & 20.75\% \\
        \multicolumn{1}{|c|}{0.1}     & 21.84\% & 21.91\% & 24.08\% \\
        \multicolumn{1}{|c|}{0.5}     & 26.75\% & 29.47\% & 35.26\% \\
        \hline
    \end{tabular}
    \caption{\scriptsize Table displaying the relative errors for UQ-VAE.
            Relative error of MAP estimate: 15.15\%.}
    \label{TableRelativeErrorsn2601ns0M500}
\end{table}
\begin{figure}[H]
    \centering
    \includegraphics[scale=\figscale]{Figures/poisson_2d/general/mesh_n2601.png}
    \includegraphics[scale=\figscale]{Figures/poisson_2d/general/parameter_test_n2601_128.png}
    \includegraphics[scale=\figscale]{Figures/poisson_2d/ns0/n2601/traditional/parameter_cross_section.png}
    \includegraphics[scale=\figscale]{Figures/poisson_2d/ns0/n2601/traditional/posterior_covariance.png}\\
    \includegraphics[scale=\figscale]{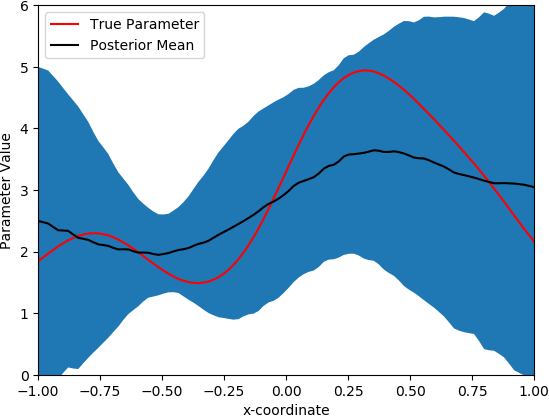}
    \includegraphics[scale=\figscale]{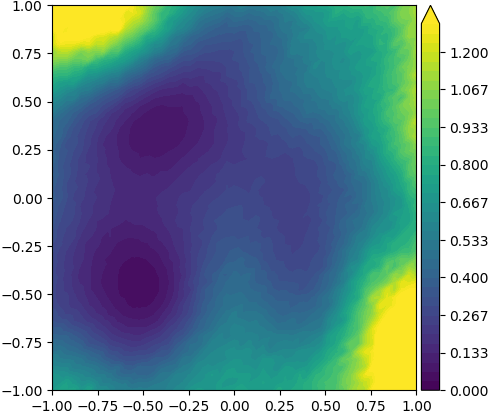}
    \includegraphics[scale=\figscale]{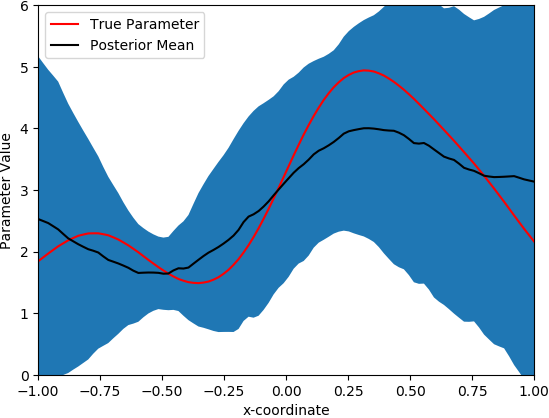}
    \includegraphics[scale=\figscale]{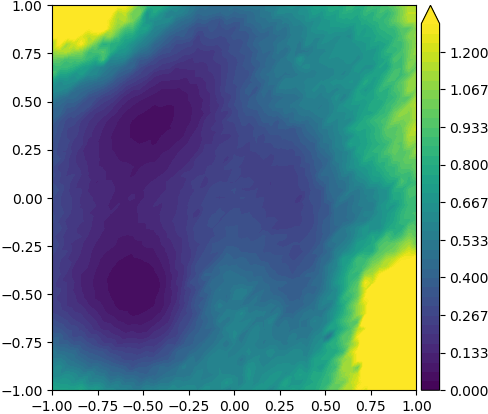}\\
    \includegraphics[scale=\figscale]{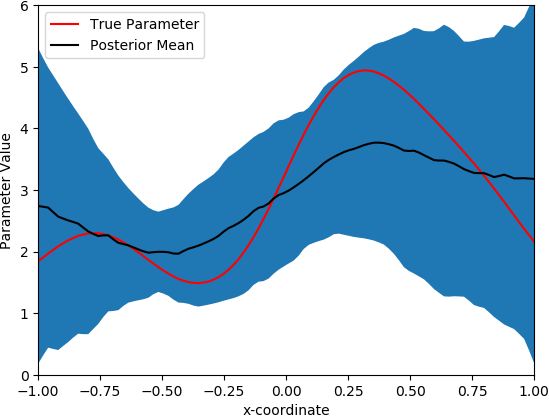}
    \includegraphics[scale=\figscale]{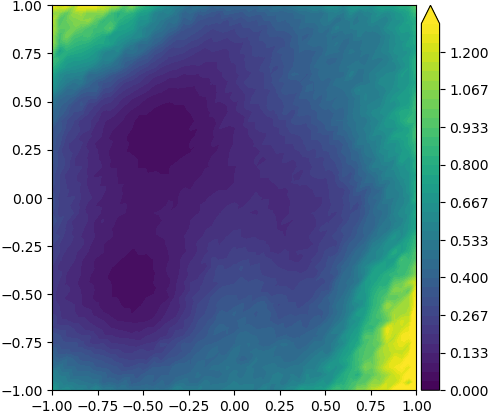}
    \includegraphics[scale=\figscale]{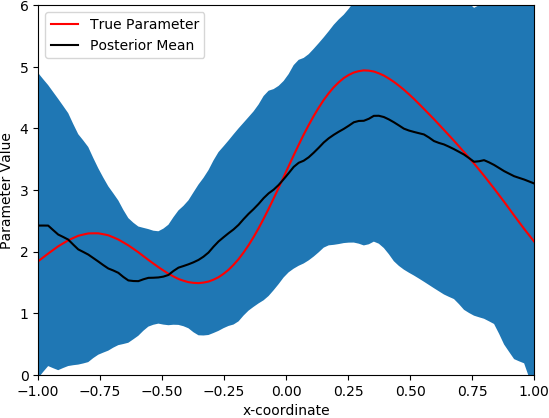}
    \includegraphics[scale=\figscale]{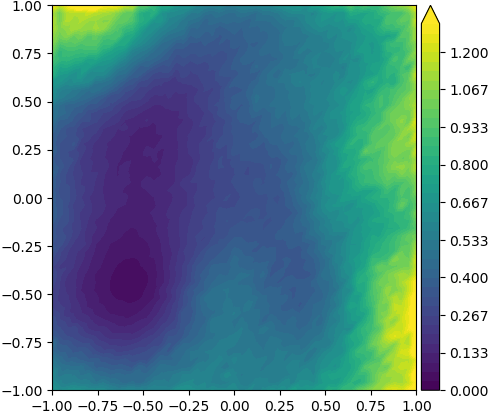}\\
    \includegraphics[scale=\figscale]{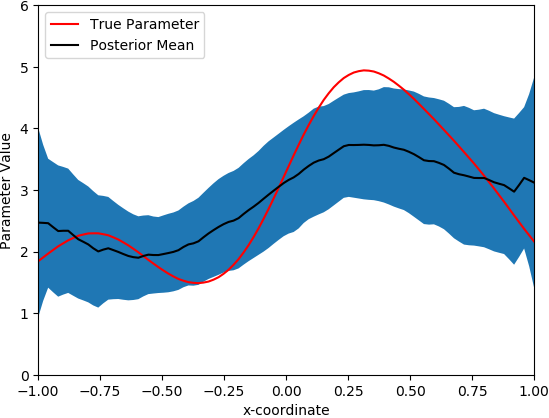}
    \includegraphics[scale=\figscale]{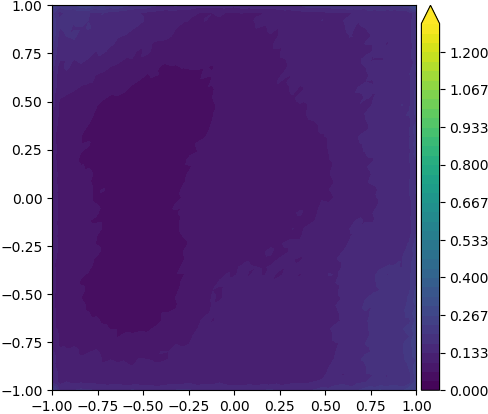}
    \includegraphics[scale=\figscale]{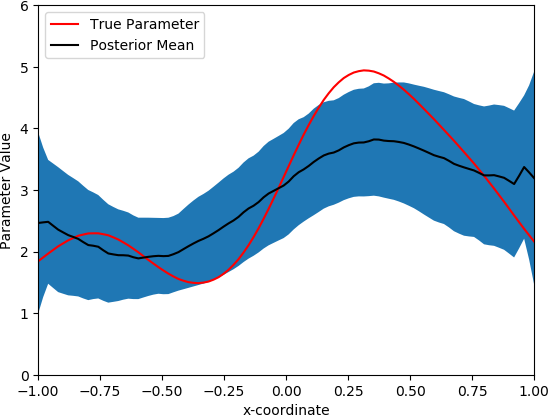}
    \includegraphics[scale=\figscale]{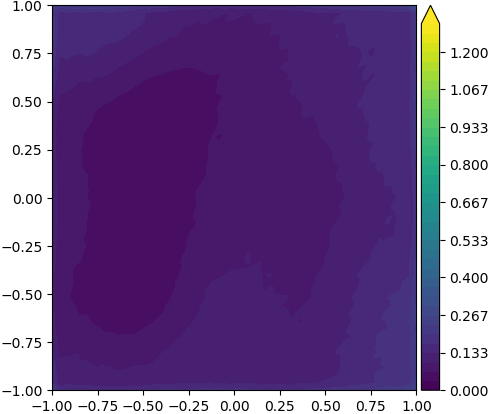}\\
    \includegraphics[scale=\figscale]{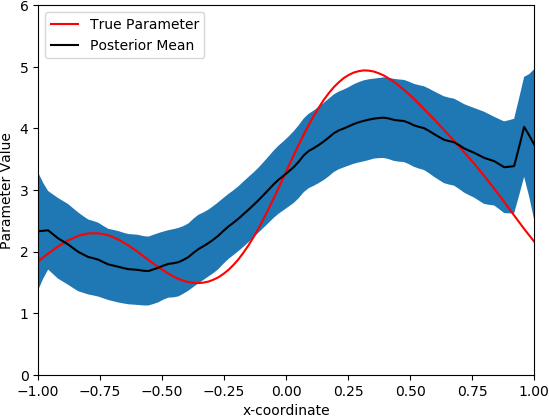}
    \includegraphics[scale=\figscale]{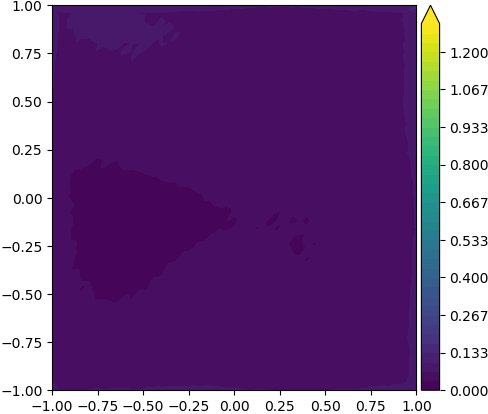}
    \includegraphics[scale=\figscale]{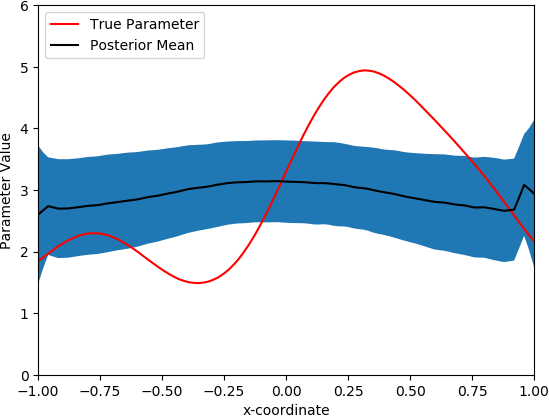}
    \includegraphics[scale=\figscale]{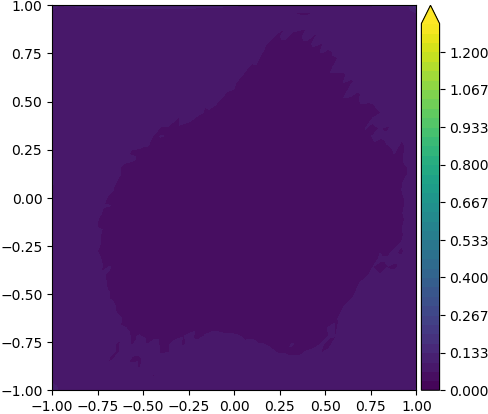}
    \caption{\scriptsize Top row left to right: mesh with sensors denoted with a red cross,
        true PoI,
        cross-sectional uncertainty estimate and pointwise posterior
        variance from Laplace approximation.
        Second to fourth rows: $\penJS =
        0.00001,0.001,0.1,0.5$. First and third columns: cross-sectional
        uncertainty estimates. Second and fourth columns: approximate pointwise
        posterior variance. First and second columns: exact PtO map.  Third and
        fourth columns: learned PtO map.}
        \label{FigPoisson2Dn2601ns0M500}
\end{figure}

%------------------------------------------------------------------------------
\subsection{$\noiselevel=0$, $\numdatatrain=1000$} \label{SecHeat2Dn2601ns0M1000}
%------------------------------------------------------------------------------
\begin{table}[H]
    \scriptsize
    \centering
    \begin{tabular}{c | c | c | c |}
        \cline{2-4}
        & \multicolumn{2}{c|}{Relative Error: $\param$} & \multicolumn{1}{c|}{Relative Error: $\stateobs$}\\
        \hline
        \multicolumn{1}{|c|}{$\penJS$} & Exact PtO & Learned PtO & Learned PtO\\
        \hline
        \multicolumn{1}{|c|}{0.00001} & 20.65\% & 18.15\% & 18.55\% \\
        \multicolumn{1}{|c|}{0.001}   & 18.54\% & 19.04\% & 15.02\% \\
        \multicolumn{1}{|c|}{0.1}     & 18.74\% & 19.28\% & 22.72\% \\
        \multicolumn{1}{|c|}{0.5}     & 21.84\% & 22.74\% & 21.74\% \\
        \hline
    \end{tabular}
    \caption{\scriptsize Table displaying the relative errors for UQ-VAE.
            Relative error of MAP estimate: 15.15\%.}
    \label{TableRelativeErrorsn2601ns0M1000}
\end{table}
\begin{figure}[H]
    \centering
    \includegraphics[scale=\figscale]{Figures/poisson_2d/general/mesh_n2601.png}
    \includegraphics[scale=\figscale]{Figures/poisson_2d/general/parameter_test_n2601_128.png}
    \includegraphics[scale=\figscale]{Figures/poisson_2d/ns0/n2601/traditional/parameter_cross_section.png}
    \includegraphics[scale=\figscale]{Figures/poisson_2d/ns0/n2601/traditional/posterior_covariance.png}\\
    \includegraphics[scale=\figscale]{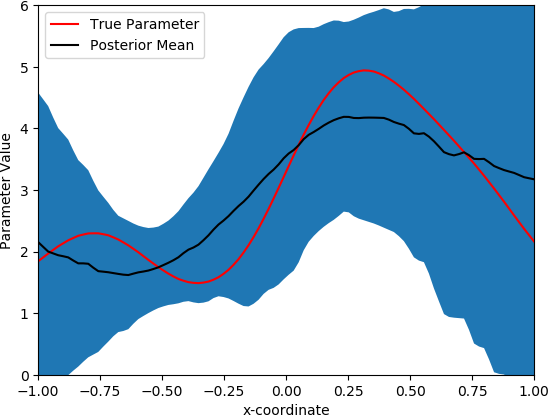}
    \includegraphics[scale=\figscale]{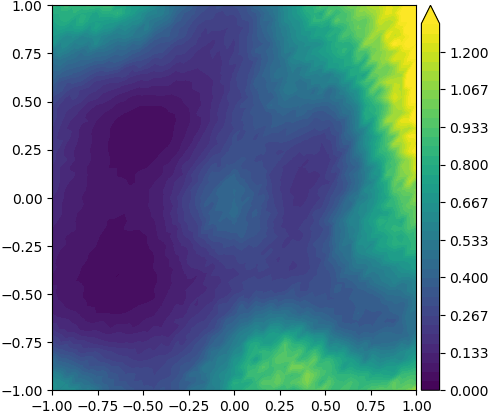}
    \includegraphics[scale=\figscale]{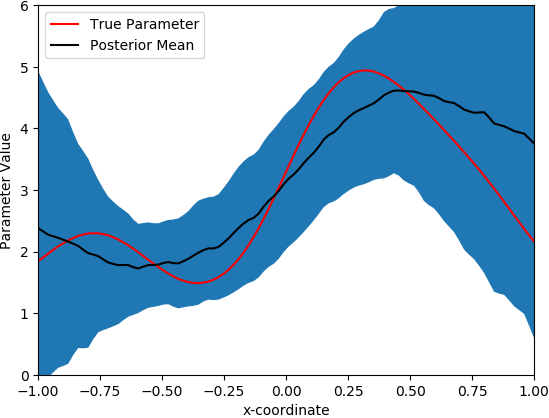}
    \includegraphics[scale=\figscale]{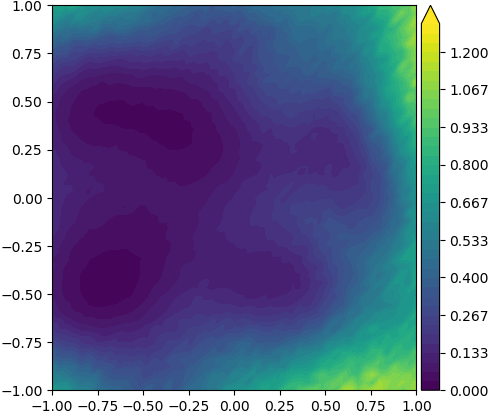}\\
    \includegraphics[scale=\figscale]{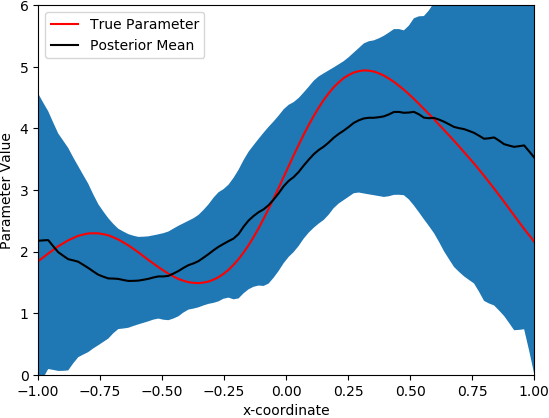}
    \includegraphics[scale=\figscale]{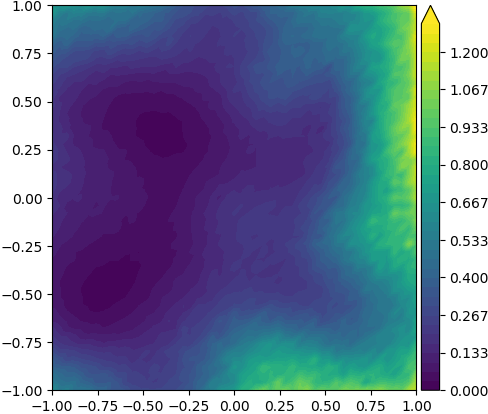}
    \includegraphics[scale=\figscale]{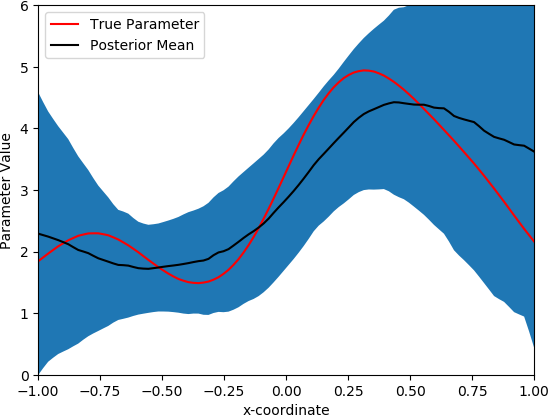}
    \includegraphics[scale=\figscale]{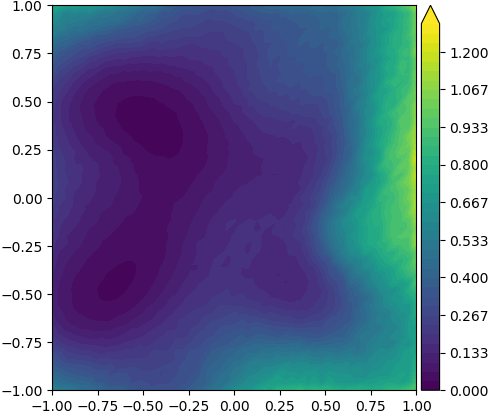}\\
    \includegraphics[scale=\figscale]{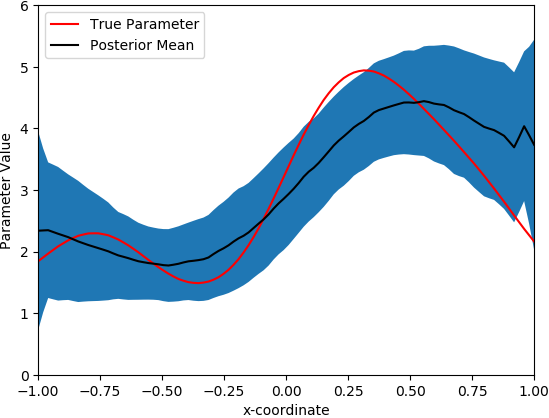}
    \includegraphics[scale=\figscale]{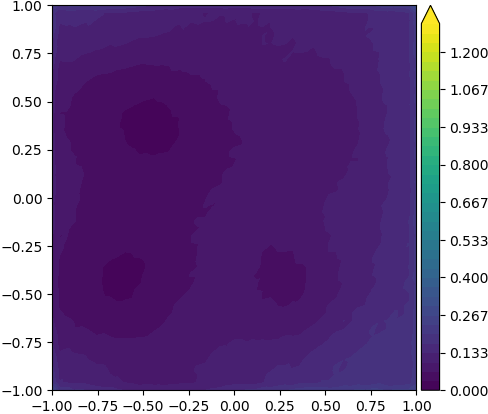}
    \includegraphics[scale=\figscale]{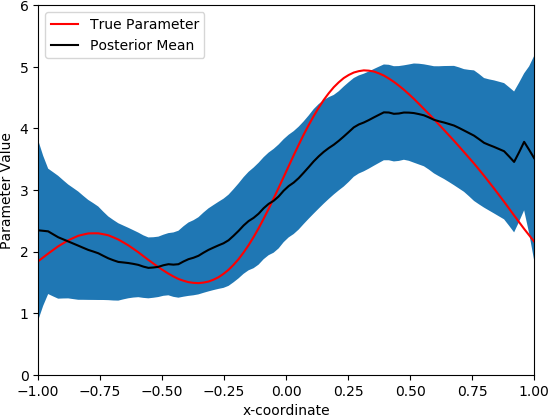}
    \includegraphics[scale=\figscale]{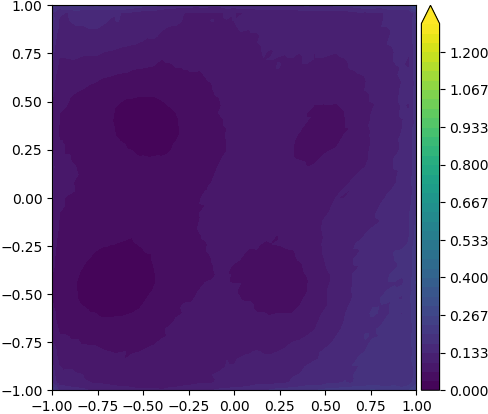}\\
    \includegraphics[scale=\figscale]{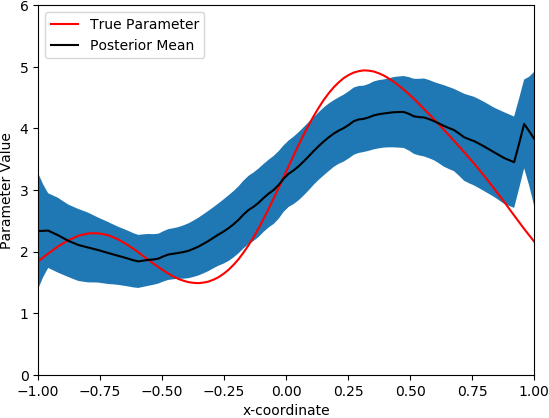}
    \includegraphics[scale=\figscale]{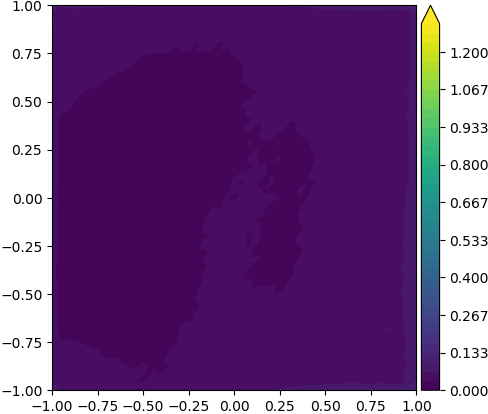}
    \includegraphics[scale=\figscale]{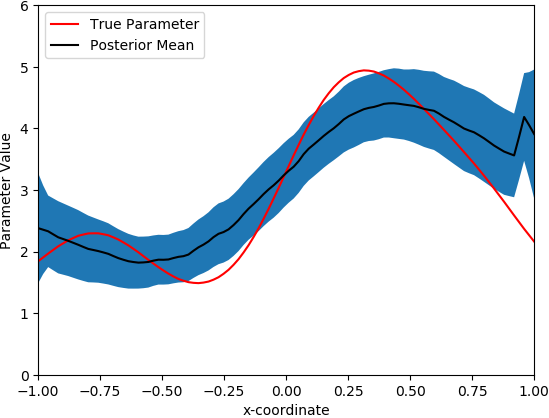}
    \includegraphics[scale=\figscale]{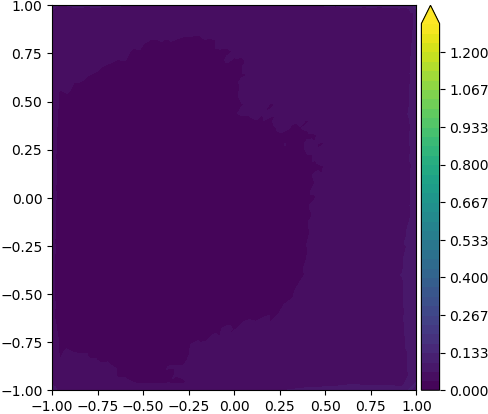}
    \caption{\scriptsize Top row left to right: mesh with sensors denoted with a red cross,
        true PoI,
        cross-sectional uncertainty estimate and pointwise posterior
        variance from Laplace approximation.
        Second to fourth rows: $\penJS =
        0.00001,0.001,0.1,0.5$. First and third columns: cross-sectional
        uncertainty estimates. Second and fourth columns: approximate pointwise
        posterior variance. First and second columns: exact PtO map.  Third and
        fourth columns: learned PtO map.}
        \label{FigPoisson2Dn2601ns0M1000}
\end{figure}

%------------------------------------------------------------------------------
\subsection{$\noiselevel=0$, $\numdatatrain=5000$} \label{SecHeat2Dn2601ns0M5000}
%------------------------------------------------------------------------------
\begin{table}[H]
    \scriptsize
    \centering
    \begin{tabular}{c | c | c | c |}
        \cline{2-4}
        & \multicolumn{2}{c|}{Relative Error: $\param$} & \multicolumn{1}{c|}{Relative Error: $\stateobs$}\\
        \hline
        \multicolumn{1}{|c|}{$\penJS$} & Exact PtO & Learned PtO & Learned PtO\\
        \hline
        \multicolumn{1}{|c|}{0.00001} & 14.49\% & 15.07\% & 3.21\% \\
        \multicolumn{1}{|c|}{0.001}   & 14.28\% & 14.07\% & 3.13\% \\
        \multicolumn{1}{|c|}{0.1}     & 14.72\% & 14.48\% & 3.19\% \\
        \multicolumn{1}{|c|}{0.5}     & 14.89\% & 15.47\% & 4.72\% \\
        \hline
    \end{tabular}
    \caption{\scriptsize Table displaying the relative errors for UQ-VAE.
            Relative error of MAP estimate: 15.15\%.}
    \label{TableRelativeErrorsn2601ns0M5000}
\end{table}
\begin{figure}[H]
    \centering
    \includegraphics[scale=\figscale]{Figures/poisson_2d/general/mesh_n2601.png}
    \includegraphics[scale=\figscale]{Figures/poisson_2d/general/parameter_test_n2601_128.png}
    \includegraphics[scale=\figscale]{Figures/poisson_2d/ns0/n2601/traditional/parameter_cross_section.png}
    \includegraphics[scale=\figscale]{Figures/poisson_2d/ns0/n2601/traditional/posterior_covariance.png}\\
    \includegraphics[scale=\figscale]{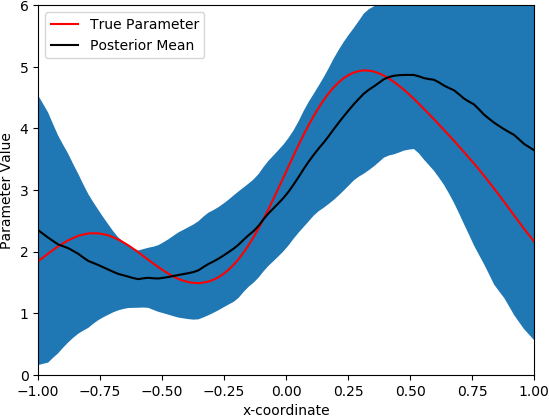}
    \includegraphics[scale=\figscale]{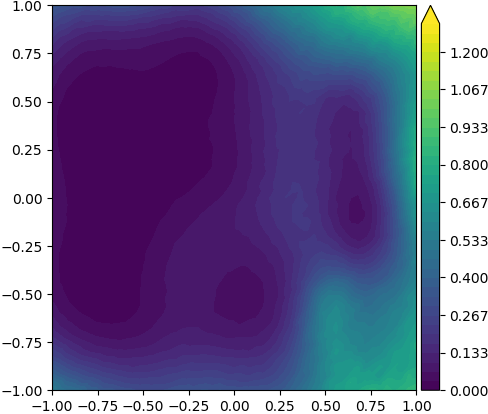}
    \includegraphics[scale=\figscale]{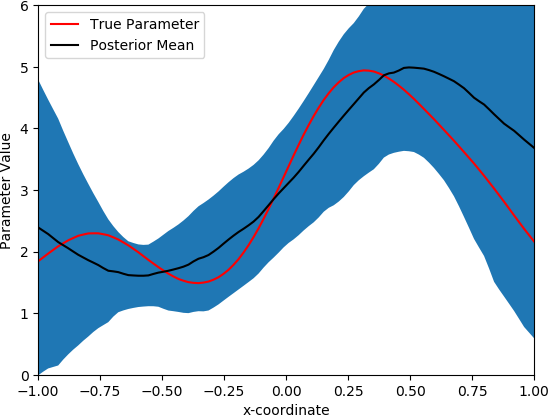}
    \includegraphics[scale=\figscale]{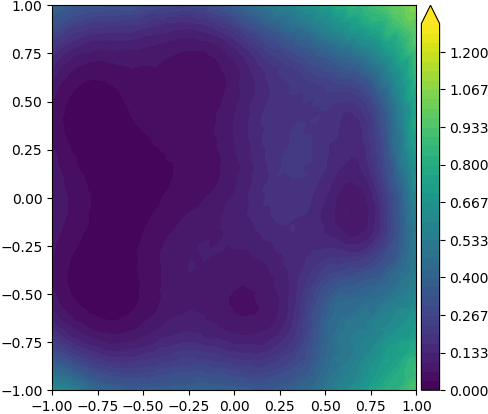}\\
    \includegraphics[scale=\figscale]{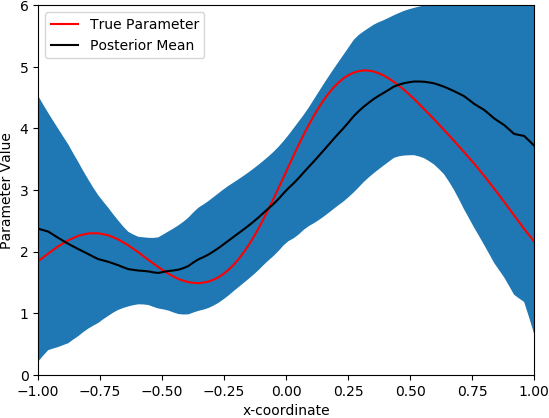}
    \includegraphics[scale=\figscale]{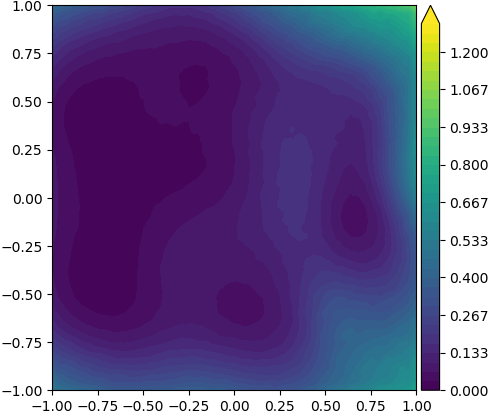}
    \includegraphics[scale=\figscale]{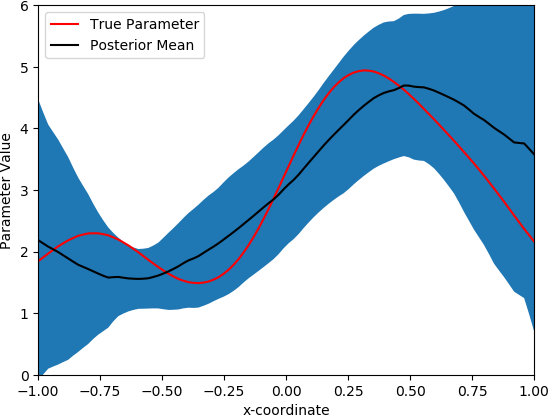}
    \includegraphics[scale=\figscale]{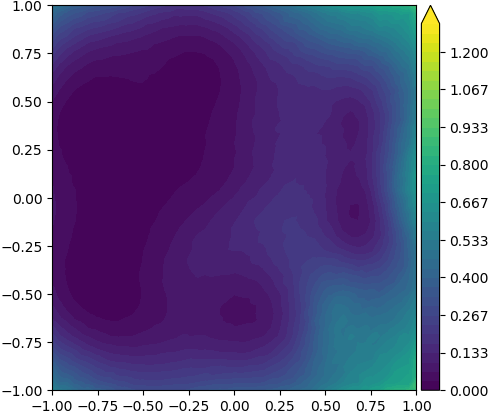}\\
    \includegraphics[scale=\figscale]{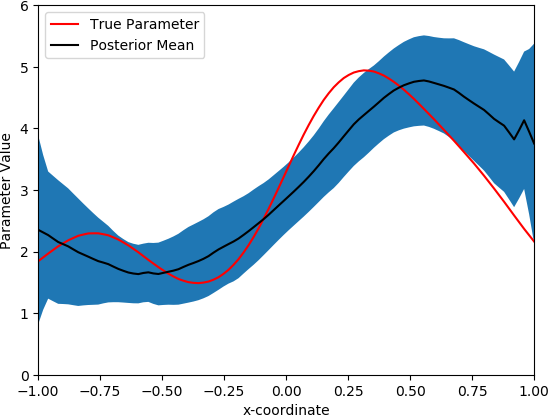}
    \includegraphics[scale=\figscale]{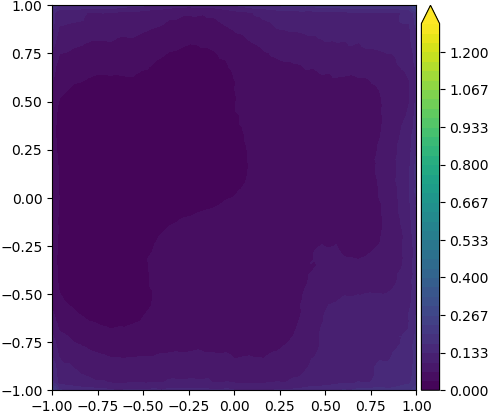}
    \includegraphics[scale=\figscale]{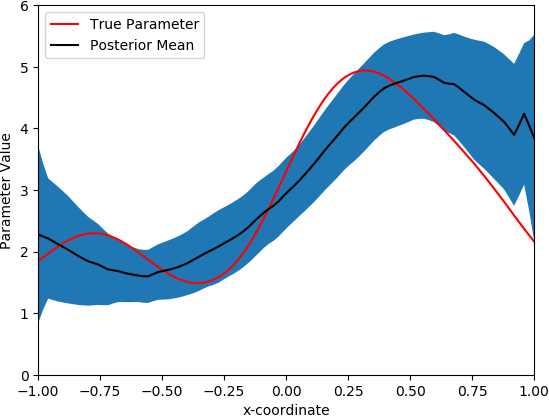}
    \includegraphics[scale=\figscale]{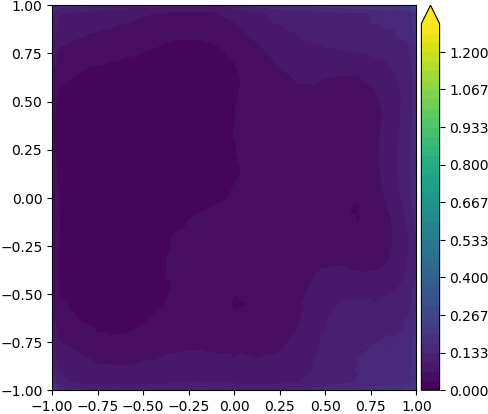}\\
    \includegraphics[scale=\figscale]{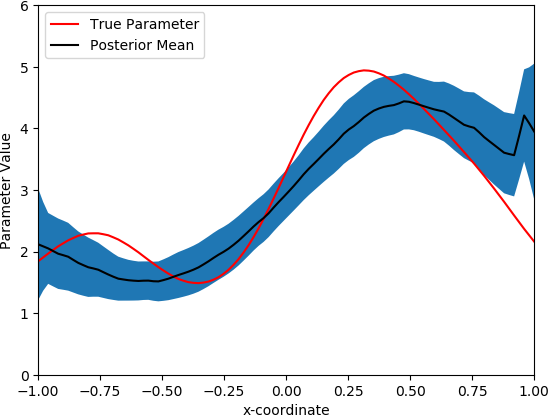}
    \includegraphics[scale=\figscale]{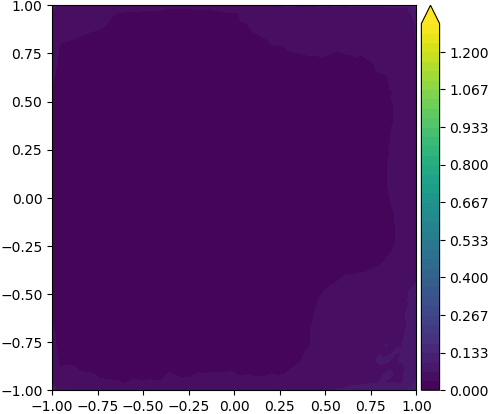}
    \includegraphics[scale=\figscale]{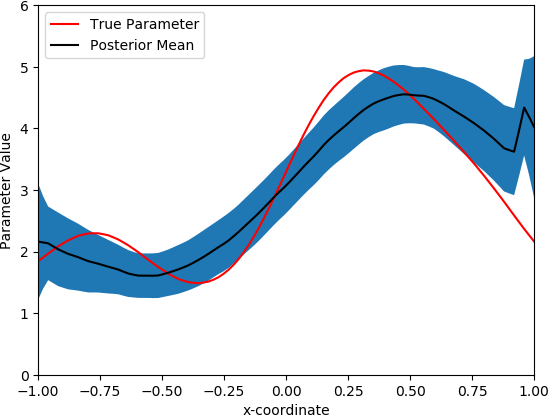}
    \includegraphics[scale=\figscale]{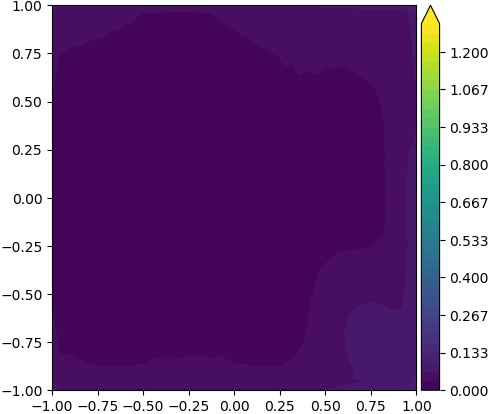}
    \caption{\scriptsize Top row left to right: mesh with sensors denoted with a red cross,
        true PoI,
        cross-sectional uncertainty estimate and pointwise posterior
        variance from Laplace approximation.
        Second to fourth rows: $\penJS =
        0.00001,0.001,0.1,0.5$. First and third columns: cross-sectional
        uncertainty estimates. Second and fourth columns: approximate pointwise
        posterior variance. First and second columns: exact PtO map.  Third and
        fourth columns: learned PtO map.}
        \label{FigPoisson2Dn2601ns0M5000}
\end{figure}

%------------------------------------------------------------------------------
\subsection{$\noiselevel=0.01$, $\numdatatrain=50$} \label{SecHeat2Dn2601ns1M50}
%------------------------------------------------------------------------------
\begin{table}[H]
    \scriptsize
    \centering
    \begin{tabular}{c | c | c | c |}
        \cline{2-4}
        & \multicolumn{2}{c|}{Relative Error: $\param$} & \multicolumn{1}{c|}{Relative Error: $\stateobs$}\\
        \hline
        \multicolumn{1}{|c|}{$\penJS$} & Exact PtO & Learned PtO & Learned PtO\\
        \hline
        \multicolumn{1}{|c|}{0.00001} & 29.19\% & 30.67\% & 24.59\% \\
        \multicolumn{1}{|c|}{0.001}   & 39.52\% & 29.25\% & 31.88\% \\
        \multicolumn{1}{|c|}{0.1}     & 28.99\% & 28.95\% & 34.97\% \\
        \multicolumn{1}{|c|}{0.5}     & 32.92\% & 30.05\% & 41.18\% \\
        \hline
    \end{tabular}
    \caption{\scriptsize Table displaying the relative errors for UQ-VAE.
            Relative error of MAP estimate: 25.11\%.}
    \label{TableRelativeErrorsn2601nspt01M50}
\end{table}
\begin{figure}[H]
    \centering
    \includegraphics[scale=\figscale]{Figures/poisson_2d/general/mesh_n2601.png}
    \includegraphics[scale=\figscale]{Figures/poisson_2d/general/parameter_test_n2601_128.png}
    \includegraphics[scale=\figscale]{Figures/poisson_2d/nspt01/n2601/traditional/parameter_cross_section.png}
    \includegraphics[scale=\figscale]{Figures/poisson_2d/nspt01/n2601/traditional/posterior_covariance.png}\\
    \includegraphics[scale=\figscale]{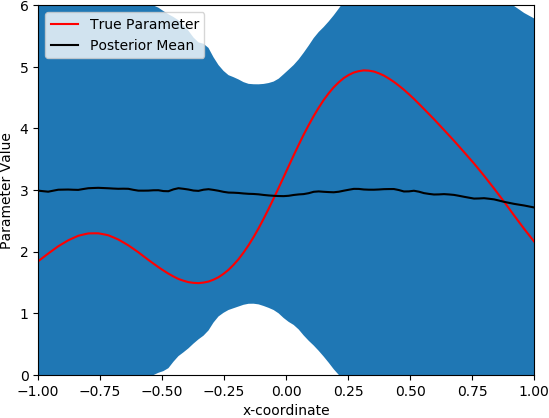}
    \includegraphics[scale=\figscale]{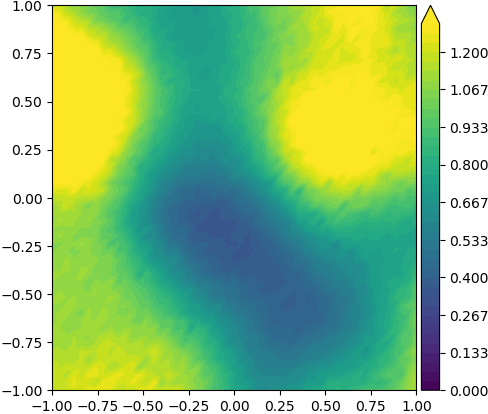}
    \includegraphics[scale=\figscale]{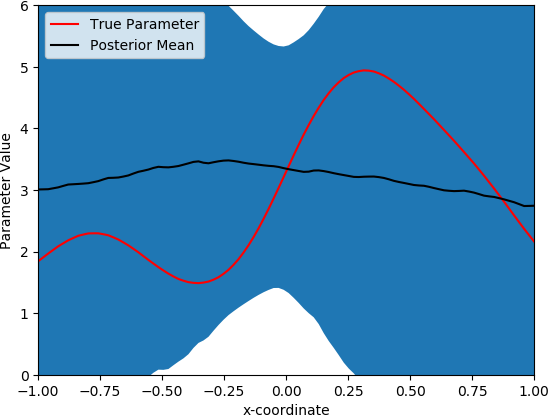}
    \includegraphics[scale=\figscale]{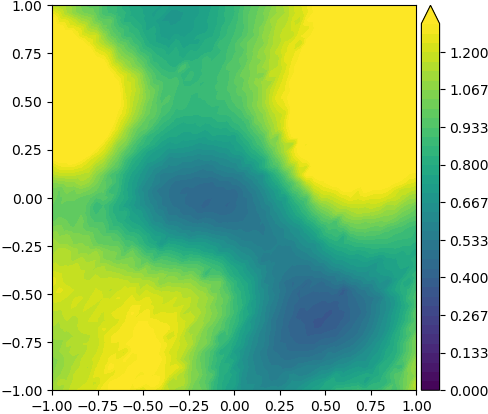}\\
    \includegraphics[scale=\figscale]{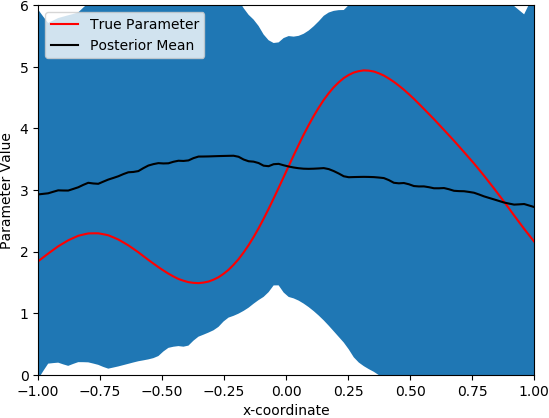}
    \includegraphics[scale=\figscale]{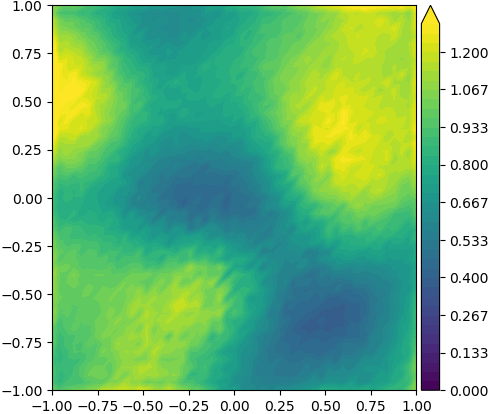}
    \includegraphics[scale=\figscale]{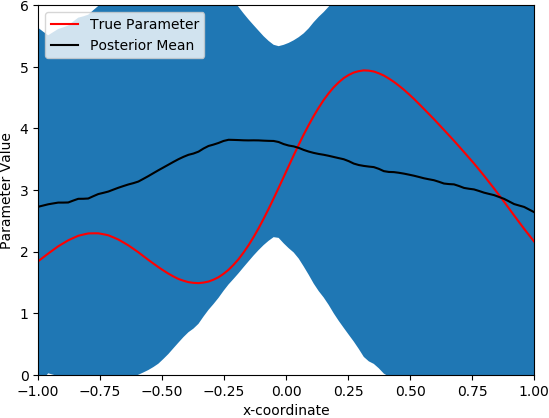}
    \includegraphics[scale=\figscale]{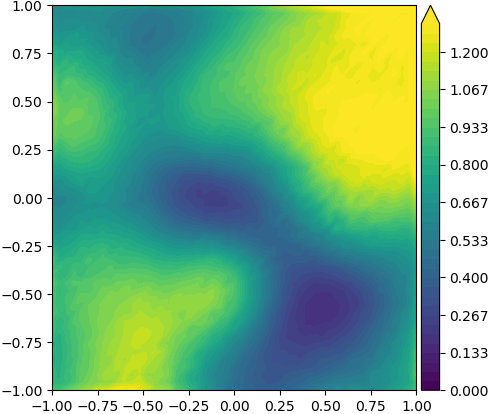}\\
    \includegraphics[scale=\figscale]{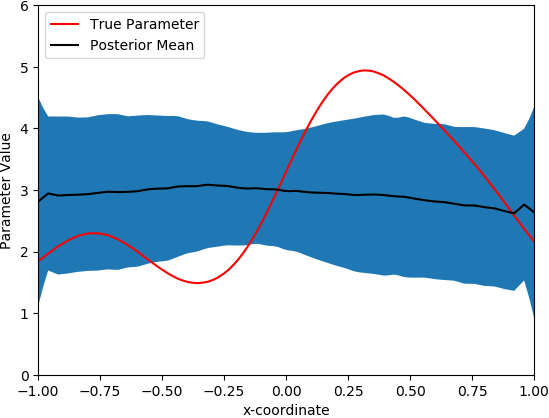}
    \includegraphics[scale=\figscale]{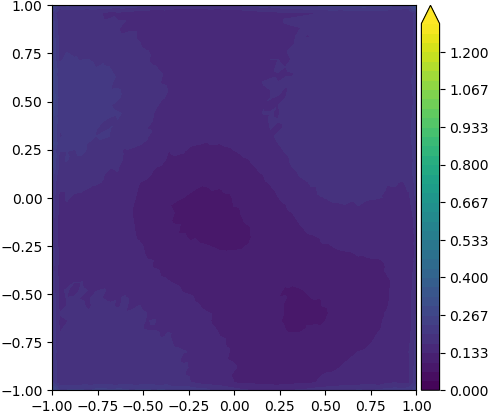}
    \includegraphics[scale=\figscale]{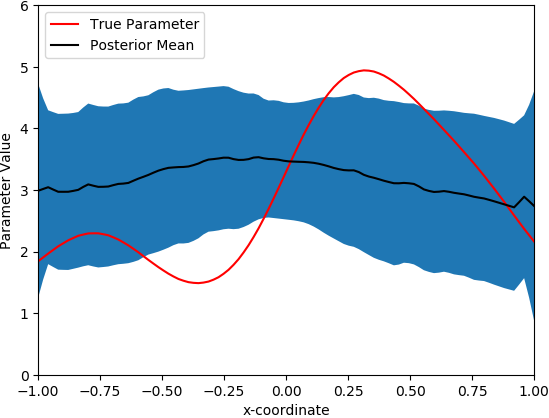}
    \includegraphics[scale=\figscale]{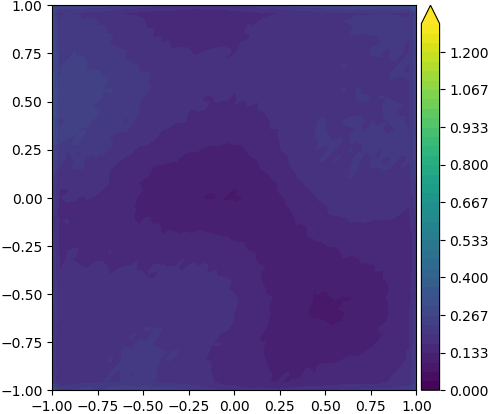}\\
    \includegraphics[scale=\figscale]{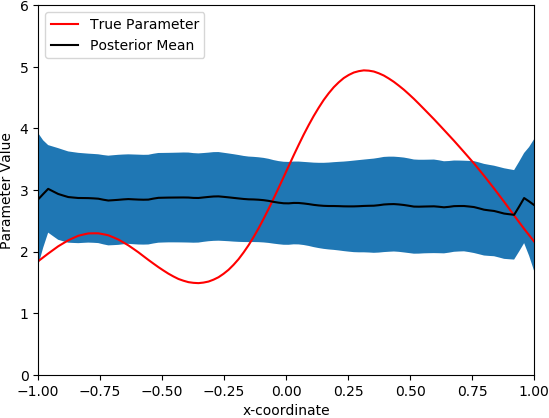}
    \includegraphics[scale=\figscale]{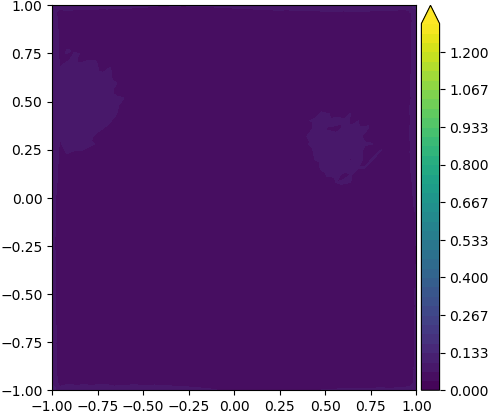}
    \includegraphics[scale=\figscale]{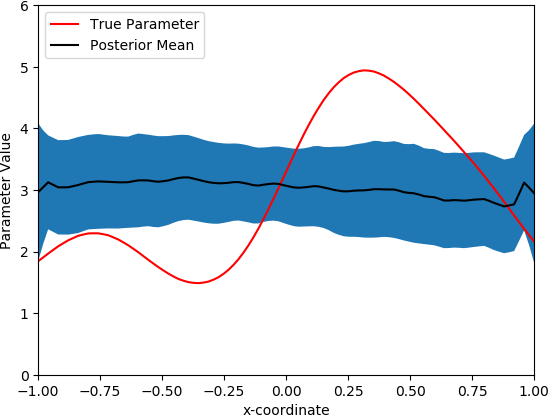}
    \includegraphics[scale=\figscale]{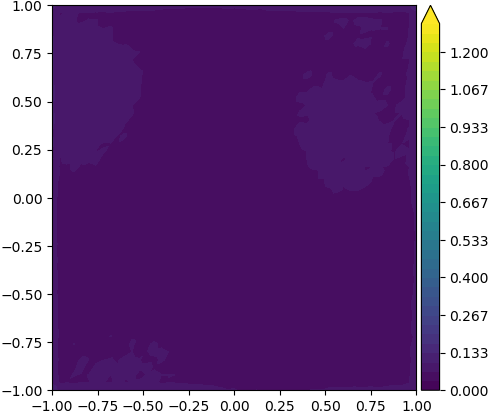}
    \caption{\scriptsize Top row left to right: mesh with sensors denoted with a red cross,
        true PoI,
        cross-sectional uncertainty estimate and pointwise posterior
        variance from Laplace approximation.
        Second to fourth rows: $\penJS =
        0.00001,0.001,0.1,0.5$. First and third columns: cross-sectional
        uncertainty estimates. Second and fourth columns: approximate pointwise
        posterior variance. First and second columns: exact PtO map.  Third and
        fourth columns: learned PtO map.}
        \label{FigPoisson2Dn2601nspt01M50}
\end{figure}

%------------------------------------------------------------------------------
\subsection{$\noiselevel=0.01$, $\numdatatrain=500$} \label{SecHeat2Dn2601ns1M500}
%------------------------------------------------------------------------------
\begin{table}[H]
    \scriptsize
    \centering
    \begin{tabular}{c | c | c | c |}
        \cline{2-4}
        & \multicolumn{2}{c|}{Relative Error: $\param$} & \multicolumn{1}{c|}{Relative Error: $\stateobs$}\\
        \hline
        \multicolumn{1}{|c|}{$\penJS$} & Exact PtO & Learned PtO & Learned PtO\\
        \hline
        \multicolumn{1}{|c|}{0.00001} & 24.17\% & 23.42\% & 22.02\% \\
        \multicolumn{1}{|c|}{0.001}   & 22.79\% & 23.23\% & 17.41\% \\
        \multicolumn{1}{|c|}{0.1}     & 25.51\% & 29.52\% & 28.16\% \\
        \multicolumn{1}{|c|}{0.5}     & 29.93\% & 29.82\% & 32.68\% \\
        \hline
    \end{tabular}
    \caption{\scriptsize Table displaying the relative errors for UQ-VAE.
            Relative error of MAP estimate: 25.11\%.}
    \label{TableRelativeErrorsn2601nspt01M500}
\end{table}
\begin{figure}[H]
    \centering
    \includegraphics[scale=\figscale]{Figures/poisson_2d/general/mesh_n2601.png}
    \includegraphics[scale=\figscale]{Figures/poisson_2d/general/parameter_test_n2601_128.png}
    \includegraphics[scale=\figscale]{Figures/poisson_2d/nspt01/n2601/traditional/parameter_cross_section.png}
    \includegraphics[scale=\figscale]{Figures/poisson_2d/nspt01/n2601/traditional/posterior_covariance.png}\\
    \includegraphics[scale=\figscale]{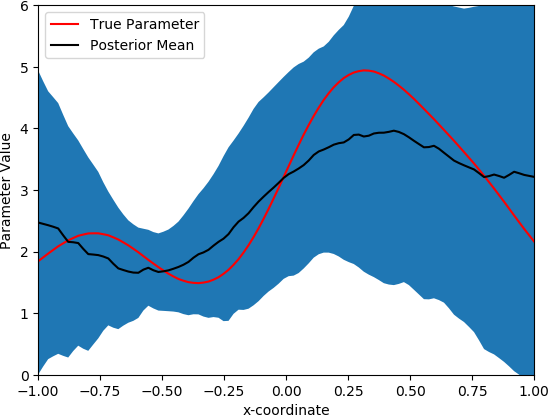}
    \includegraphics[scale=\figscale]{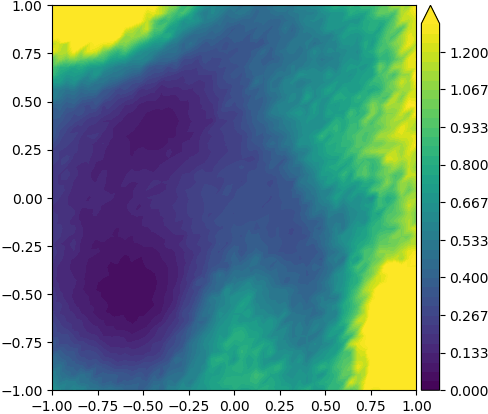}
    \includegraphics[scale=\figscale]{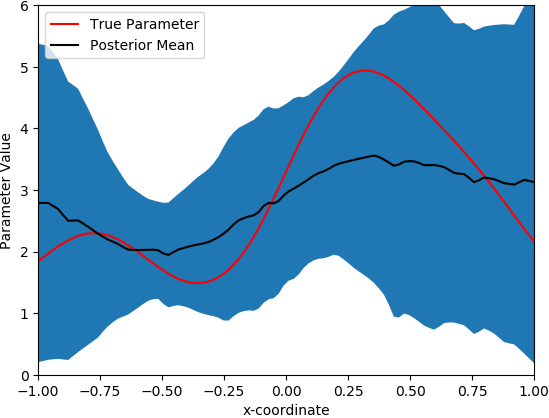}
    \includegraphics[scale=\figscale]{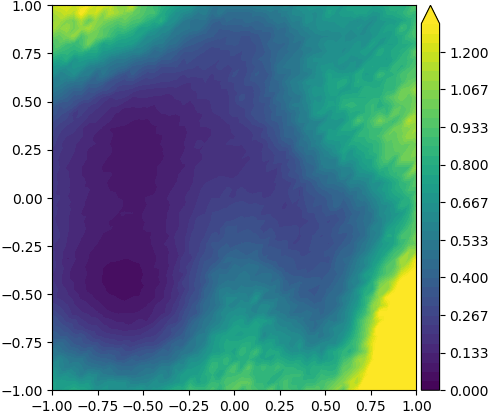}\\
    \includegraphics[scale=\figscale]{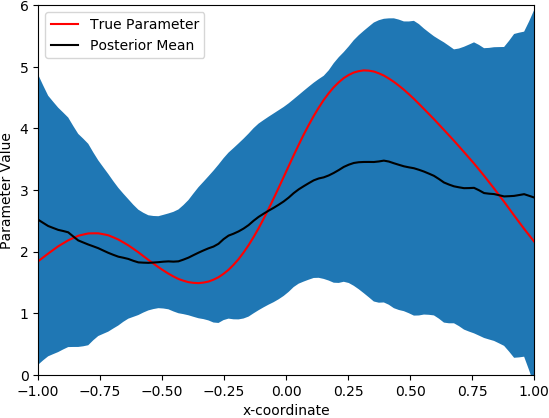}
    \includegraphics[scale=\figscale]{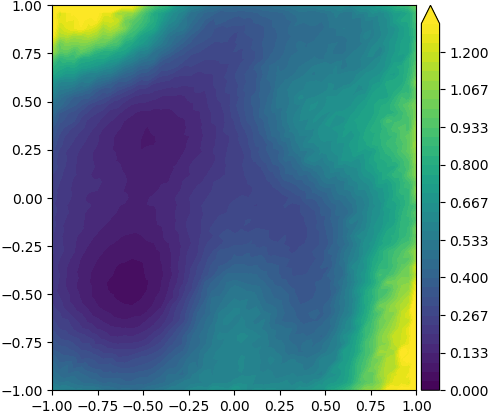}
    \includegraphics[scale=\figscale]{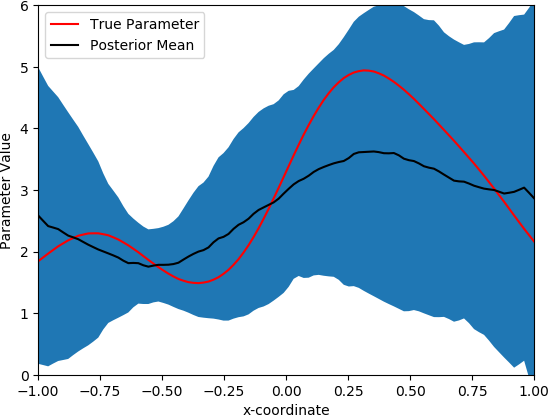}
    \includegraphics[scale=\figscale]{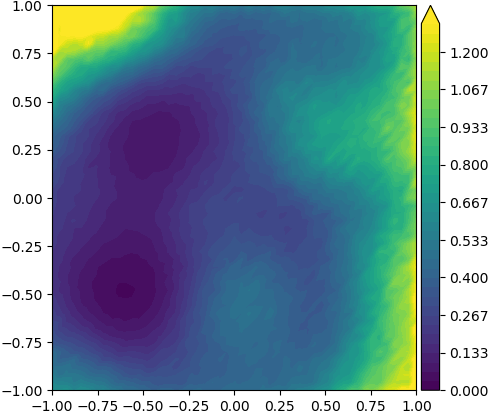}\\
    \includegraphics[scale=\figscale]{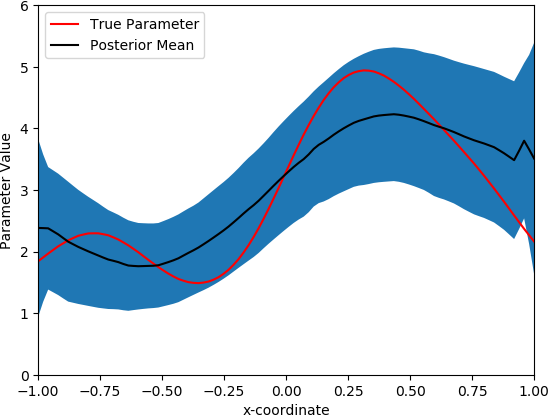}
    \includegraphics[scale=\figscale]{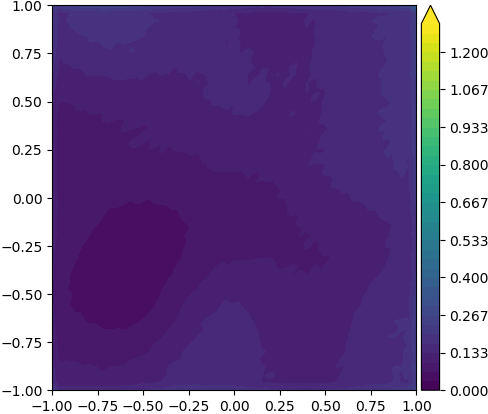}
    \includegraphics[scale=\figscale]{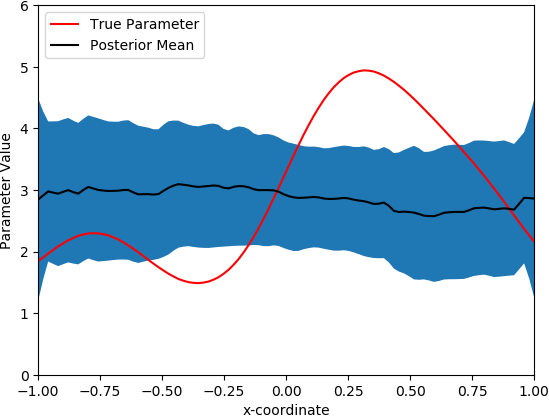}
    \includegraphics[scale=\figscale]{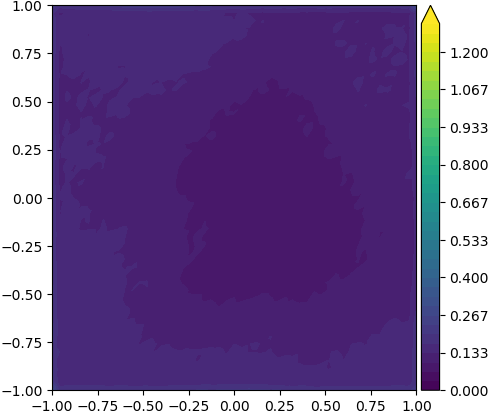}\\
    \includegraphics[scale=\figscale]{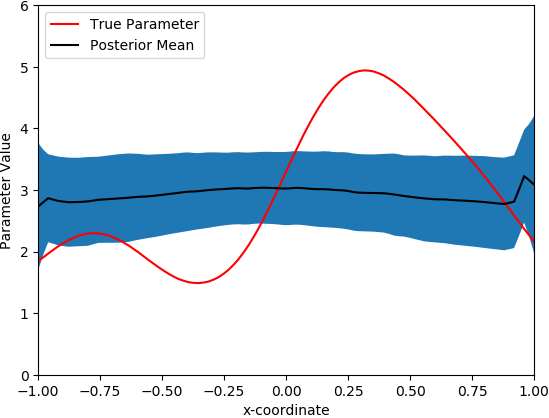}
    \includegraphics[scale=\figscale]{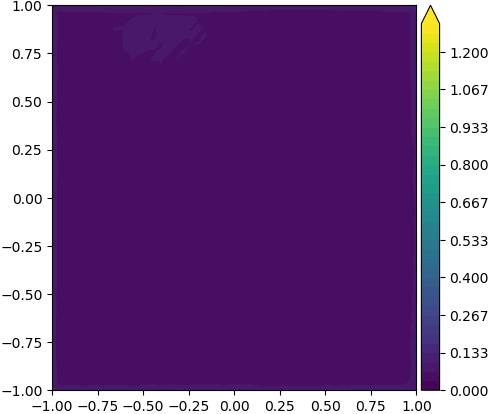}
    \includegraphics[scale=\figscale]{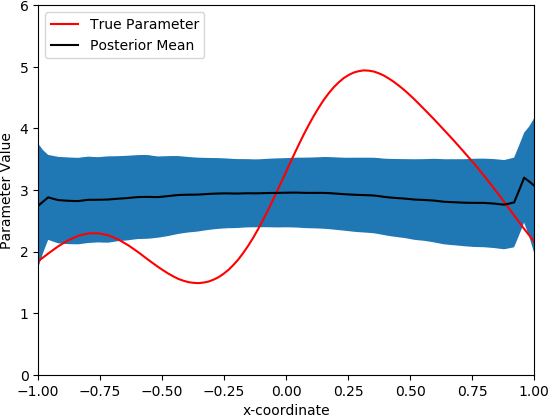}
    \includegraphics[scale=\figscale]{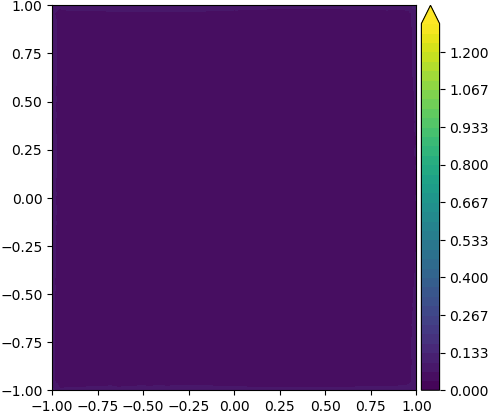}
    \caption{\scriptsize Top row left to right: mesh with sensors denoted with a red cross,
        true PoI,
        cross-sectional uncertainty estimate and pointwise posterior
        variance from Laplace approximation.
        Second to fourth rows: $\penJS =
        0.00001,0.001,0.1,0.5$. First and third columns: cross-sectional
        uncertainty estimates. Second and fourth columns: approximate pointwise
        posterior variance. First and second columns: exact PtO map.  Third and
        fourth columns: learned PtO map.}
        \label{FigPoisson2Dn2601nspt01M500}
\end{figure}

%------------------------------------------------------------------------------
\subsection{$\noiselevel=0.01$, $\numdatatrain=1000$} \label{SecHeat2Dn2601ns1M1000}
%------------------------------------------------------------------------------
\begin{table}[H]
    \scriptsize
    \centering
    \begin{tabular}{c | c | c | c |}
        \cline{2-4}
        & \multicolumn{2}{c|}{Relative Error: $\param$} & \multicolumn{1}{c|}{Relative Error: $\stateobs$}\\
        \hline
        \multicolumn{1}{|c|}{$\penJS$} & Exact PtO & Learned PtO & Learned PtO\\
        \hline
        \multicolumn{1}{|c|}{0.00001} & 21.99\% & 23.64\% & 19.98\% \\
        \multicolumn{1}{|c|}{0.001}   & 22.24\% & 21.78\% & 14.42\% \\
        \multicolumn{1}{|c|}{0.1}     & 22.84\% & 22.73\% & 20.28\% \\
        \multicolumn{1}{|c|}{0.5}     & 24.16\% & 24.04\% & 22.78\% \\
        \hline
    \end{tabular}
    \caption{\scriptsize Table displaying the relative errors for UQ-VAE.
            Relative error of MAP estimate: 25.11\%.}
    \label{TableRelativeErrorsn2601nspt01M1000}
\end{table}
\begin{figure}[H]
    \centering
    \includegraphics[scale=\figscale]{Figures/poisson_2d/general/mesh_n2601.png}
    \includegraphics[scale=\figscale]{Figures/poisson_2d/general/parameter_test_n2601_128.png}
    \includegraphics[scale=\figscale]{Figures/poisson_2d/nspt01/n2601/traditional/parameter_cross_section.png}
    \includegraphics[scale=\figscale]{Figures/poisson_2d/nspt01/n2601/traditional/posterior_covariance.png}\\
    \includegraphics[scale=\figscale]{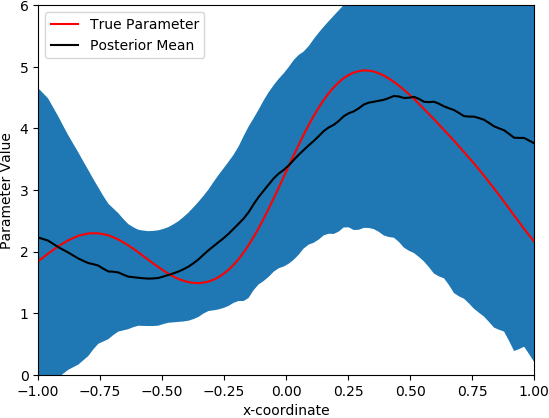}
    \includegraphics[scale=\figscale]{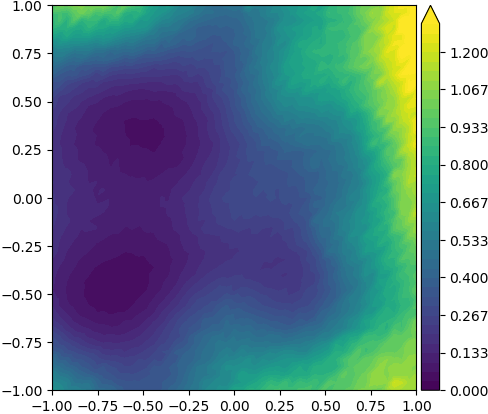}
    \includegraphics[scale=\figscale]{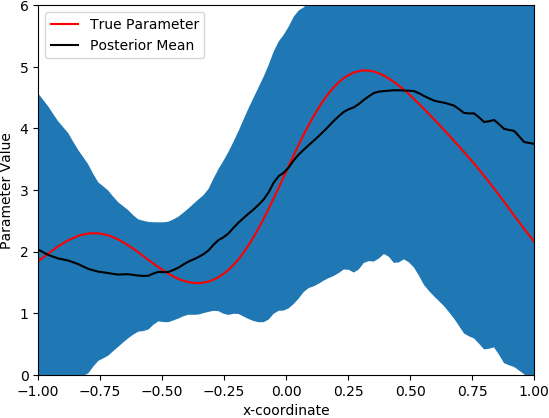}
    \includegraphics[scale=\figscale]{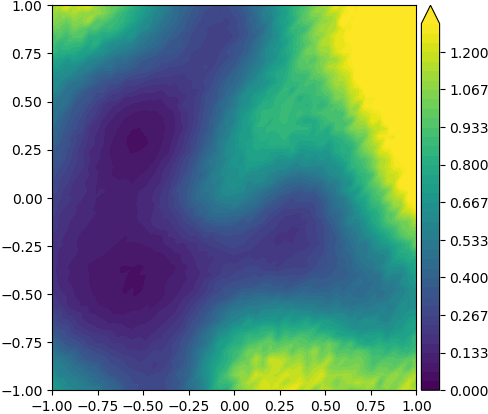}\\
    \includegraphics[scale=\figscale]{Figures/poisson_2d/nspt01/n2601/data1000/aug/jspt001/parameter_cross_section_128.png}
    \includegraphics[scale=\figscale]{Figures/poisson_2d/nspt01/n2601/data1000/aug/jspt001/posterior_covariance_128.png}
    \includegraphics[scale=\figscale]{Figures/poisson_2d/nspt01/n2601/data1000/aware/jspt001/parameter_cross_section_128.png}
    \includegraphics[scale=\figscale]{Figures/poisson_2d/nspt01/n2601/data1000/aware/jspt001/posterior_covariance_128.png}\\
    \includegraphics[scale=\figscale]{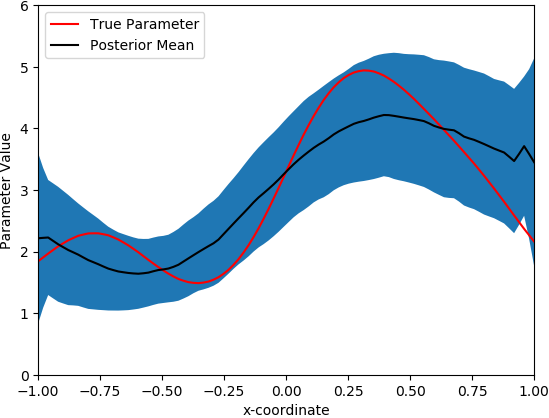}
    \includegraphics[scale=\figscale]{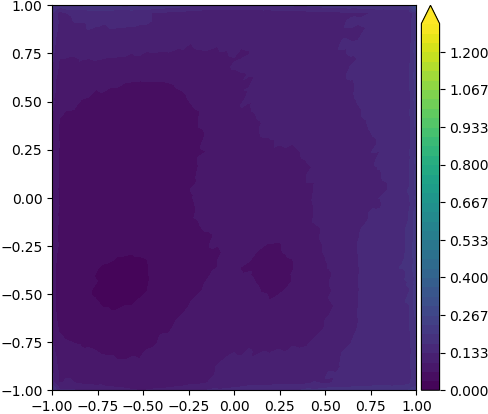}
    \includegraphics[scale=\figscale]{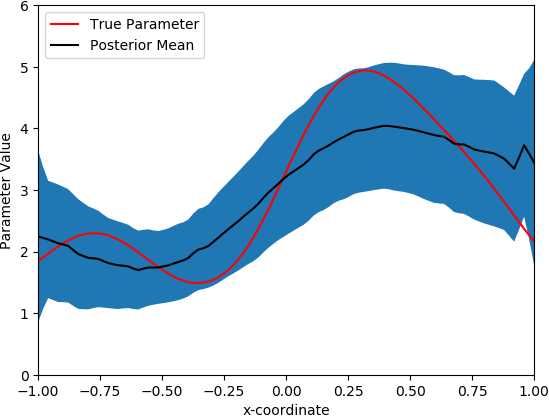}
    \includegraphics[scale=\figscale]{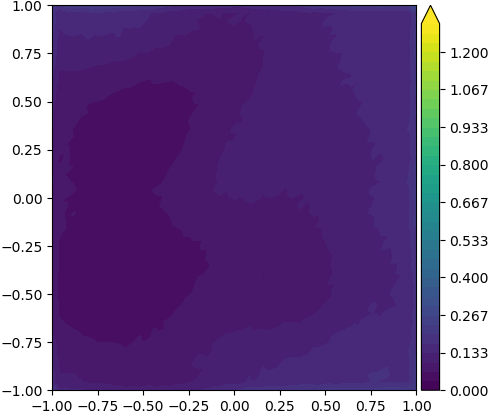}\\
    \includegraphics[scale=\figscale]{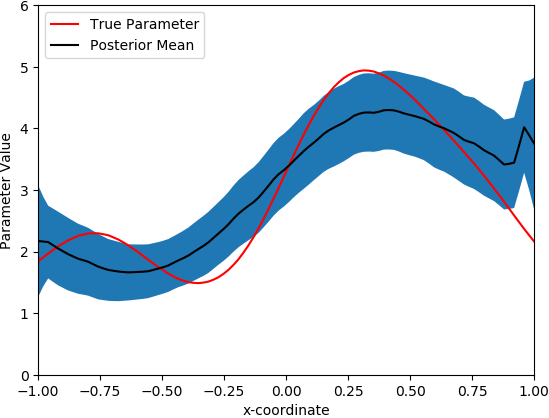}
    \includegraphics[scale=\figscale]{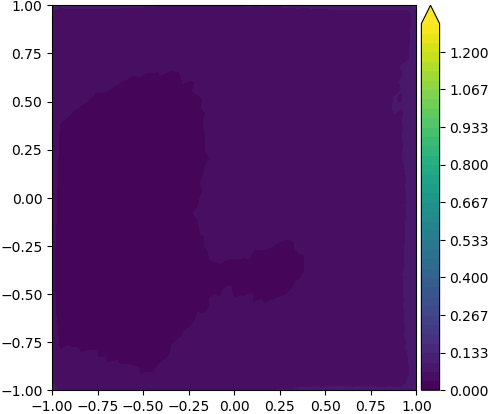}
    \includegraphics[scale=\figscale]{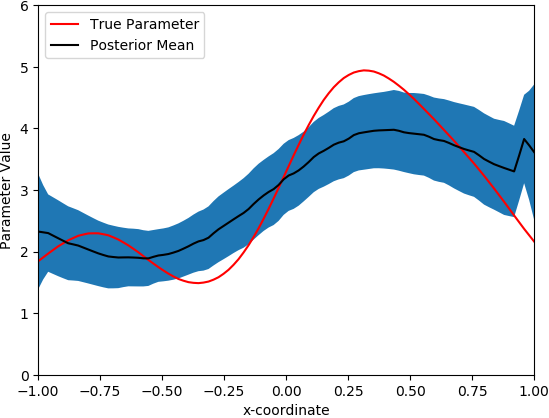}
    \includegraphics[scale=\figscale]{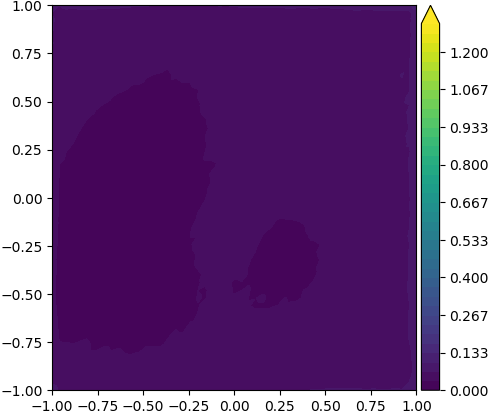}
    \caption{\scriptsize Top row left to right: mesh with sensors denoted with a red cross,
        true PoI,
        cross-sectional uncertainty estimate and pointwise posterior
        variance from Laplace approximation.
        Second to fourth rows: $\penJS =
        0.00001,0.001,0.1,0.5$. First and third columns: cross-sectional
        uncertainty estimates. Second and fourth columns: approximate pointwise
        posterior variance. First and second columns: exact PtO map.  Third and
        fourth columns: learned PtO map.}
        \label{FigPoisson2Dn2601nspt01M1000}
\end{figure}

%------------------------------------------------------------------------------
\subsection{$\noiselevel=0.01$, $\numdatatrain=5000$} \label{SecHeat2Dn2601ns1M5000}
%------------------------------------------------------------------------------
\begin{table}[H]
    \scriptsize
    \centering
    \begin{tabular}{c | c | c | c |}
        \cline{2-4}
        & \multicolumn{2}{c|}{Relative Error: $\param$} & \multicolumn{1}{c|}{Relative Error: $\stateobs$}\\
        \hline
        \multicolumn{1}{|c|}{$\penJS$} & Exact PtO & Learned PtO & Learned PtO\\
        \hline
        \multicolumn{1}{|c|}{0.00001} & 21.28\% & 21.14\% & 4.29\% \\
        \multicolumn{1}{|c|}{0.001}   & 21.34\% & 20.92\% & 3.79\% \\
        \multicolumn{1}{|c|}{0.1}     & 21.37\% & 21.47\% & 5.27\% \\
        \multicolumn{1}{|c|}{0.5}     & 21.72\% & 21.19\% & 3.57\% \\
        \hline
    \end{tabular}
    \caption{\scriptsize Table displaying the relative errors for UQ-VAE.
            Relative error of MAP estimate: 25.11\%.}
    \label{TableRelativeErrorsn2601nspt01M5000}
\end{table}
\begin{figure}[H]
    \centering
    \includegraphics[scale=\figscale]{Figures/poisson_2d/general/mesh_n2601.png}
    \includegraphics[scale=\figscale]{Figures/poisson_2d/general/parameter_test_n2601_128.png}
    \includegraphics[scale=\figscale]{Figures/poisson_2d/nspt01/n2601/traditional/parameter_cross_section.png}
    \includegraphics[scale=\figscale]{Figures/poisson_2d/nspt01/n2601/traditional/posterior_covariance.png}\\
    \includegraphics[scale=\figscale]{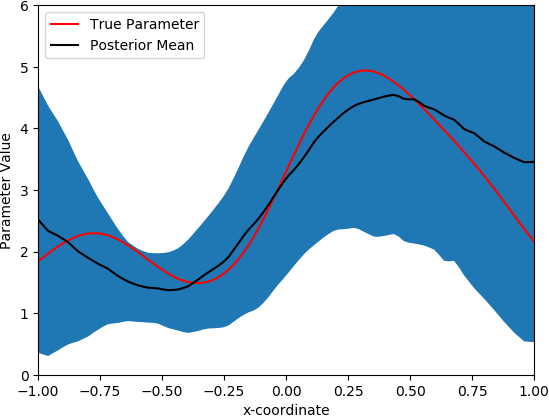}
    \includegraphics[scale=\figscale]{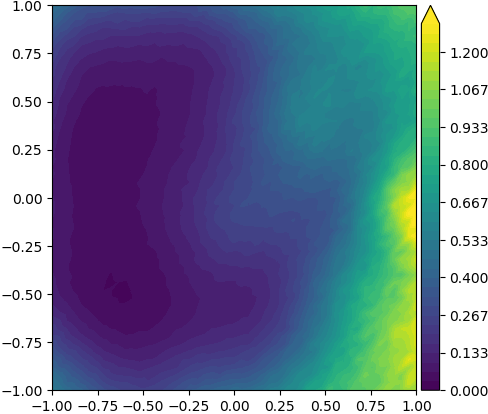}
    \includegraphics[scale=\figscale]{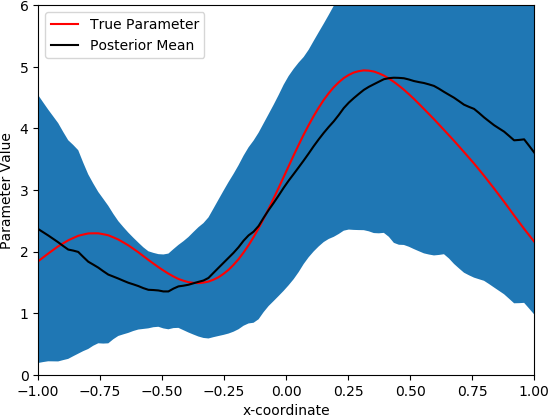}
    \includegraphics[scale=\figscale]{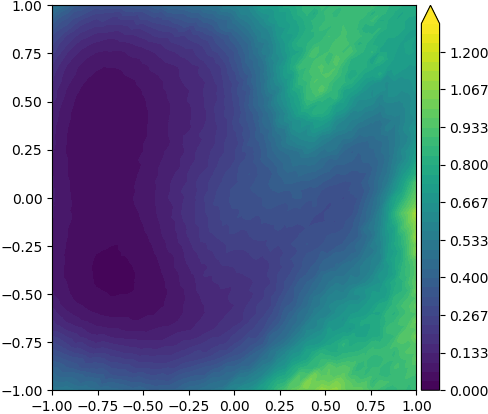}\\
    \includegraphics[scale=\figscale]{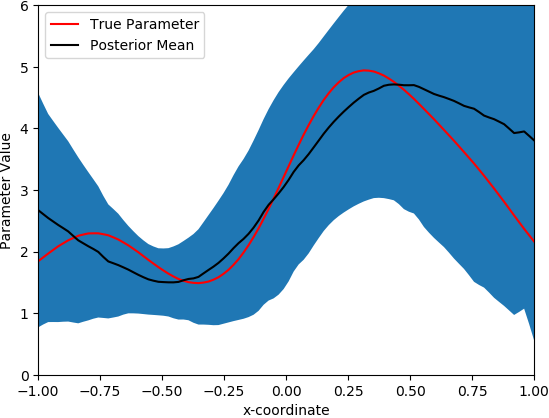}
    \includegraphics[scale=\figscale]{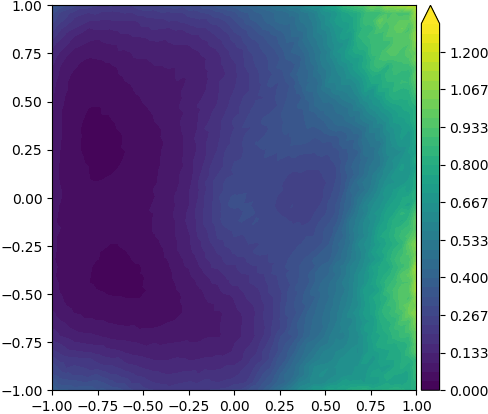}
    \includegraphics[scale=\figscale]{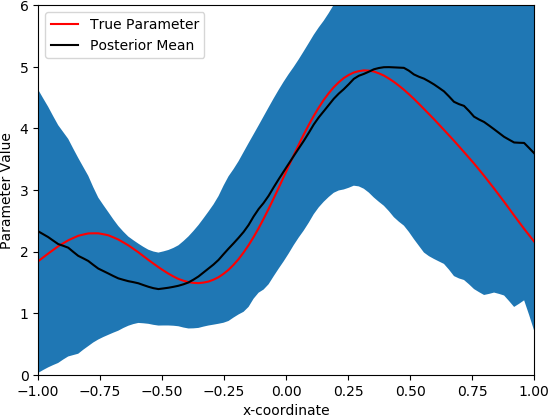}
    \includegraphics[scale=\figscale]{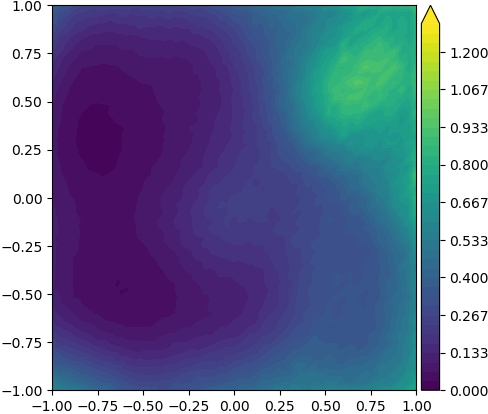}\\
    \includegraphics[scale=\figscale]{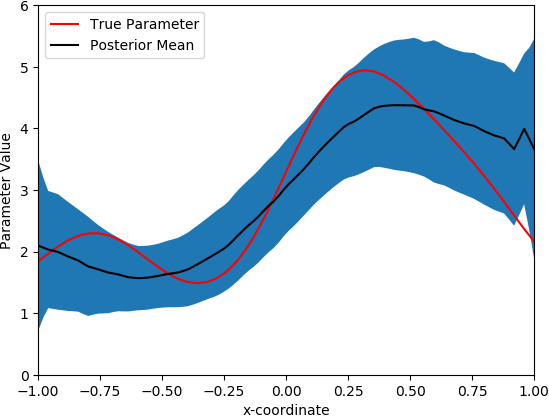}
    \includegraphics[scale=\figscale]{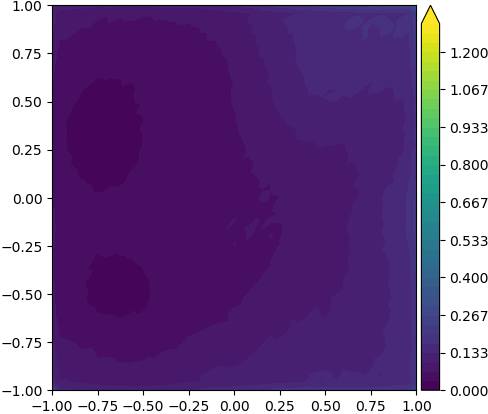}
    \includegraphics[scale=\figscale]{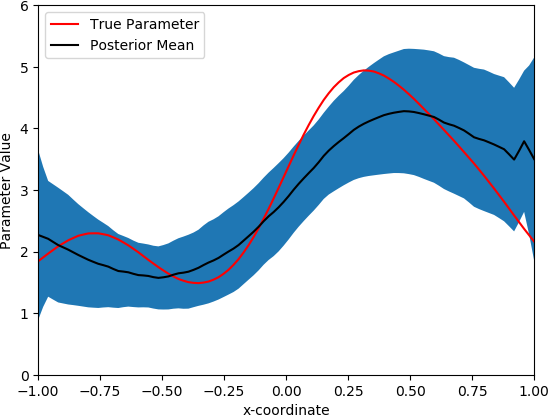}
    \includegraphics[scale=\figscale]{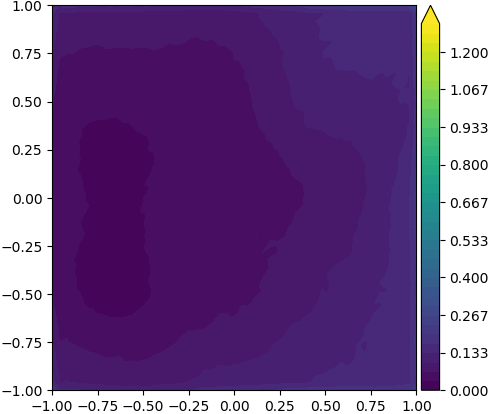}\\
    \includegraphics[scale=\figscale]{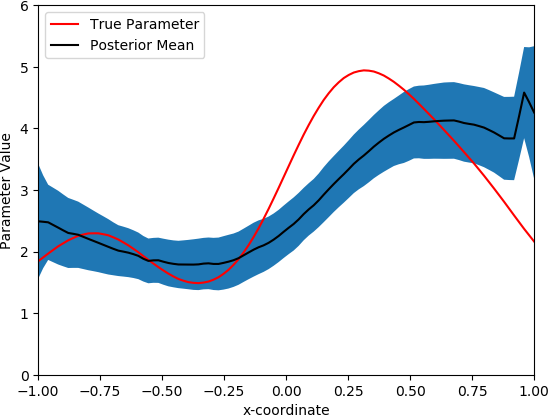}
    \includegraphics[scale=\figscale]{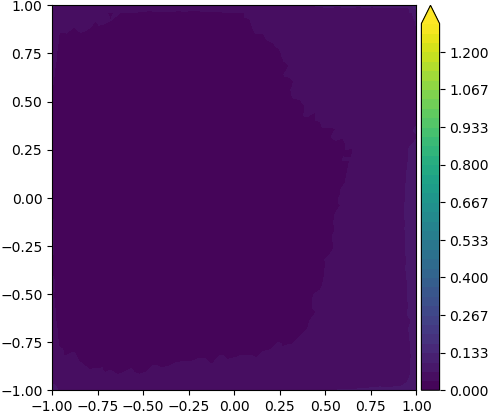}
    \includegraphics[scale=\figscale]{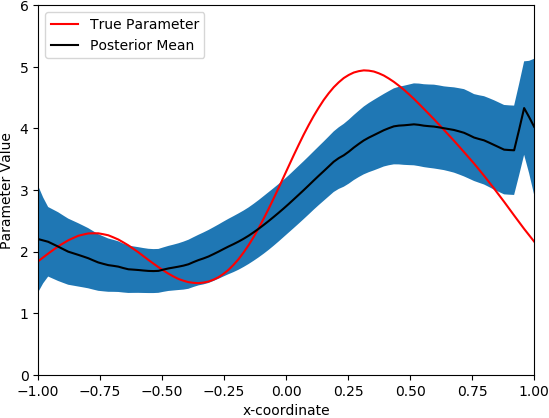}
    \includegraphics[scale=\figscale]{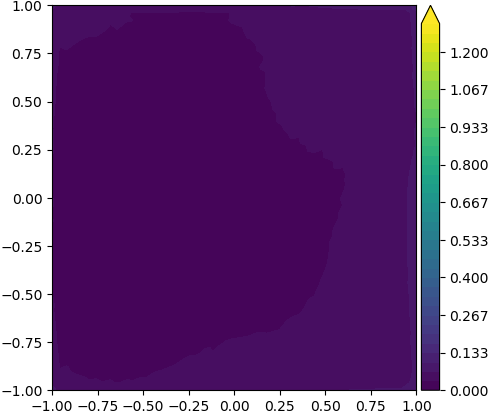}
    \caption{\scriptsize Top row left to right: mesh with sensors denoted with a red cross,
        true PoI,
        cross-sectional uncertainty estimate and pointwise posterior
        variance from Laplace approximation.
        Second to fourth rows: $\penJS =
        0.00001,0.001,0.1,0.5$. First and third columns: cross-sectional
        uncertainty estimates. Second and fourth columns: approximate pointwise
        posterior variance. First and second columns: exact PtO map.  Third and
        fourth columns: learned PtO map.}
        \label{FigPoisson2Dn2601nspt01M5000}
\end{figure}

%------------------------------------------------------------------------------
\subsection{$\noiselevel=0.05$, $\numdatatrain=50$} \label{SecHeat2Dn2601ns5M50}
%------------------------------------------------------------------------------
\begin{table}[H]
    \scriptsize
    \centering
    \begin{tabular}{c | c | c | c |}
        \cline{2-4}
        & \multicolumn{2}{c|}{Relative Error: $\param$} & \multicolumn{1}{c|}{Relative Error: $\stateobs$}\\
        \hline
        \multicolumn{1}{|c|}{$\penJS$} & Exact PtO & Learned PtO & Learned PtO\\
        \hline
        \multicolumn{1}{|c|}{0.00001} & 31.76\% & 33.37\% & 22.32\% \\
        \multicolumn{1}{|c|}{0.001}   & 34.78\% & 34.52\% & 22.24\% \\
        \multicolumn{1}{|c|}{0.1}     & 33.45\% & 31.00\% & 39.73\% \\
        \multicolumn{1}{|c|}{0.5}     & 32.87\% & 33.00\% & 38.23\% \\
        \hline
    \end{tabular}
    \caption{\scriptsize Table displaying the relative errors for UQ-VAE.
            Relative error of MAP estimate: 43.19\%.}
    \label{TableRelativeErrorsn2601nspt05M50}
\end{table}
\begin{figure}[H]
    \centering
    \includegraphics[scale=\figscale]{Figures/poisson_2d/general/mesh_n2601.png}
    \includegraphics[scale=\figscale]{Figures/poisson_2d/general/parameter_test_n2601_128.png}
    \includegraphics[scale=\figscale]{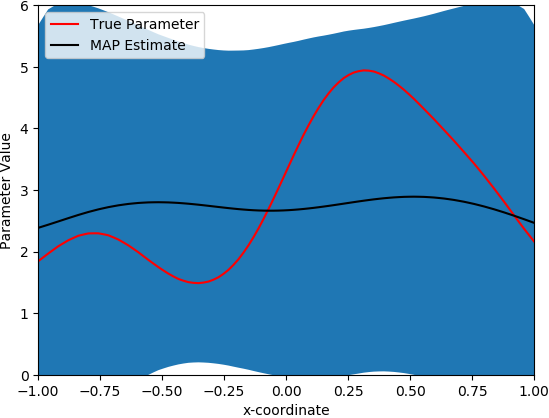}
    \includegraphics[scale=\figscale]{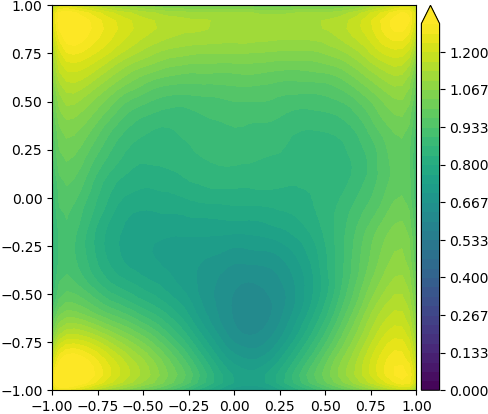}\\
    \includegraphics[scale=\figscale]{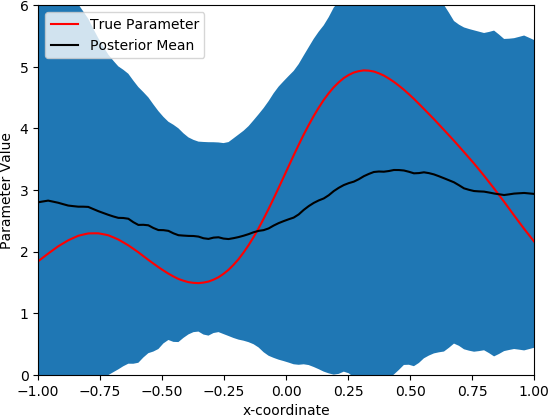}
    \includegraphics[scale=\figscale]{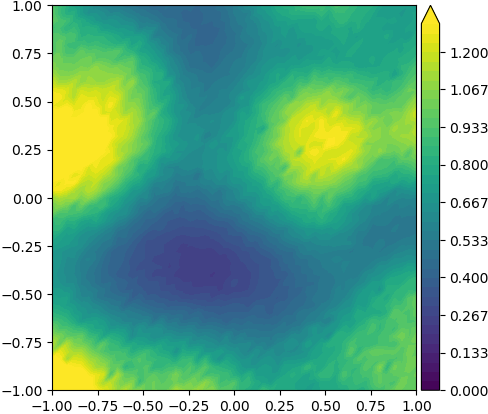}
    \includegraphics[scale=\figscale]{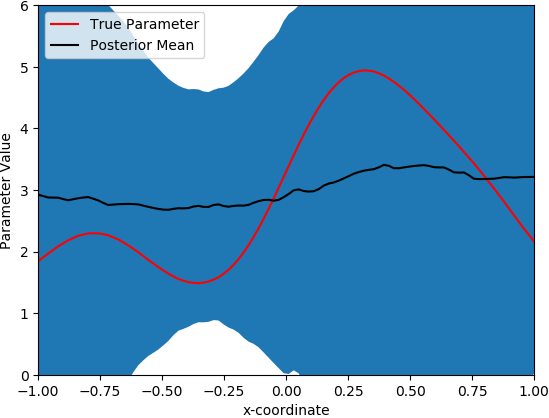}
    \includegraphics[scale=\figscale]{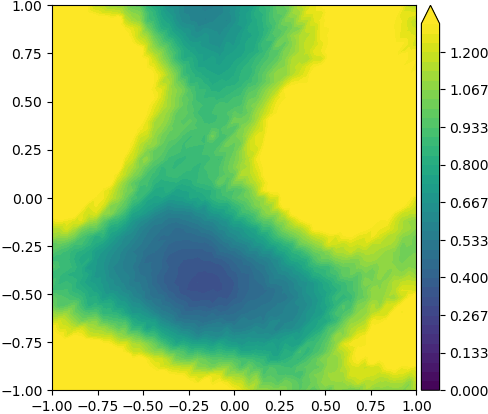}\\
    \includegraphics[scale=\figscale]{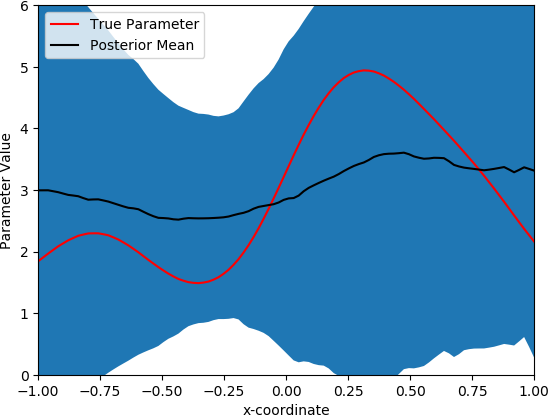}
    \includegraphics[scale=\figscale]{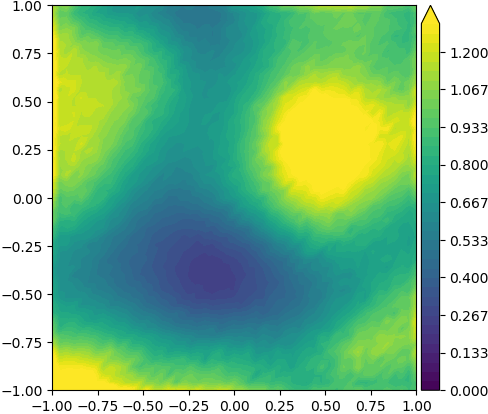}
    \includegraphics[scale=\figscale]{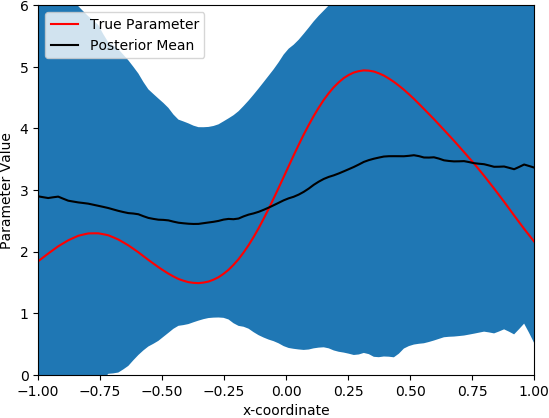}
    \includegraphics[scale=\figscale]{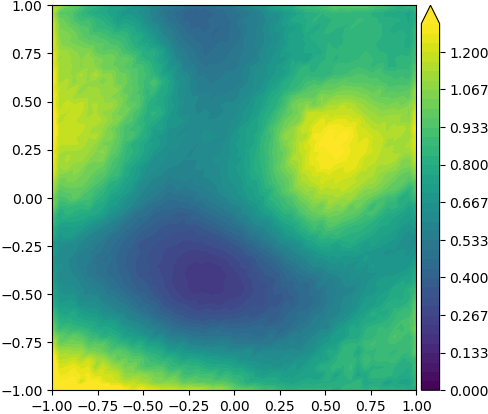}\\
    \includegraphics[scale=\figscale]{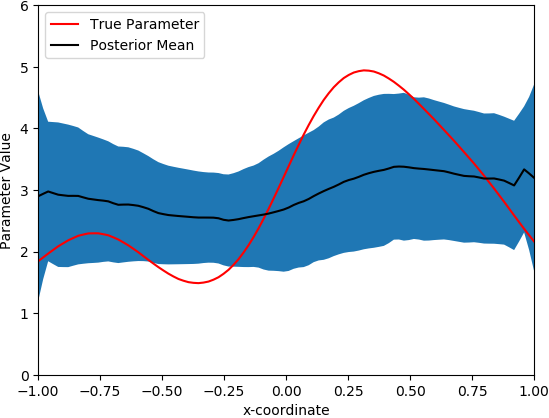}
    \includegraphics[scale=\figscale]{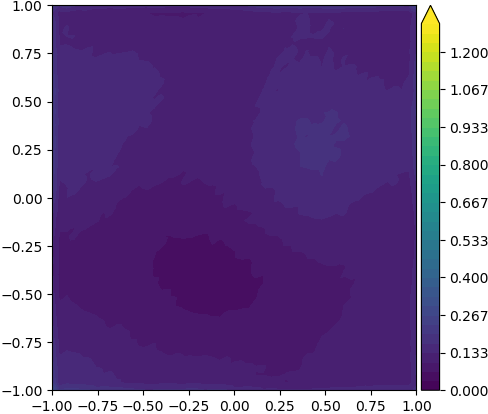}
    \includegraphics[scale=\figscale]{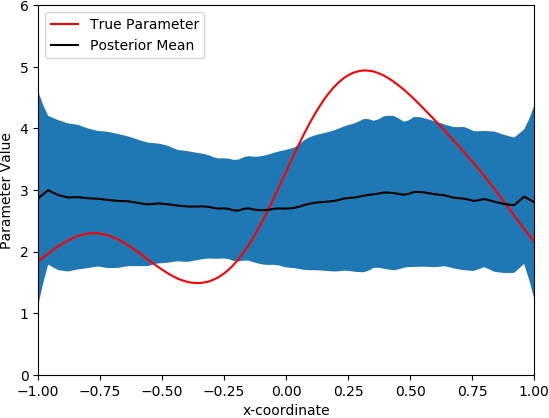}
    \includegraphics[scale=\figscale]{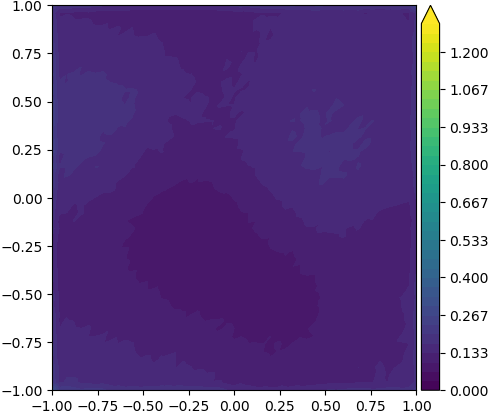}\\
    \includegraphics[scale=\figscale]{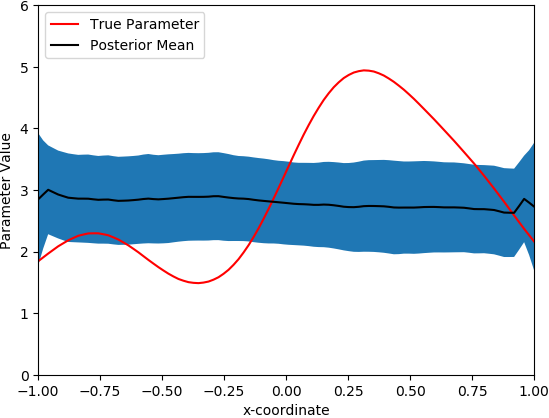}
    \includegraphics[scale=\figscale]{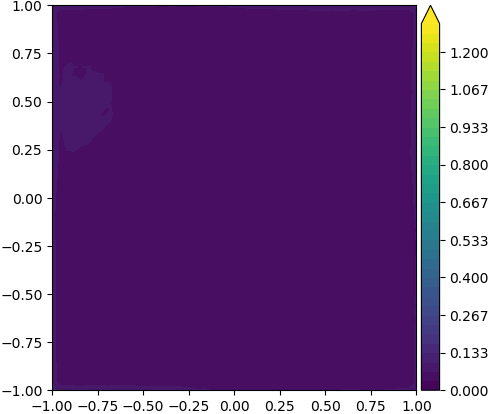}
    \includegraphics[scale=\figscale]{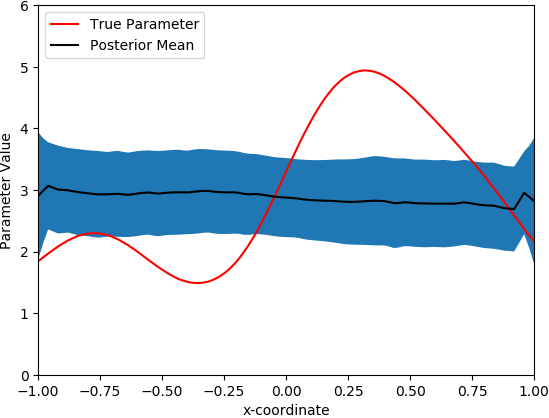}
    \includegraphics[scale=\figscale]{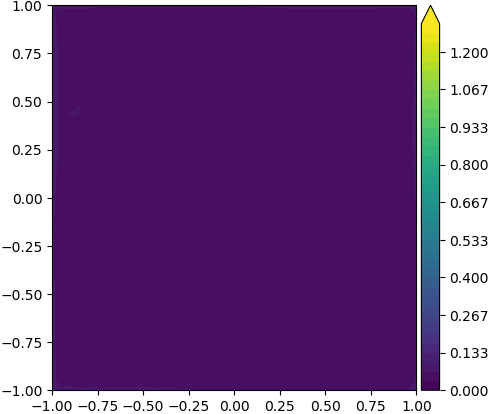}
    \caption{\scriptsize Top row left to right: mesh with sensors denoted with a red cross,
        true PoI,
        cross-sectional uncertainty estimate and pointwise posterior
        variance from Laplace approximation.
        Second to fourth rows: $\penJS =
        0.00001,0.001,0.1,0.5$. First and third columns: cross-sectional
        uncertainty estimates. Second and fourth columns: approximate pointwise
        posterior variance. First and second columns: exact PtO map.  Third and
        fourth columns: learned PtO map.}
        \label{FigPoisson2Dn2601nspt05M50}
\end{figure}

%------------------------------------------------------------------------------
\subsection{$\noiselevel=0.05$, $\numdatatrain=500$} \label{SecHeat2Dn2601ns5M500}
%------------------------------------------------------------------------------
\begin{table}[H]
    \scriptsize
    \centering
    \begin{tabular}{c | c | c | c |}
        \cline{2-4}
        & \multicolumn{2}{c|}{Relative Error: $\param$} & \multicolumn{1}{c|}{Relative Error: $\stateobs$}\\
        \hline
        \multicolumn{1}{|c|}{$\penJS$} & Exact PtO & Learned PtO & Learned PtO\\
        \hline
        \multicolumn{1}{|c|}{0.00001} & 31.72\% & 30.05\% & 21.22\% \\
        \multicolumn{1}{|c|}{0.001}   & 29.79\% & 30.69\% & 25.16\% \\
        \multicolumn{1}{|c|}{0.1}     & 30.31\% & 30.61\% & 26.02\% \\
        \multicolumn{1}{|c|}{0.5}     & 31.56\% & 31.33\% & 38.14\% \\
        \hline
    \end{tabular}
    \caption{\scriptsize Table displaying the relative errors for UQ-VAE.
            Relative error of MAP estimate: 43.19\%.}
    \label{TableRelativeErrorsn2601nspt05M500}
\end{table}
\begin{figure}[H]
    \centering
    \includegraphics[scale=\figscale]{Figures/poisson_2d/general/mesh_n2601.png}
    \includegraphics[scale=\figscale]{Figures/poisson_2d/general/parameter_test_n2601_128.png}
    \includegraphics[scale=\figscale]{Figures/poisson_2d/nspt05/n2601/traditional/parameter_cross_section.png}
    \includegraphics[scale=\figscale]{Figures/poisson_2d/nspt05/n2601/traditional/posterior_covariance.png}\\
    \includegraphics[scale=\figscale]{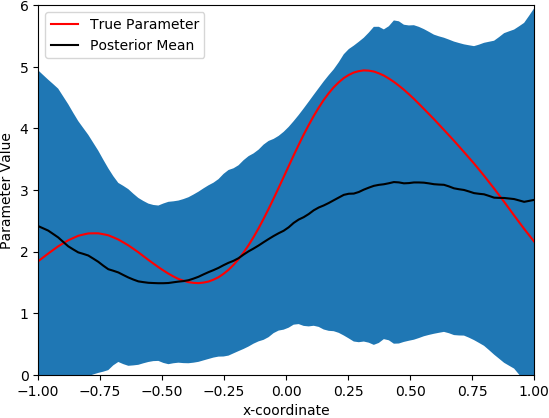}
    \includegraphics[scale=\figscale]{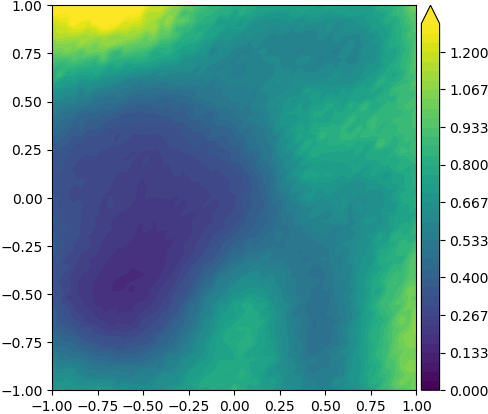}
    \includegraphics[scale=\figscale]{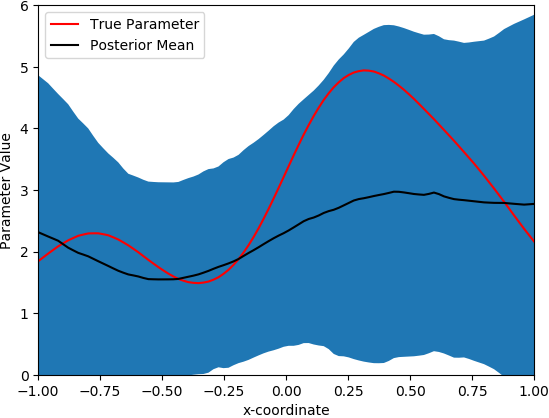}
    \includegraphics[scale=\figscale]{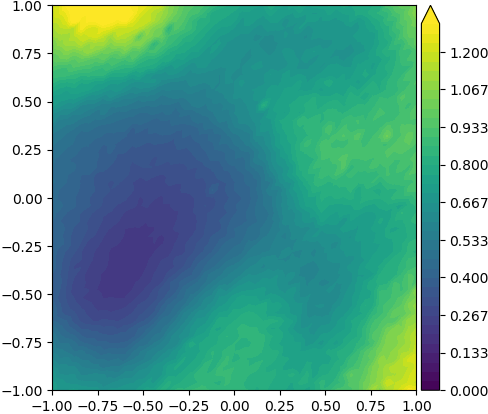}\\
    \includegraphics[scale=\figscale]{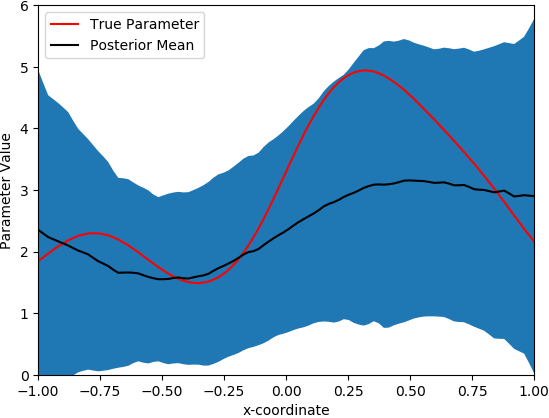}
    \includegraphics[scale=\figscale]{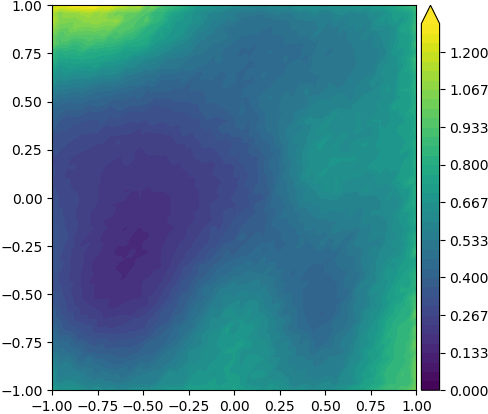}
    \includegraphics[scale=\figscale]{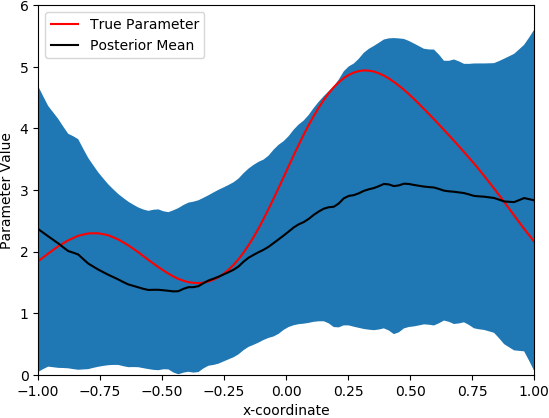}
    \includegraphics[scale=\figscale]{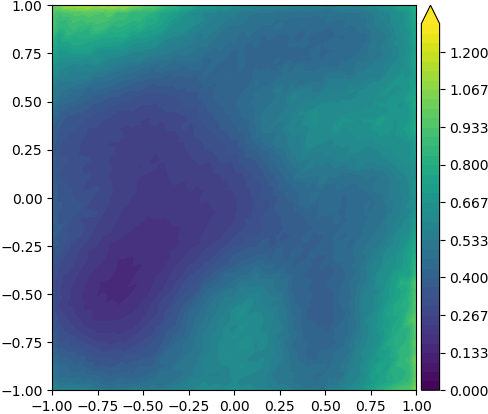}\\
    \includegraphics[scale=\figscale]{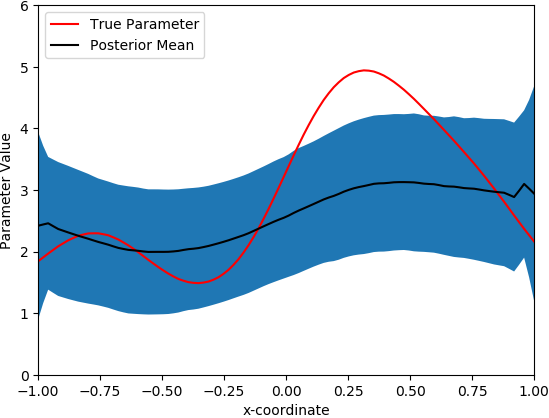}
    \includegraphics[scale=\figscale]{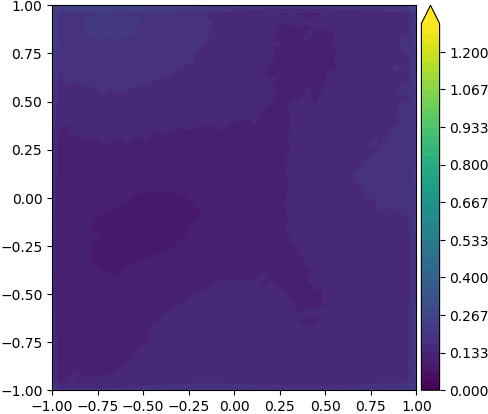}
    \includegraphics[scale=\figscale]{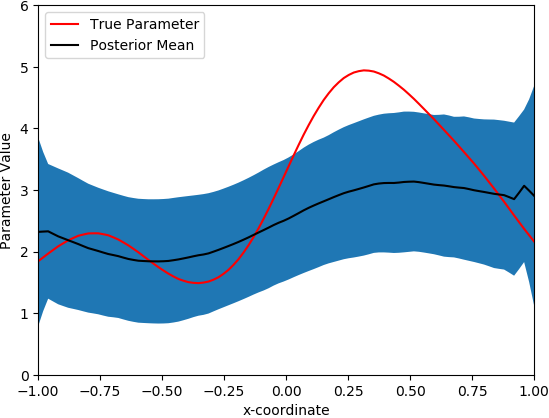}
    \includegraphics[scale=\figscale]{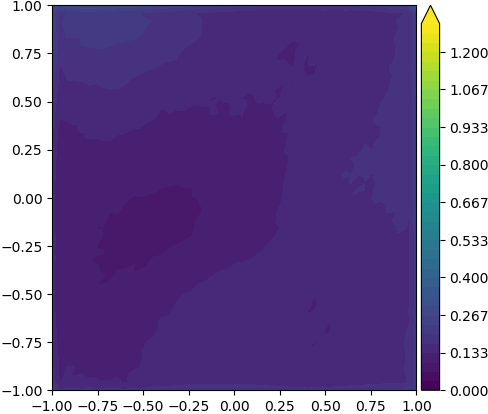}\\
    \includegraphics[scale=\figscale]{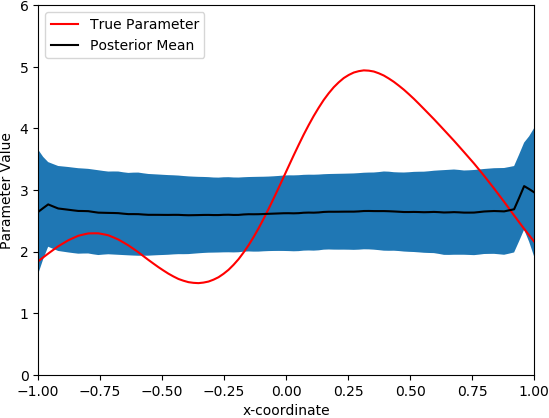}
    \includegraphics[scale=\figscale]{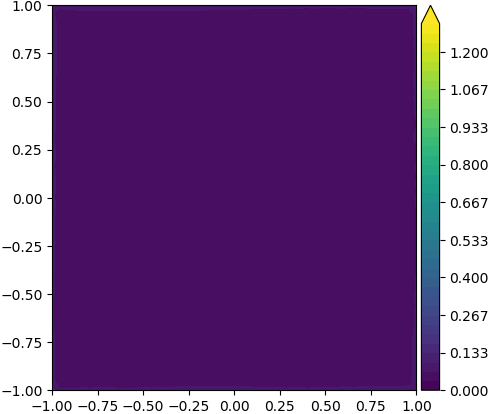}
    \includegraphics[scale=\figscale]{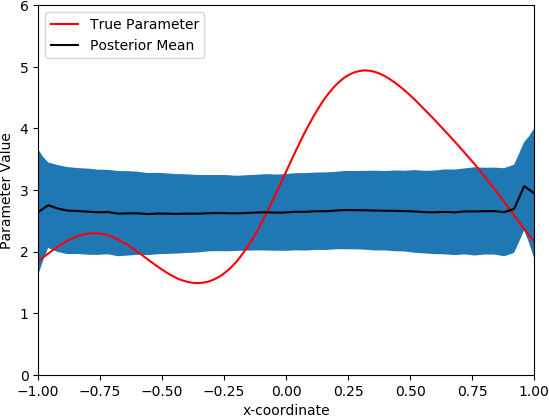}
    \includegraphics[scale=\figscale]{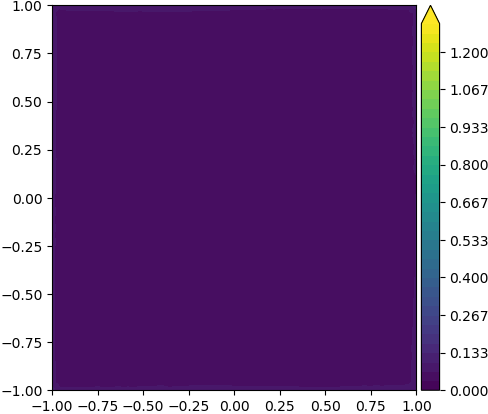}
    \caption{\scriptsize Top row left to right: mesh with sensors denoted with a red cross,
        true PoI,
        cross-sectional uncertainty estimate and pointwise posterior
        variance from Laplace approximation.
        Second to fourth rows: $\penJS =
        0.00001,0.001,0.1,0.5$. First and third columns: cross-sectional
        uncertainty estimates. Second and fourth columns: approximate pointwise
        posterior variance. First and second columns: exact PtO map.  Third and
        fourth columns: learned PtO map.}
        \label{FigPoisson2Dn2601nspt05M500}
\end{figure}

%------------------------------------------------------------------------------
\subsection{$\noiselevel=0.05$, $\numdatatrain=1000$} \label{SecHeat2Dn2601ns5M1000}
%------------------------------------------------------------------------------
\begin{table}[H]
    \scriptsize
    \centering
    \begin{tabular}{c | c | c | c |}
        \cline{2-4}
        & \multicolumn{2}{c|}{Relative Error: $\param$} & \multicolumn{1}{c|}{Relative Error: $\stateobs$}\\
        \hline
        \multicolumn{1}{|c|}{$\penJS$} & Exact PtO & Learned PtO & Learned PtO\\
        \hline
        \multicolumn{1}{|c|}{0.00001} & 31.96\% & 32.59\% & 18.36\% \\
        \multicolumn{1}{|c|}{0.001}   & 32.14\% & 31.85\% & 20.17\% \\
        \multicolumn{1}{|c|}{0.1}     & 31.69\% & 30.87\% & 25.05\% \\
        \multicolumn{1}{|c|}{0.5}     & 33.31\% & 33.09\% & 28.05\% \\
        \hline
    \end{tabular}
    \caption{\scriptsize Table displaying the relative errors for UQ-VAE.
            Relative error of MAP estimate: 43.19\%.}
    \label{TableRelativeErrorsn2601nspt05M1000}
\end{table}
\begin{figure}[H]
    \centering
    \includegraphics[scale=\figscale]{Figures/poisson_2d/general/mesh_n2601.png}
    \includegraphics[scale=\figscale]{Figures/poisson_2d/general/parameter_test_n2601_128.png}
    \includegraphics[scale=\figscale]{Figures/poisson_2d/nspt05/n2601/traditional/parameter_cross_section.png}
    \includegraphics[scale=\figscale]{Figures/poisson_2d/nspt05/n2601/traditional/posterior_covariance.png}\\
    \includegraphics[scale=\figscale]{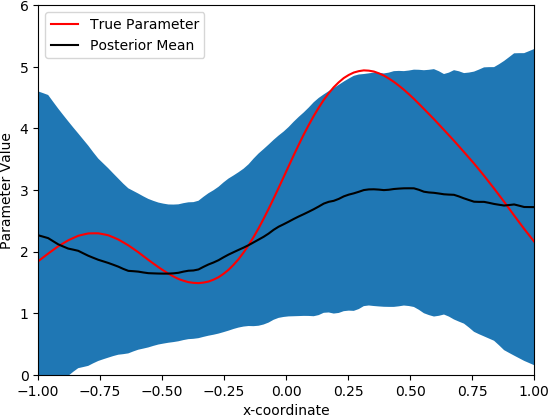}
    \includegraphics[scale=\figscale]{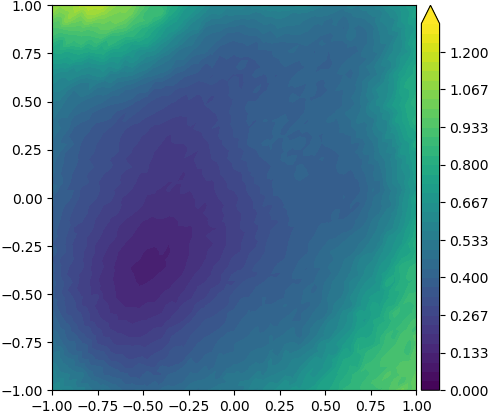}
    \includegraphics[scale=\figscale]{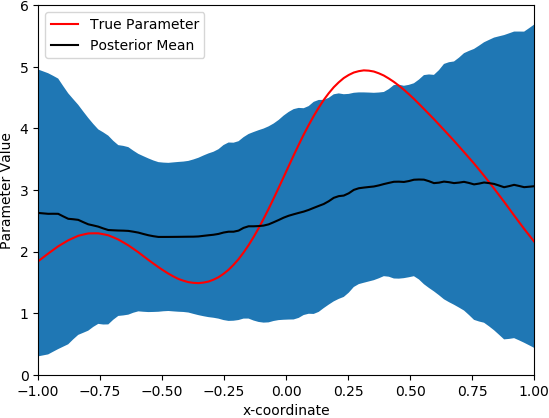}
    \includegraphics[scale=\figscale]{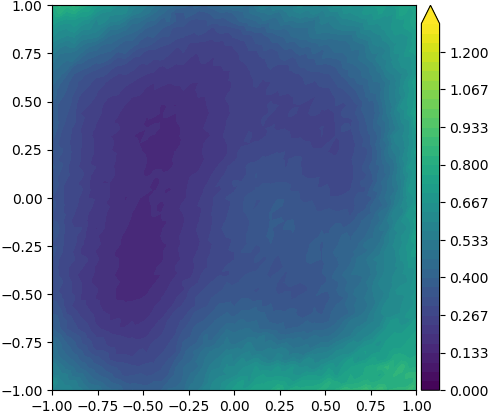}\\
    \includegraphics[scale=\figscale]{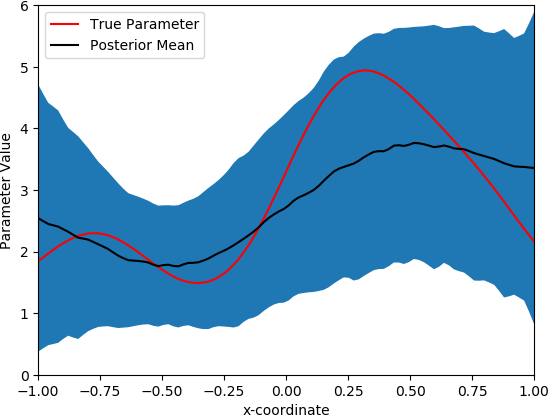}
    \includegraphics[scale=\figscale]{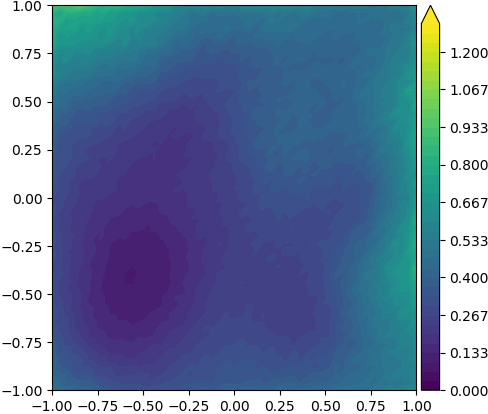}
    \includegraphics[scale=\figscale]{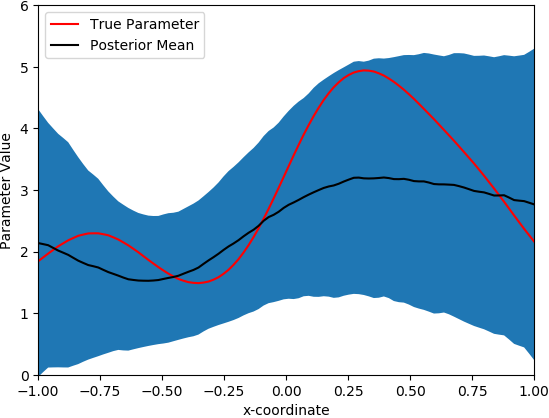}
    \includegraphics[scale=\figscale]{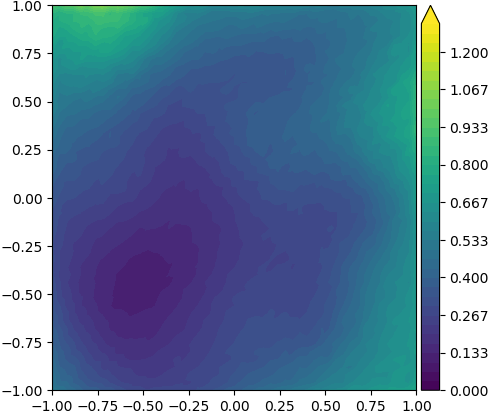}\\
    \includegraphics[scale=\figscale]{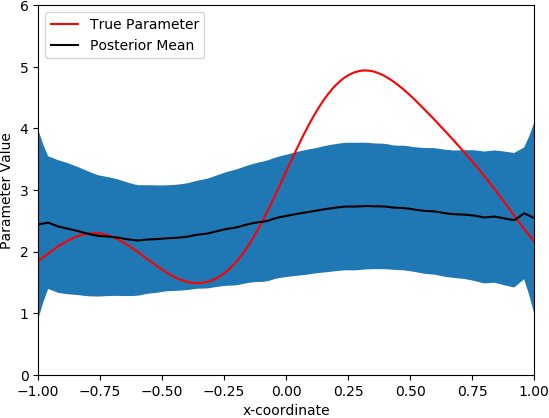}
    \includegraphics[scale=\figscale]{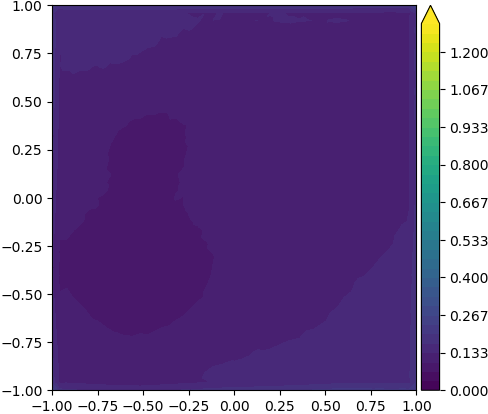}
    \includegraphics[scale=\figscale]{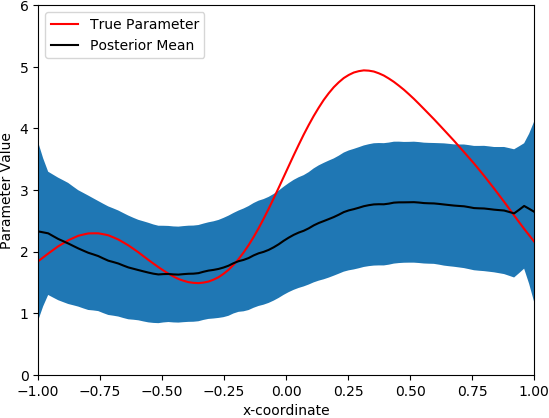}
    \includegraphics[scale=\figscale]{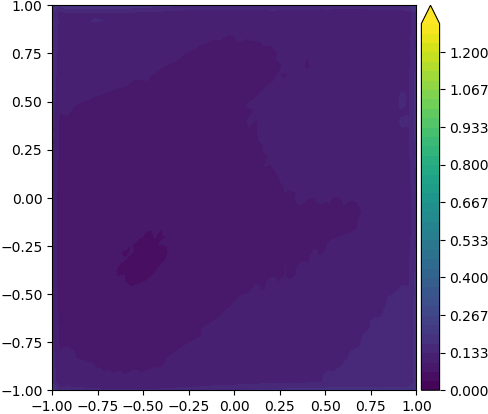}\\
    \includegraphics[scale=\figscale]{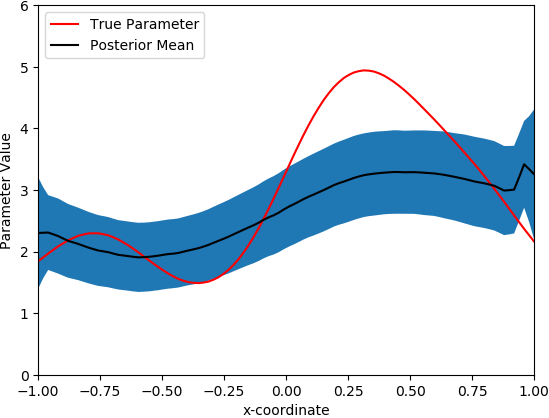}
    \includegraphics[scale=\figscale]{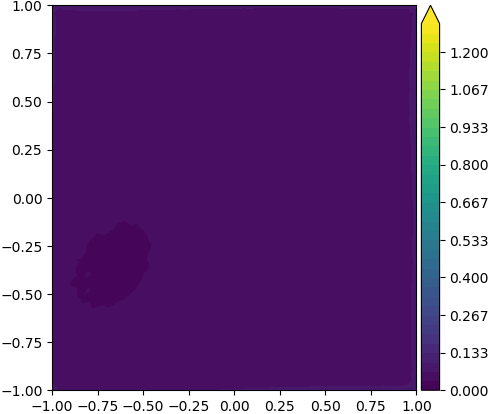}
    \includegraphics[scale=\figscale]{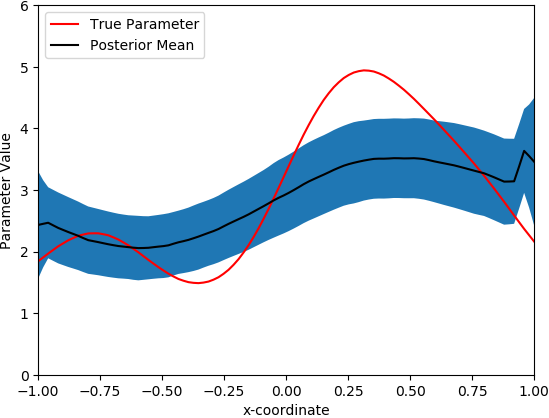}
    \includegraphics[scale=\figscale]{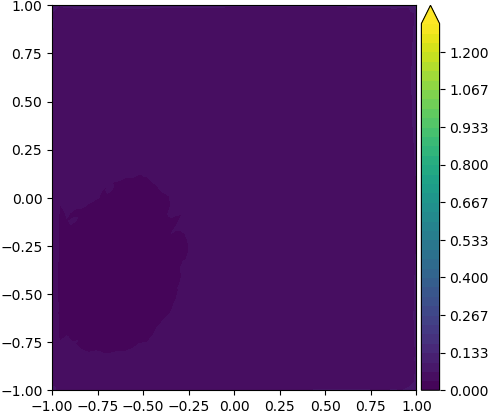}
    \caption{\scriptsize Top row left to right: mesh with sensors denoted with a red cross,
        true PoI,
        cross-sectional uncertainty estimate and pointwise posterior
        variance from Laplace approximation.
        Second to fourth rows: $\penJS =
        0.00001,0.001,0.1,0.5$. First and third columns: cross-sectional
        uncertainty estimates. Second and fourth columns: approximate pointwise
        posterior variance. First and second columns: exact PtO map.  Third and
        fourth columns: learned PtO map.}
        \label{FigPoisson2Dn2601nspt05M1000}
\end{figure}

%------------------------------------------------------------------------------
\subsection{$\noiselevel=0.05$, $\numdatatrain=5000$} \label{SecHeat2Dn2601ns5M5000}
%------------------------------------------------------------------------------
\begin{table}[H]
    \scriptsize
    \centering
    \begin{tabular}{c | c | c | c |}
        \cline{2-4}
        & \multicolumn{2}{c|}{Relative Error: $\param$} & \multicolumn{1}{c|}{Relative Error: $\stateobs$}\\
        \hline
        \multicolumn{1}{|c|}{$\penJS$} & Exact PtO & Learned PtO & Learned PtO\\
        \hline
        \multicolumn{1}{|c|}{0.00001} & 36.78\% & 35.41\% & 8.11\% \\
        \multicolumn{1}{|c|}{0.001}   & 36.90\% & 35.82\% & 8.10\% \\
        \multicolumn{1}{|c|}{0.1}     & 37.84\% & 36.48\% & 7.94\% \\
        \multicolumn{1}{|c|}{0.5}     & 36.88\% & 36.89\% & 9.41\% \\
        \hline
    \end{tabular}
    \caption{\scriptsize Table displaying the relative errors for UQ-VAE.
            Relative error of MAP estimate: 43.19\%.}
    \label{TableRelativeErrorsn2601nspt05M5000}
\end{table}
\begin{figure}[H]
    \centering
    \includegraphics[scale=\figscale]{Figures/poisson_2d/general/mesh_n2601.png}
    \includegraphics[scale=\figscale]{Figures/poisson_2d/general/parameter_test_n2601_128.png}
    \includegraphics[scale=\figscale]{Figures/poisson_2d/nspt05/n2601/traditional/parameter_cross_section.png}
    \includegraphics[scale=\figscale]{Figures/poisson_2d/nspt05/n2601/traditional/posterior_covariance.png}\\
    \includegraphics[scale=\figscale]{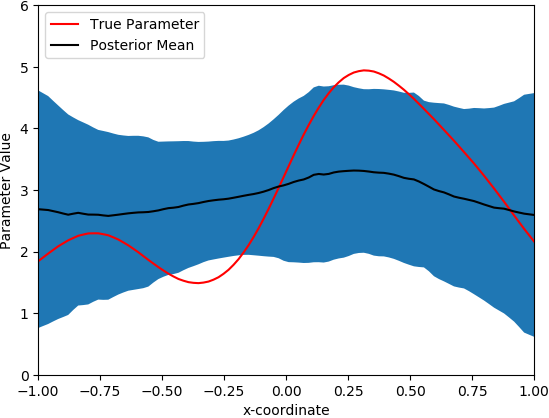}
    \includegraphics[scale=\figscale]{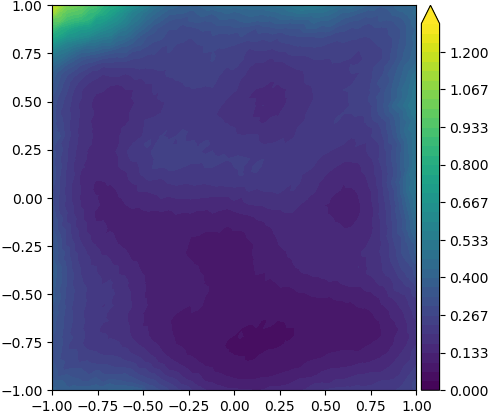}
    \includegraphics[scale=\figscale]{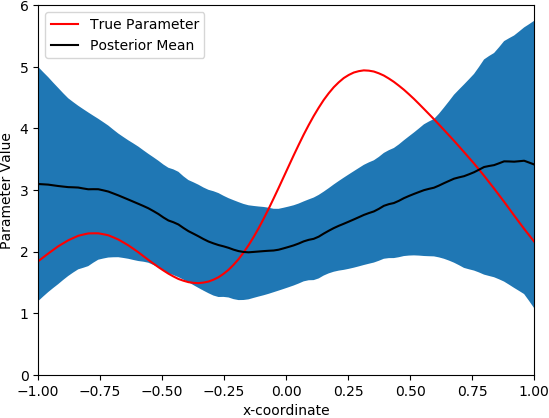}
    \includegraphics[scale=\figscale]{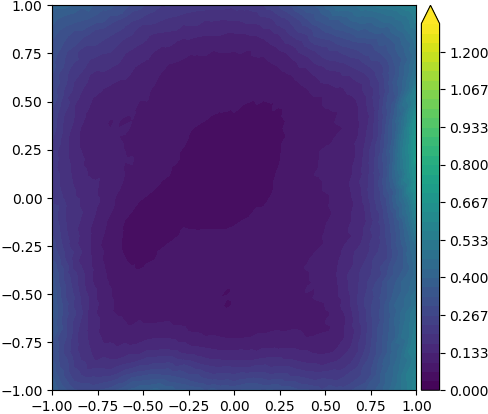}\\
    \includegraphics[scale=\figscale]{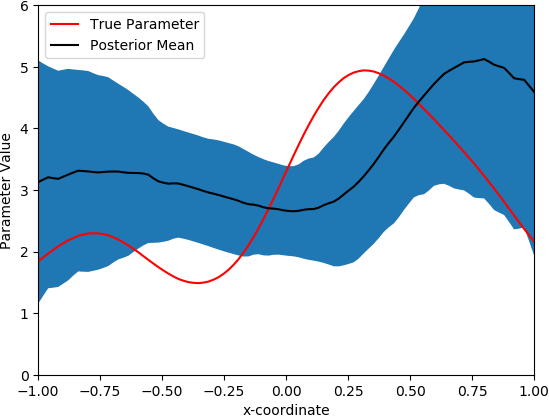}
    \includegraphics[scale=\figscale]{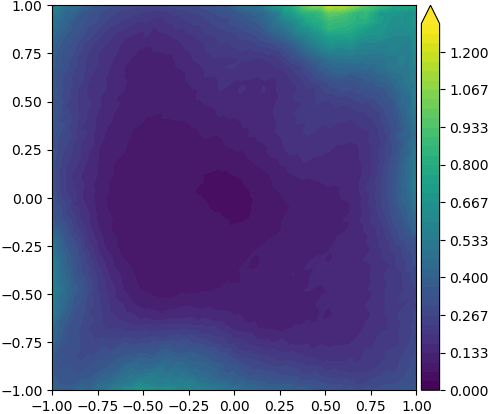}
    \includegraphics[scale=\figscale]{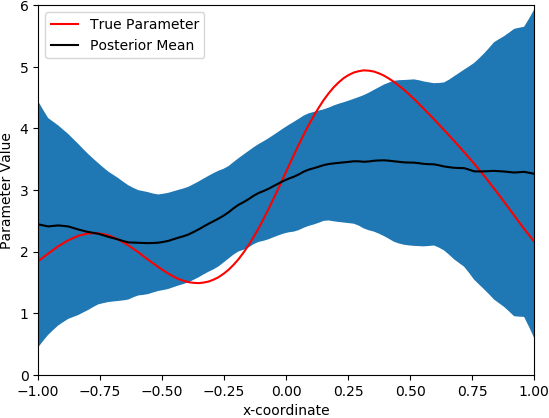}
    \includegraphics[scale=\figscale]{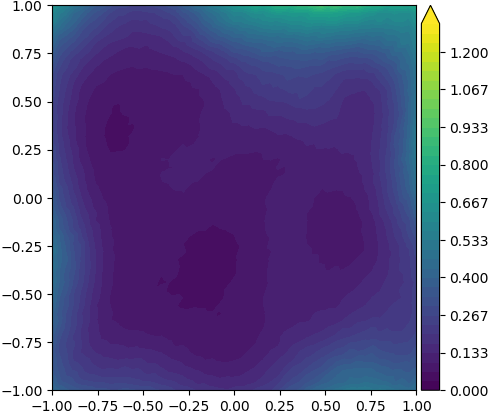}\\
    \includegraphics[scale=\figscale]{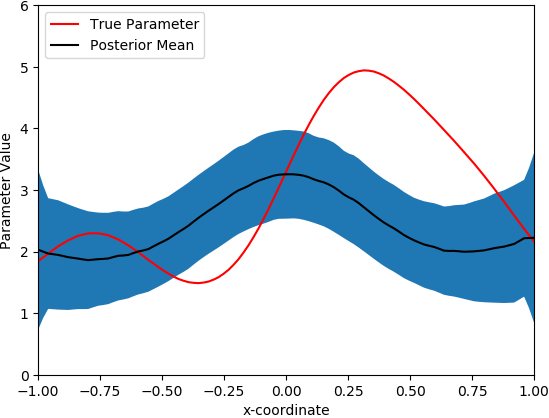}
    \includegraphics[scale=\figscale]{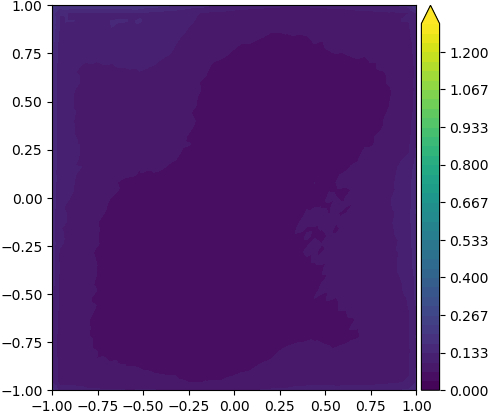}
    \includegraphics[scale=\figscale]{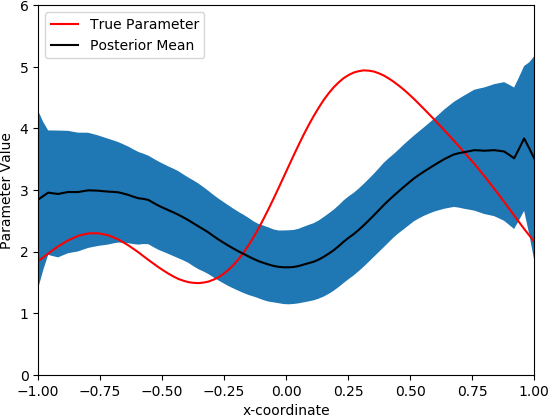}
    \includegraphics[scale=\figscale]{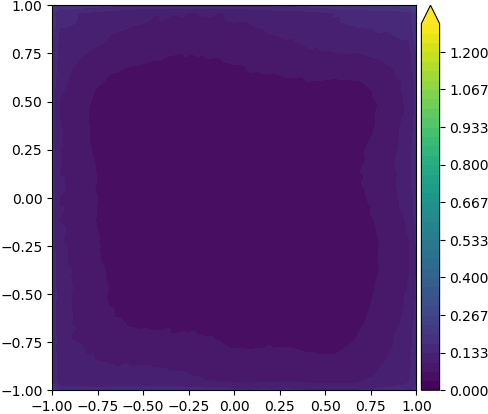}\\
    \includegraphics[scale=\figscale]{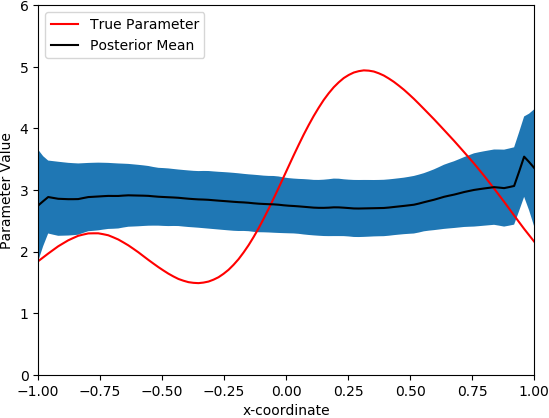}
    \includegraphics[scale=\figscale]{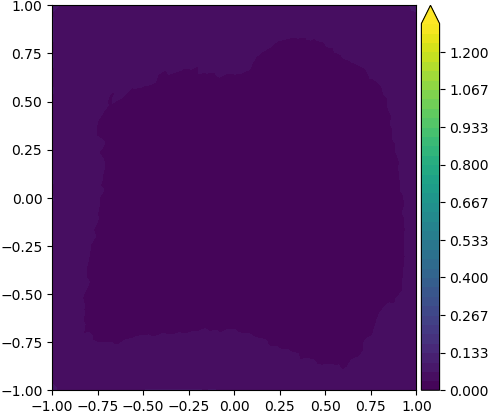}
    \includegraphics[scale=\figscale]{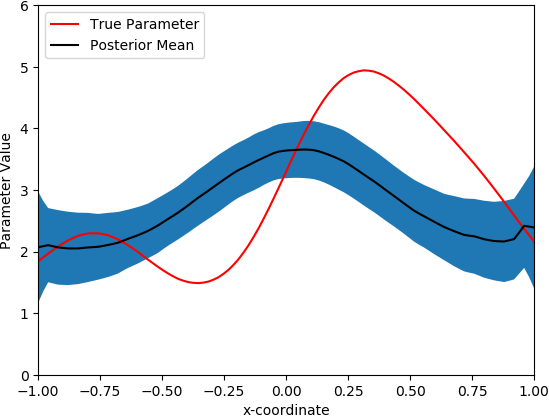}
    \includegraphics[scale=\figscale]{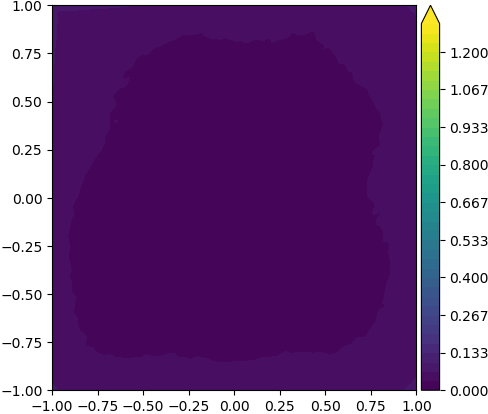}
    \caption{\scriptsize Top row left to right: mesh with sensors denoted with a red cross,
        true PoI,
        cross-sectional uncertainty estimate and pointwise posterior
        variance from Laplace approximation.
        Second to fourth rows: $\penJS =
        0.00001,0.001,0.1,0.5$. First and third columns: cross-sectional
        uncertainty estimates. Second and fourth columns: approximate pointwise
        posterior variance. First and second columns: exact PtO map.  Third and
        fourth columns: learned PtO map.}
        \label{FigPoisson2Dn2601nspt05M5000}
\end{figure}

%------------------------------------------------------------------------------
\subsection{$\noiselevel=0.1$, $\numdatatrain=50$} \label{SecHeat2Dn2601ns10M50}
%------------------------------------------------------------------------------
\begin{table}[H]
    \scriptsize
    \centering
    \begin{tabular}{c | c | c | c |}
        \cline{2-4}
        & \multicolumn{2}{c|}{Relative Error: $\param$} & \multicolumn{1}{c|}{Relative Error: $\stateobs$}\\
        \hline
        \multicolumn{1}{|c|}{$\penJS$} & Exact PtO & Learned PtO & Learned PtO\\
        \hline
        \multicolumn{1}{|c|}{0.00001} & 34.95\% & 36.13\% & 30.69\% \\
        \multicolumn{1}{|c|}{0.001}   & 37.49\% & 36.15\% & 27.56\% \\
        \multicolumn{1}{|c|}{0.1}     & 35.13\% & 34.02\% & 40.31\% \\
        \multicolumn{1}{|c|}{0.5}     & 38.44\% & 34.44\% & 38.70\% \\
        \hline
    \end{tabular}
    \caption{\scriptsize Table displaying the relative errors for UQ-VAE.
            Relative error of MAP estimate: 45.48\%.}
    \label{TableRelativeErrorsn2601nspt1M50}
\end{table}
\begin{figure}[H]
    \centering
    \includegraphics[scale=\figscale]{Figures/poisson_2d/general/mesh_n2601.png}
    \includegraphics[scale=\figscale]{Figures/poisson_2d/general/parameter_test_n2601_128.png}
    \includegraphics[scale=\figscale]{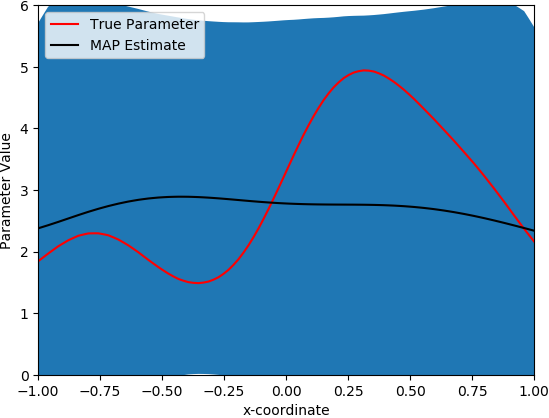}
    \includegraphics[scale=\figscale]{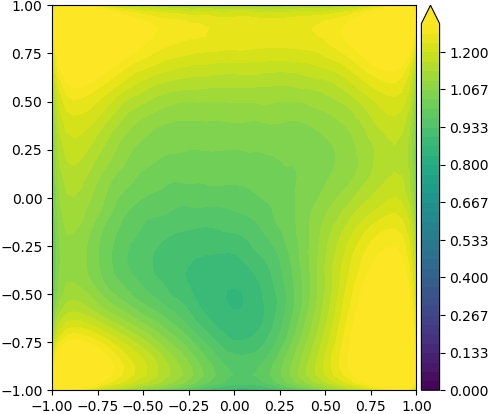}\\
    \includegraphics[scale=\figscale]{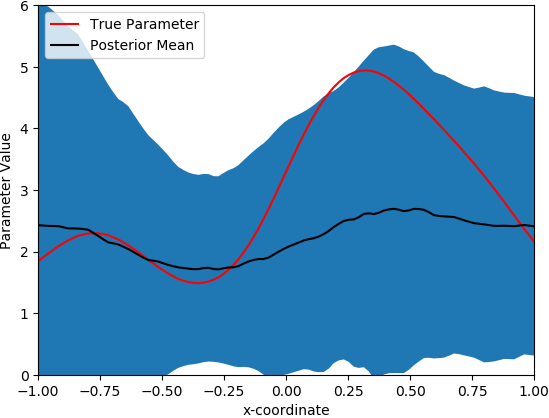}
    \includegraphics[scale=\figscale]{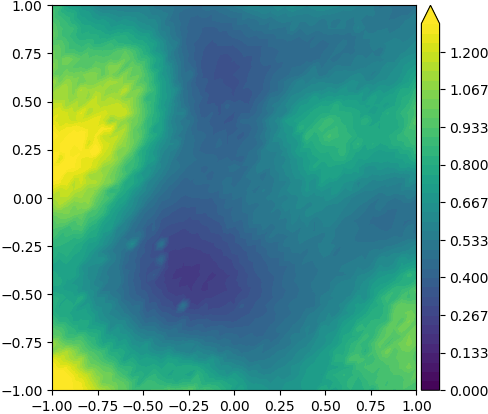}
    \includegraphics[scale=\figscale]{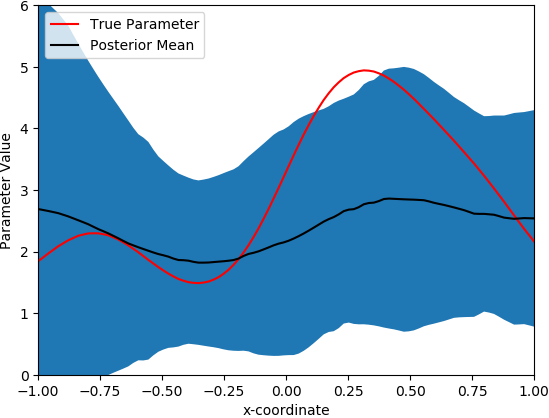}
    \includegraphics[scale=\figscale]{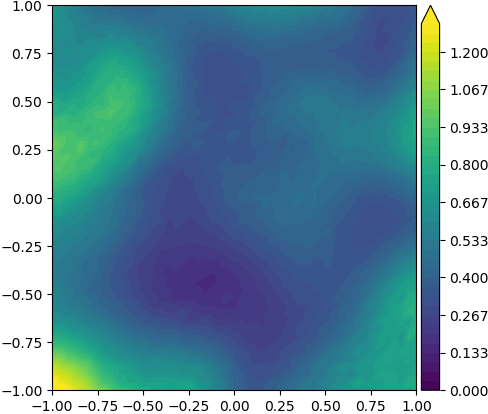}\\
    \includegraphics[scale=\figscale]{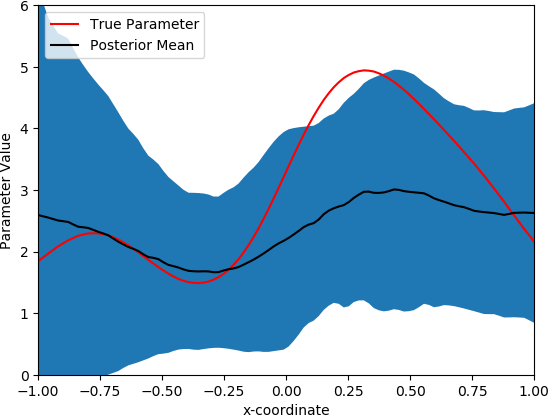}
    \includegraphics[scale=\figscale]{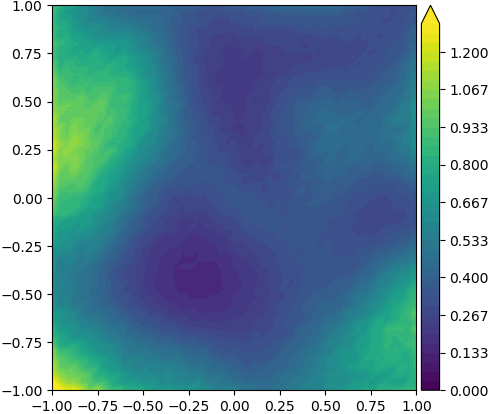}
    \includegraphics[scale=\figscale]{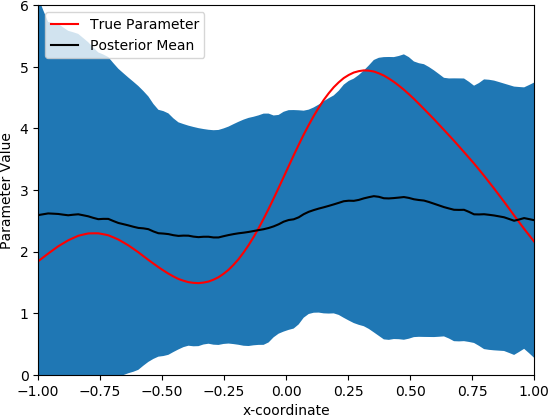}
    \includegraphics[scale=\figscale]{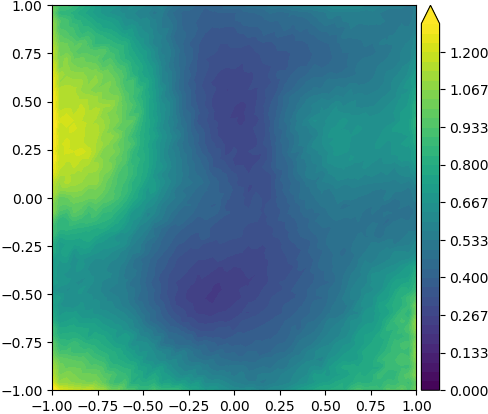}\\
    \includegraphics[scale=\figscale]{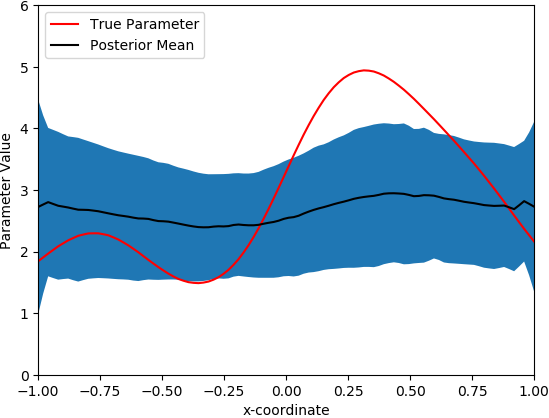}
    \includegraphics[scale=\figscale]{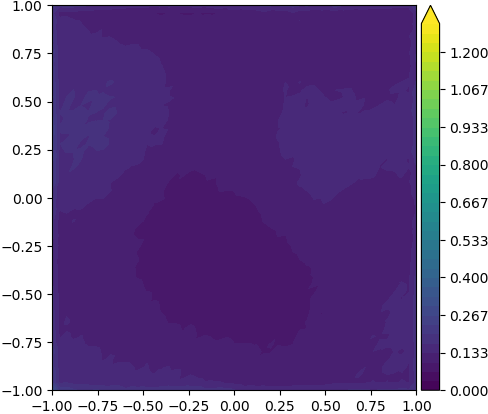}
    \includegraphics[scale=\figscale]{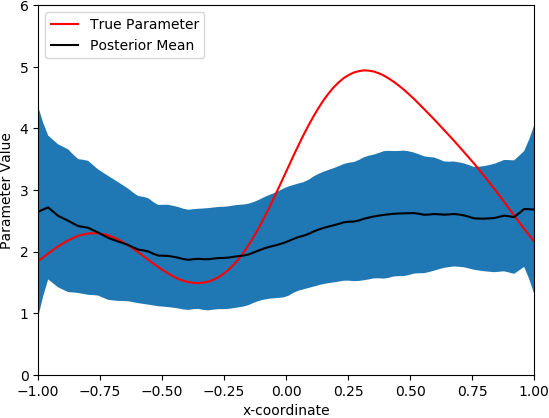}
    \includegraphics[scale=\figscale]{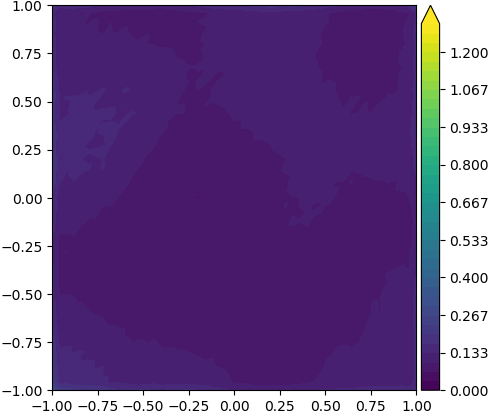}\\
    \includegraphics[scale=\figscale]{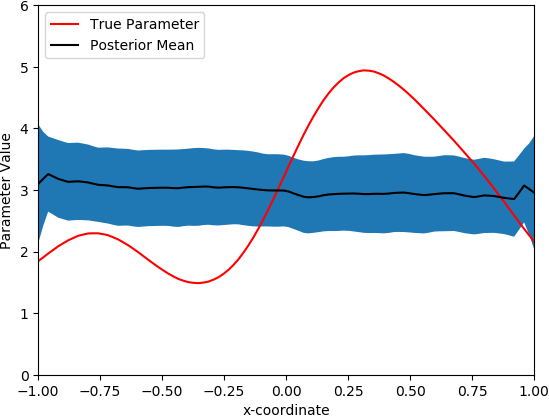}
    \includegraphics[scale=\figscale]{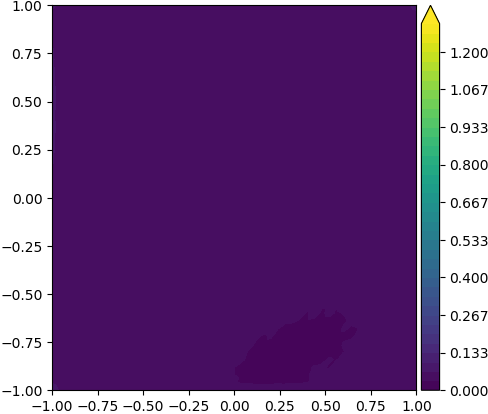}
    \includegraphics[scale=\figscale]{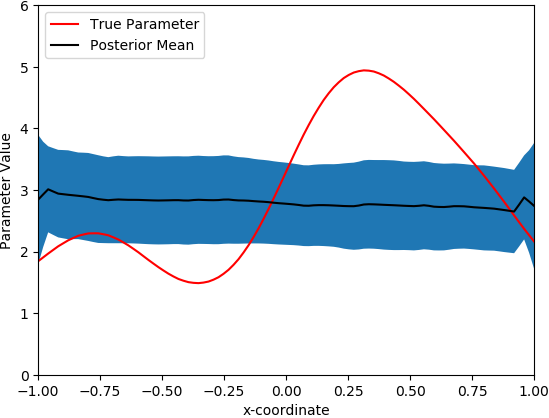}
    \includegraphics[scale=\figscale]{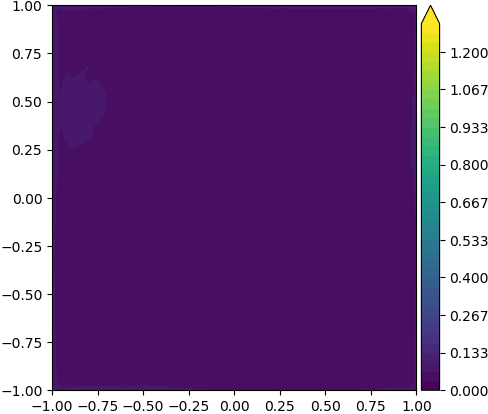}
    \caption{\scriptsize Top row left to right: mesh with sensors denoted with a red cross,
        true PoI,
        cross-sectional uncertainty estimate and pointwise posterior
        variance from Laplace approximation.
        Second to fourth rows: $\penJS =
        0.00001,0.001,0.1,0.5$. First and third columns: cross-sectional
        uncertainty estimates. Second and fourth columns: approximate pointwise
        posterior variance. First and second columns: exact PtO map.  Third and
        fourth columns: learned PtO map.}
        \label{FigPoisson2Dn2601nspt1M50}
\end{figure}

%------------------------------------------------------------------------------
\subsection{$\noiselevel=0.1$, $\numdatatrain=500$} \label{SecHeat2Dn2601ns10M500}
%------------------------------------------------------------------------------
\begin{table}[H]
    \scriptsize
    \centering
    \begin{tabular}{c | c | c | c |}
        \cline{2-4}
        & \multicolumn{2}{c|}{Relative Error: $\param$} & \multicolumn{1}{c|}{Relative Error: $\stateobs$}\\
        \hline
        \multicolumn{1}{|c|}{$\penJS$} & Exact PtO & Learned PtO & Learned PtO\\
        \hline
        \multicolumn{1}{|c|}{0.00001} & 36.61\% & 38.59\% & 23.15\% \\
        \multicolumn{1}{|c|}{0.001}   & 37.61\% & 37.82\% & 23.02\% \\
        \multicolumn{1}{|c|}{0.1}     & 35.77\% & 36.01\% & 30.22\% \\
        \multicolumn{1}{|c|}{0.5}     & 33.78\% & 35.37\% & 34.93\% \\
        \hline
    \end{tabular}
    \caption{\scriptsize Table displaying the relative errors for UQ-VAE.
            Relative error of MAP estimate: 45.48\%.}
    \label{TableRelativeErrorsn2601nspt1M500}
\end{table}
\begin{figure}[H]
    \centering
    \includegraphics[scale=\figscale]{Figures/poisson_2d/general/mesh_n2601.png}
    \includegraphics[scale=\figscale]{Figures/poisson_2d/general/parameter_test_n2601_128.png}
    \includegraphics[scale=\figscale]{Figures/poisson_2d/nspt1/n2601/traditional/parameter_cross_section.png}
    \includegraphics[scale=\figscale]{Figures/poisson_2d/nspt1/n2601/traditional/posterior_covariance.png}\\
    \includegraphics[scale=\figscale]{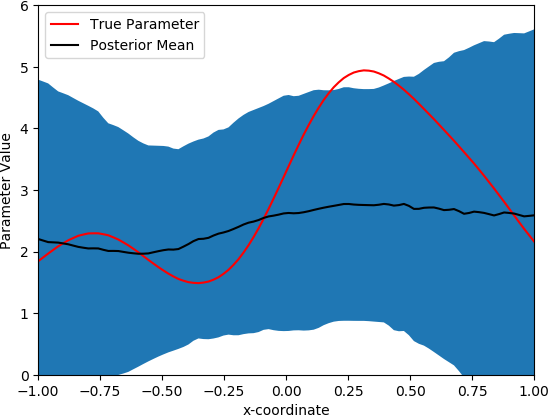}
    \includegraphics[scale=\figscale]{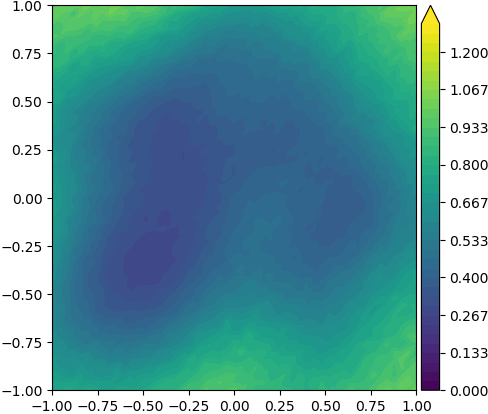}
    \includegraphics[scale=\figscale]{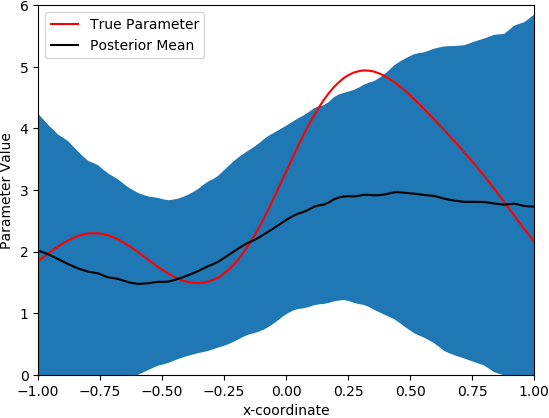}
    \includegraphics[scale=\figscale]{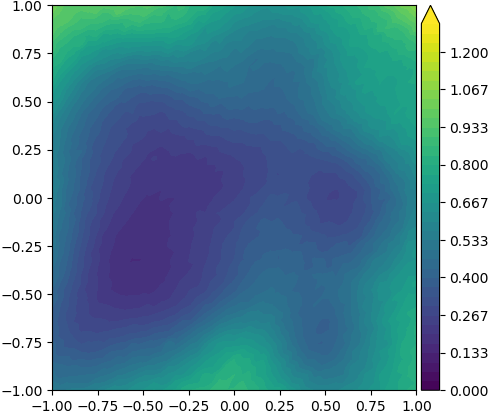}\\
    \includegraphics[scale=\figscale]{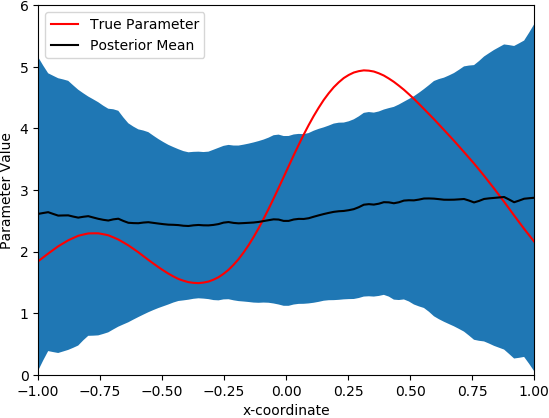}
    \includegraphics[scale=\figscale]{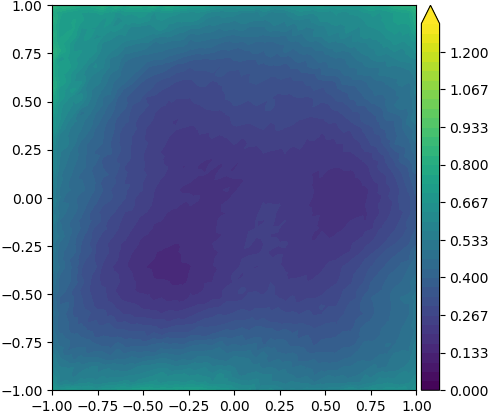}
    \includegraphics[scale=\figscale]{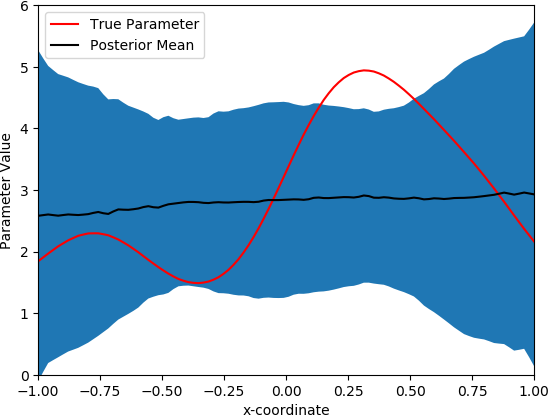}
    \includegraphics[scale=\figscale]{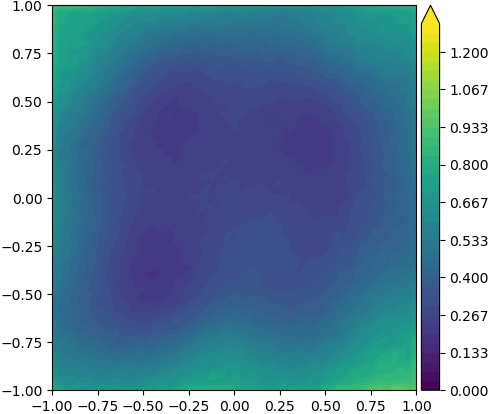}\\
    \includegraphics[scale=\figscale]{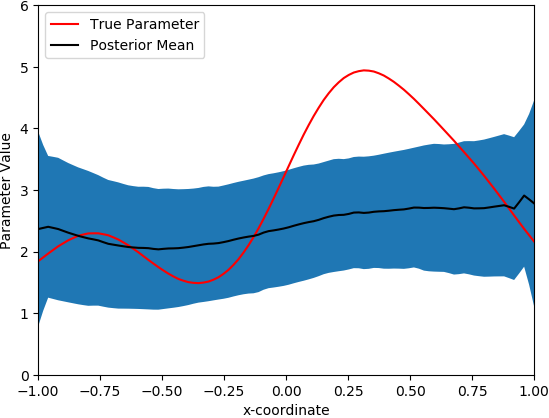}
    \includegraphics[scale=\figscale]{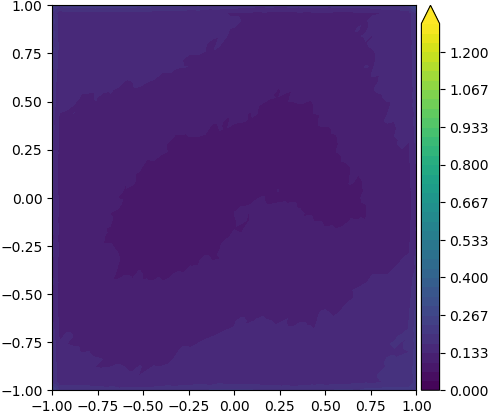}
    \includegraphics[scale=\figscale]{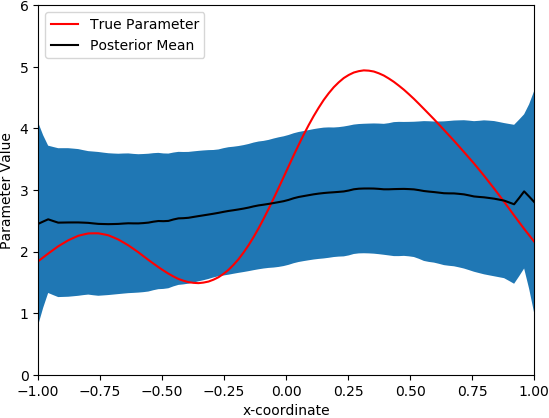}
    \includegraphics[scale=\figscale]{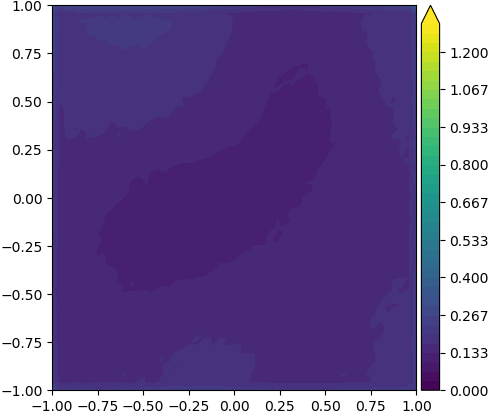}\\
    \includegraphics[scale=\figscale]{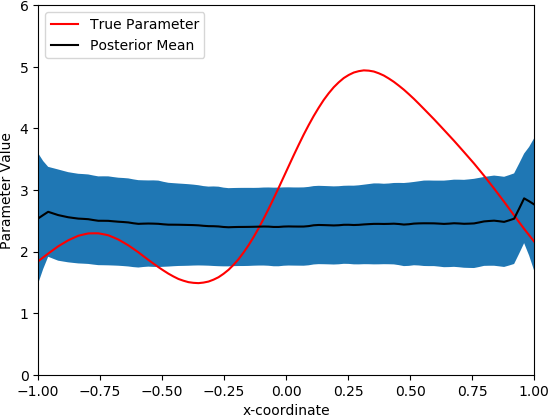}
    \includegraphics[scale=\figscale]{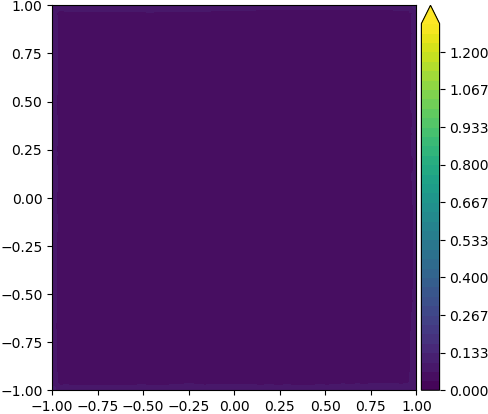}
    \includegraphics[scale=\figscale]{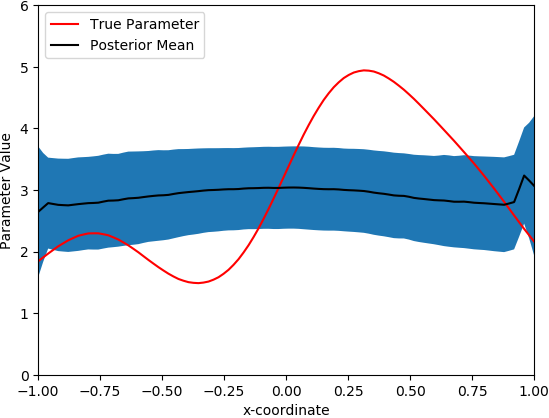}
    \includegraphics[scale=\figscale]{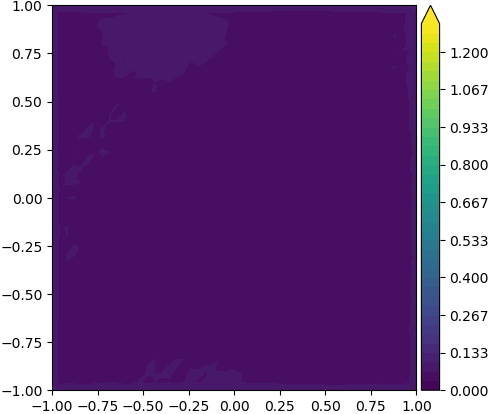}
    \caption{\scriptsize Top row left to right: mesh with sensors denoted with a red cross,
        true PoI,
        cross-sectional uncertainty estimate and pointwise posterior
        variance from Laplace approximation.
        Second to fourth rows: $\penJS =
        0.00001,0.001,0.1,0.5$. First and third columns: cross-sectional
        uncertainty estimates. Second and fourth columns: approximate pointwise
        posterior variance. First and second columns: exact PtO map.  Third and
        fourth columns: learned PtO map.}
        \label{FigPoisson2Dn2601nspt1M500}
\end{figure}

%------------------------------------------------------------------------------
\subsection{$\noiselevel=0.1$, $\numdatatrain=1000$} \label{SecHeat2Dn2601ns10M1000}
%------------------------------------------------------------------------------
\begin{table}[H]
    \scriptsize
    \centering
    \begin{tabular}{c | c | c | c |}
        \cline{2-4}
        & \multicolumn{2}{c|}{Relative Error: $\param$} & \multicolumn{1}{c|}{Relative Error: $\stateobs$}\\
        \hline
        \multicolumn{1}{|c|}{$\penJS$} & Exact PtO & Learned PtO & Learned PtO\\
        \hline
        \multicolumn{1}{|c|}{0.00001} & 39.96\% & 32.59\% & 19.83\% \\
        \multicolumn{1}{|c|}{0.001}   & 38.11\% & 31.85\% & 20.87\% \\
        \multicolumn{1}{|c|}{0.1}     & 37.77\% & 30.87\% & 26.37\% \\
        \multicolumn{1}{|c|}{0.5}     & 35.51\% & 33.09\% & 26.62\% \\
        \hline
    \end{tabular}
    \caption{\scriptsize Table displaying the relative errors for UQ-VAE.
            Relative error of MAP estimate: 45.48\%.}
    \label{TableRelativeErrorsn2601nspt1M1000}
\end{table}
\begin{figure}[H]
    \centering
    \includegraphics[scale=\figscale]{Figures/poisson_2d/general/mesh_n2601.png}
    \includegraphics[scale=\figscale]{Figures/poisson_2d/general/parameter_test_n2601_128.png}
    \includegraphics[scale=\figscale]{Figures/poisson_2d/nspt1/n2601/traditional/parameter_cross_section.png}
    \includegraphics[scale=\figscale]{Figures/poisson_2d/nspt1/n2601/traditional/posterior_covariance.png}\\
    \includegraphics[scale=\figscale]{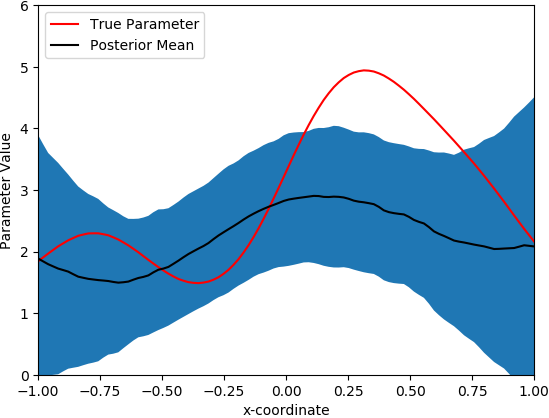}
    \includegraphics[scale=\figscale]{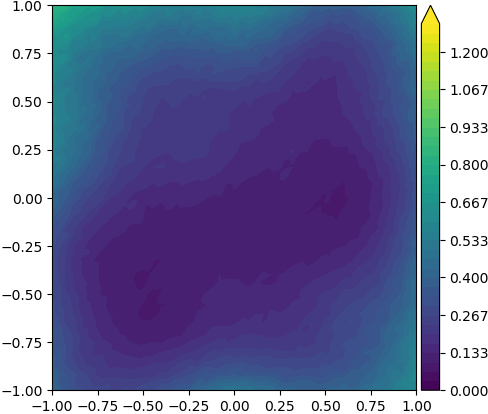}
    \includegraphics[scale=\figscale]{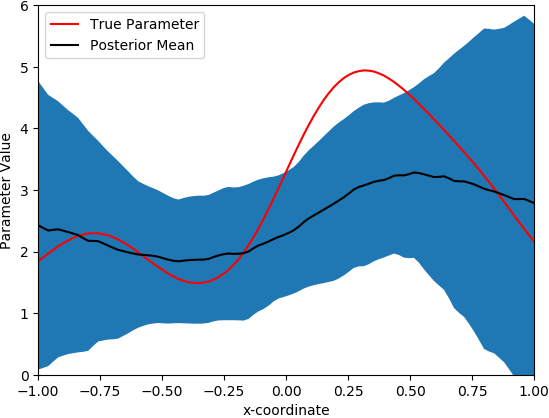}
    \includegraphics[scale=\figscale]{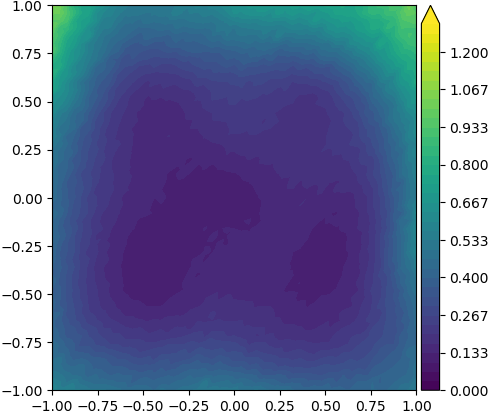}\\
    \includegraphics[scale=\figscale]{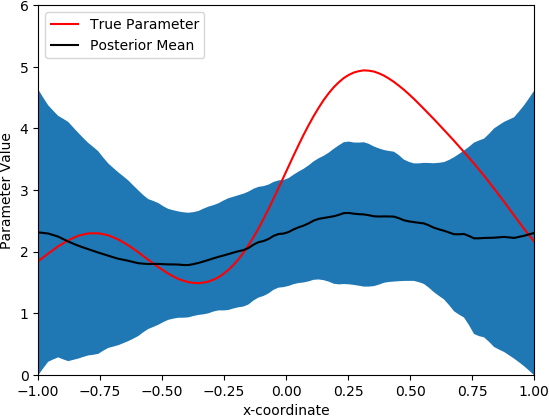}
    \includegraphics[scale=\figscale]{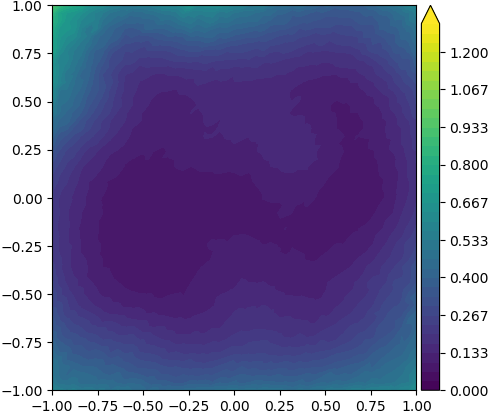}
    \includegraphics[scale=\figscale]{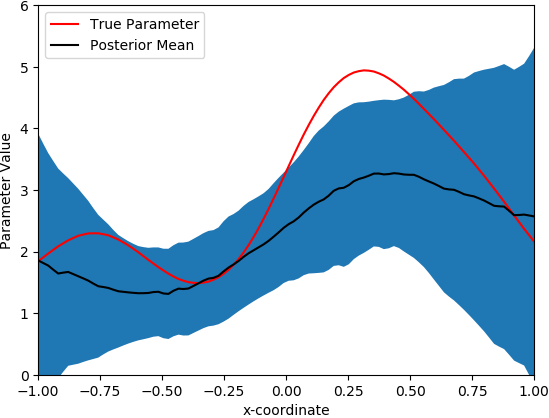}
    \includegraphics[scale=\figscale]{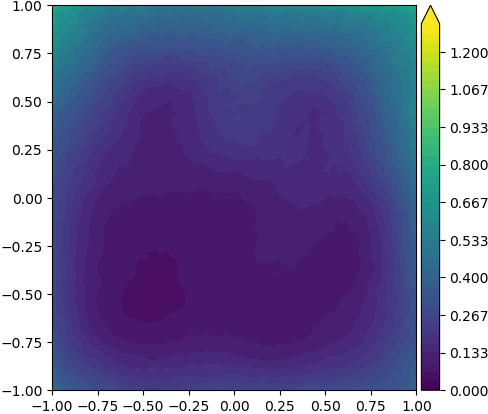}\\
    \includegraphics[scale=\figscale]{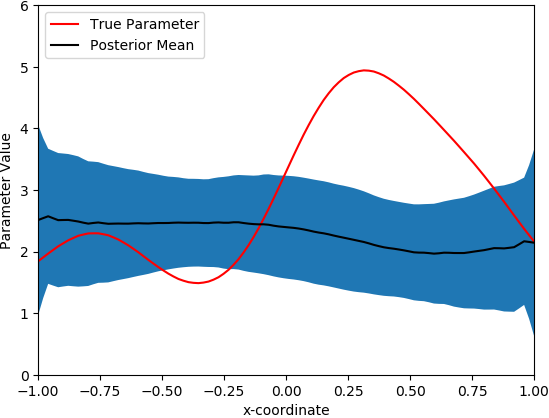}
    \includegraphics[scale=\figscale]{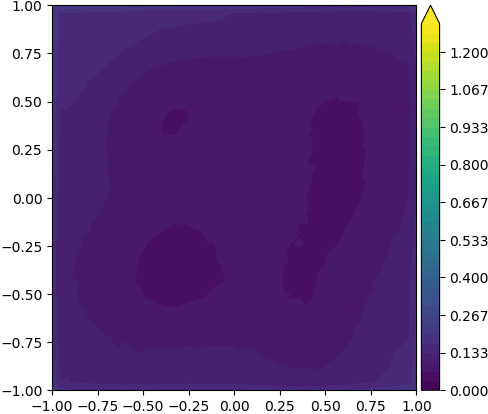}
    \includegraphics[scale=\figscale]{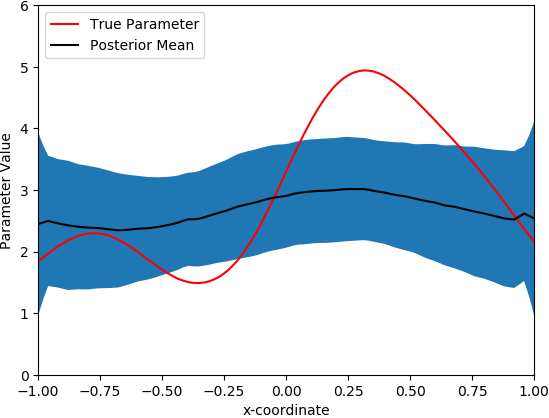}
    \includegraphics[scale=\figscale]{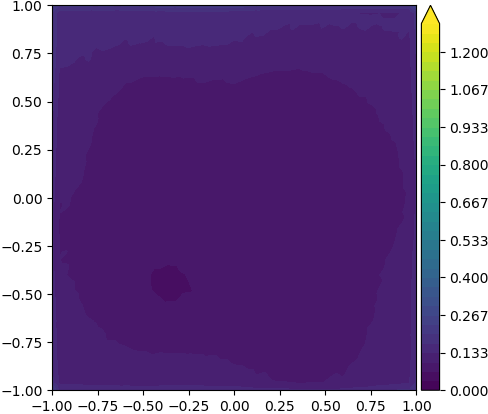}\\
    \includegraphics[scale=\figscale]{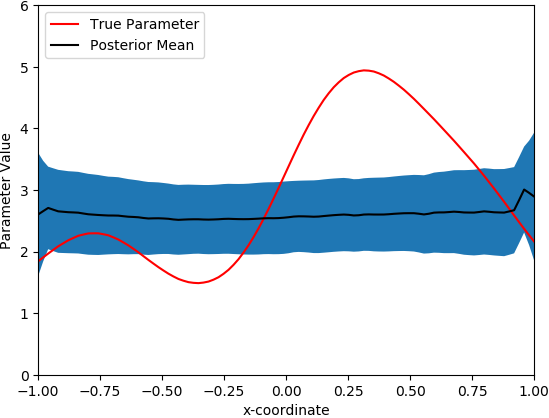}
    \includegraphics[scale=\figscale]{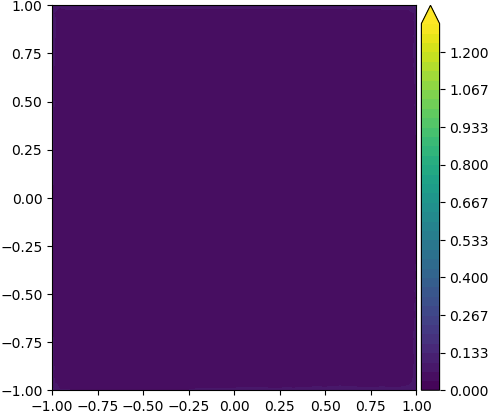}
    \includegraphics[scale=\figscale]{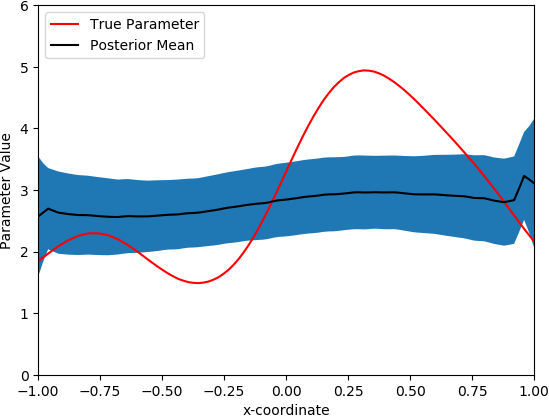}
    \includegraphics[scale=\figscale]{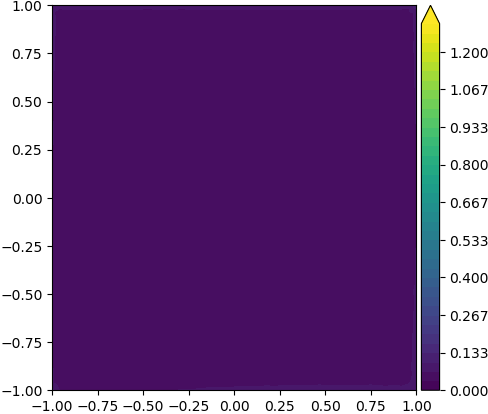}
    \caption{\scriptsize Top row left to right: mesh with sensors denoted with a red cross,
        true PoI,
        cross-sectional uncertainty estimate and pointwise posterior
        variance from Laplace approximation.
        Second to fourth rows: $\penJS =
        0.00001,0.001,0.1,0.5$. First and third columns: cross-sectional
        uncertainty estimates. Second and fourth columns: approximate pointwise
        posterior variance. First and second columns: exact PtO map.  Third and
        fourth columns: learned PtO map.}
        \label{FigPoisson2Dn2601nspt1M1000}
\end{figure}

%------------------------------------------------------------------------------
\subsection{$\noiselevel=0.1$, $\numdatatrain=5000$} \label{SecHeat2Dn2601ns10M5000}
%------------------------------------------------------------------------------
\begin{table}[H]
    \scriptsize
    \centering
    \begin{tabular}{c | c | c | c |}
        \cline{2-4}
        & \multicolumn{2}{c|}{Relative Error: $\param$} & \multicolumn{1}{c|}{Relative Error: $\stateobs$}\\
        \hline
        \multicolumn{1}{|c|}{$\penJS$} & Exact PtO & Learned PtO & Learned PtO\\
        \hline
        \multicolumn{1}{|c|}{0.00001} & 40.45\% & 35.41\% & 15.64\% \\
        \multicolumn{1}{|c|}{0.001}   & 40.78\% & 35.82\% & 15.18\% \\
        \multicolumn{1}{|c|}{0.1}     & 41.26\% & 36.48\% & 15.39\% \\
        \multicolumn{1}{|c|}{0.5}     & 40.85\% & 36.89\% & 15.35\% \\
        \hline
    \end{tabular}
    \caption{\scriptsize Table displaying the relative errors for UQ-VAE.
            Relative error of MAP estimate: 45.48\%.}
    \label{TableRelativeErrorsn2601nspt1M5000}
\end{table}
\begin{figure}[H]
    \centering
    \includegraphics[scale=\figscale]{Figures/poisson_2d/general/mesh_n2601.png}
    \includegraphics[scale=\figscale]{Figures/poisson_2d/general/parameter_test_n2601_128.png}
    \includegraphics[scale=\figscale]{Figures/poisson_2d/nspt1/n2601/traditional/parameter_cross_section.png}
    \includegraphics[scale=\figscale]{Figures/poisson_2d/nspt1/n2601/traditional/posterior_covariance.png}\\
    \includegraphics[scale=\figscale]{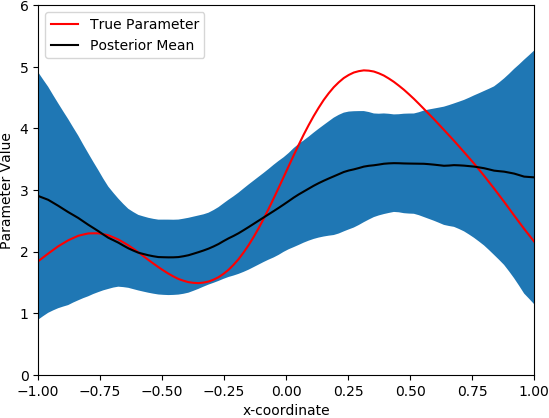}
    \includegraphics[scale=\figscale]{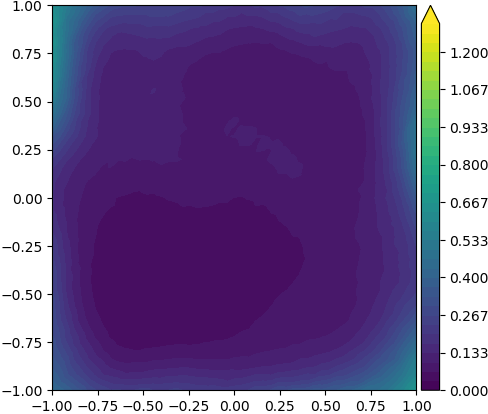}
    \includegraphics[scale=\figscale]{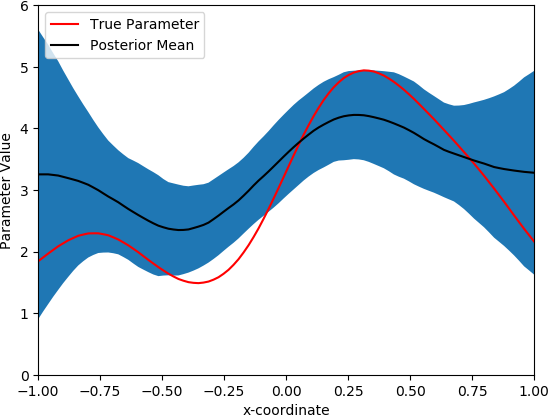}
    \includegraphics[scale=\figscale]{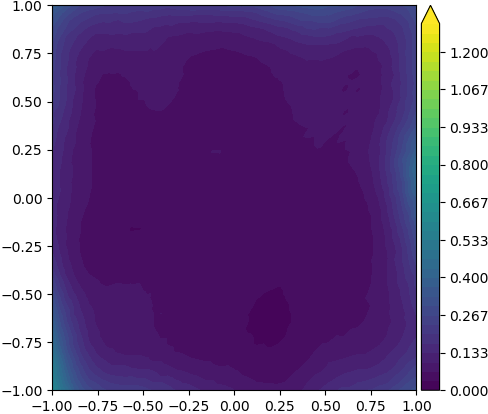}\\
    \includegraphics[scale=\figscale]{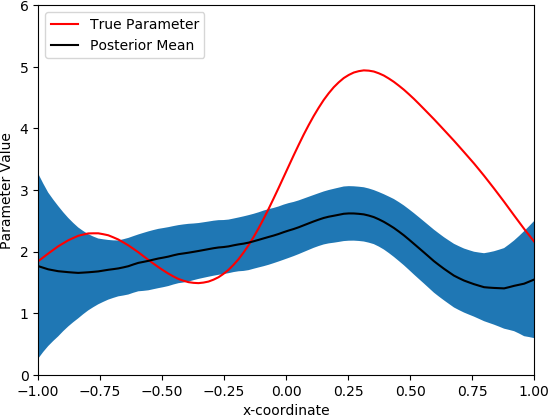}
    \includegraphics[scale=\figscale]{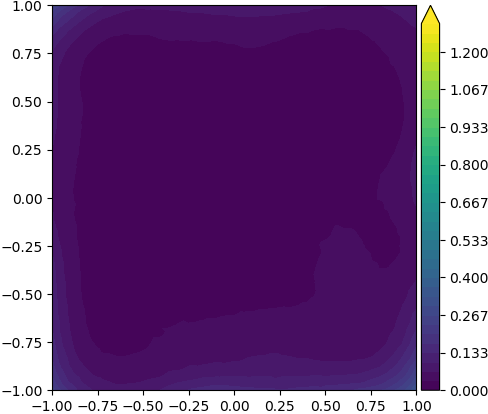}
    \includegraphics[scale=\figscale]{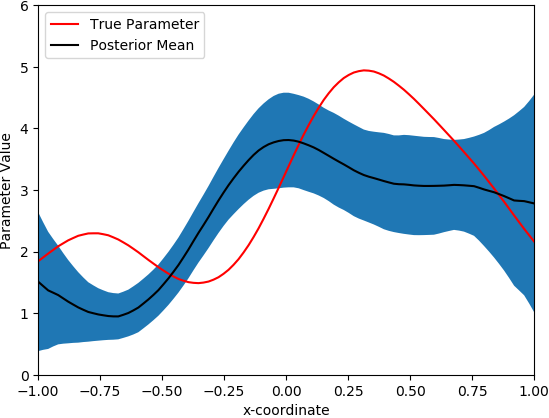}
    \includegraphics[scale=\figscale]{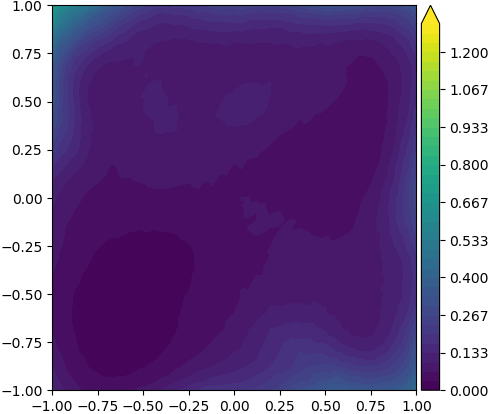}\\
    \includegraphics[scale=\figscale]{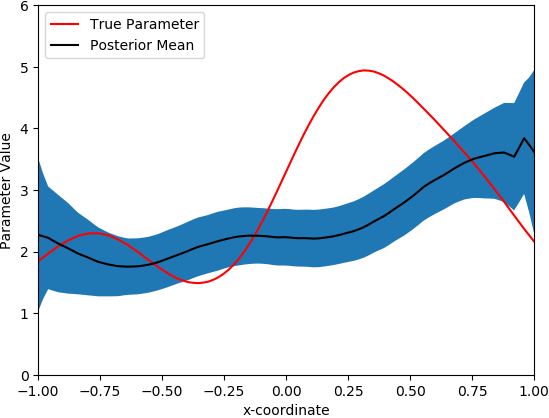}
    \includegraphics[scale=\figscale]{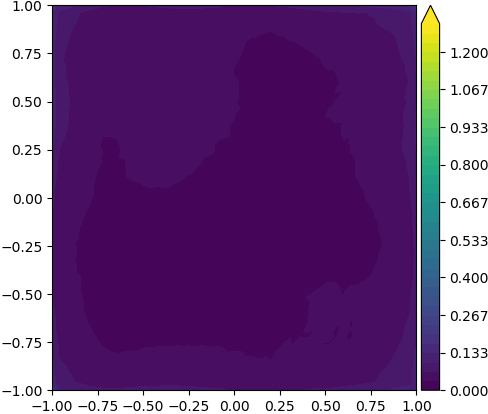}
    \includegraphics[scale=\figscale]{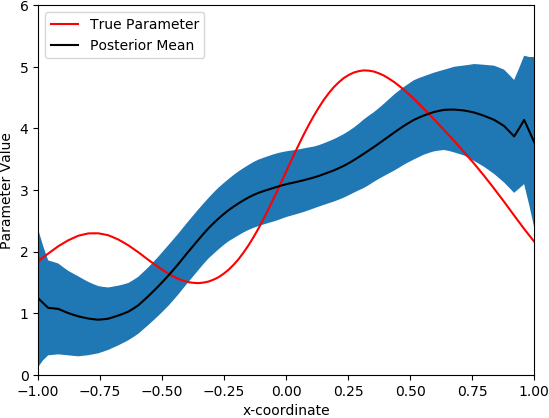}
    \includegraphics[scale=\figscale]{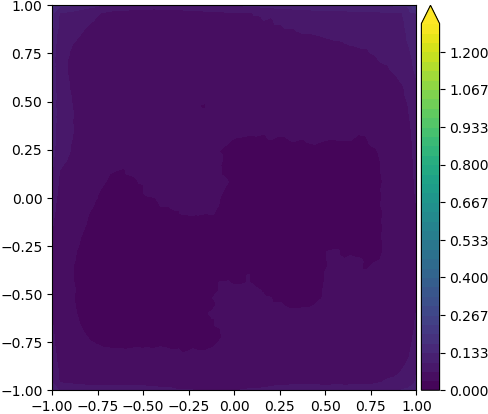}\\
    \includegraphics[scale=\figscale]{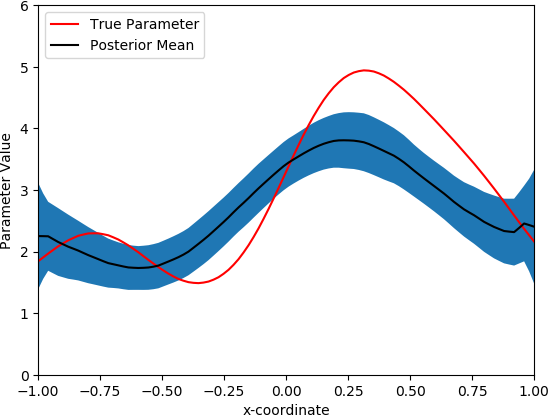}
    \includegraphics[scale=\figscale]{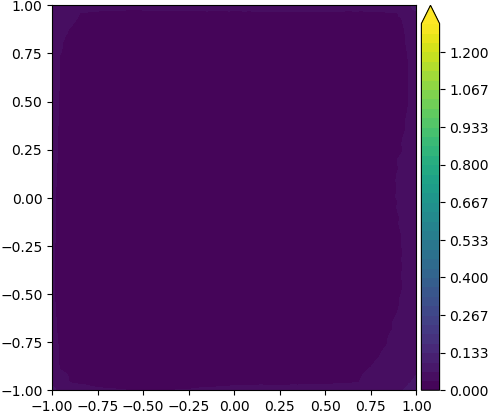}
    \includegraphics[scale=\figscale]{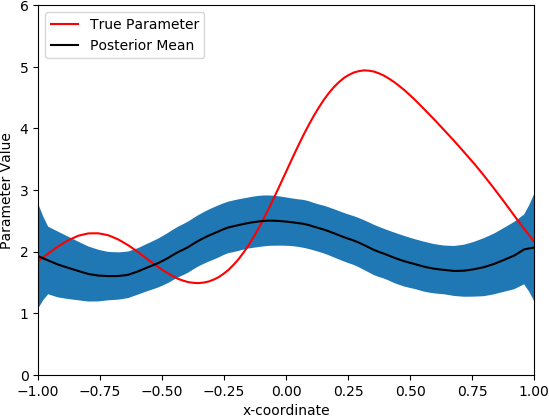}
    \includegraphics[scale=\figscale]{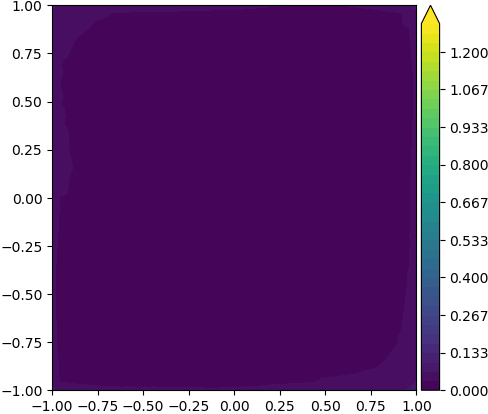}
    \caption{\scriptsize Top row left to right: mesh with sensors denoted with a red cross,
        true PoI,
        cross-sectional uncertainty estimate and pointwise posterior
        variance from Laplace approximation.
        Second to fourth rows: $\penJS =
        0.00001,0.001,0.1,0.5$. First and third columns: cross-sectional
        uncertainty estimates. Second and fourth columns: approximate pointwise
        posterior variance. First and second columns: exact PtO map.  Third and
        fourth columns: learned PtO map.}
        \label{FigPoisson2Dn2601nspt1M5000}
\end{figure}

\end{document}